\pdfoutput=1

\documentclass[11pt]{article}

\usepackage[final]{acl}
\usepackage{framed}
\usepackage{times}
\usepackage{latexsym}
\usepackage{listings}
\usepackage{makecell}
\usepackage[T1]{fontenc}

\usepackage[utf8]{inputenc}
\usepackage{booktabs}
\usepackage{array}
\usepackage{microtype}
\usepackage{inconsolata}
\usepackage{hyperref}

\usepackage{subcaption}
\usepackage{tabularx,makecell}
\usepackage{lipsum}  
\usepackage[many]{tcolorbox}
\PassOptionsToPackage{table,dvipsnames}{xcolor}
\usepackage[table]{xcolor}
\usepackage{colortbl}
\usepackage[colorinlistoftodos]{todonotes}

\usepackage{cleveref}
\definecolor{LightGray}{gray}{0.93}
\newcommand{\aspera}{ASPERA}
\newcommand{\asperaDataset}{\textit{Asper-Bench}}
\newcommand{\asperaAssistantCodebase}{ASPERA}
\crefname{appendix}{App.}{Apps.} 
\Crefname{appendix}{App.}{Apps.} 
\usepackage{enumitem}
\usepackage{relsize}
\usepackage{booktabs}
\usepackage{tabularx}
\usepackage{multirow}
\usepackage{graphicx}
\usepackage{xspace}
\usepackage{xcolor}
\usepackage{subcaption}
\usepackage{hyperref}

\usepackage{amsmath}
\usepackage{amssymb}
\usepackage{amsthm}
\usepackage{amsfonts}
\usepackage{mathtools}
\usepackage{pifont}

\usepackage{tablefootnote}

\usepackage{cleveref}

\usepackage{tikz}
\usetikzlibrary{arrows.meta}
\usetikzlibrary{shapes.arrows}
\usetikzlibrary{shapes.misc}
\usetikzlibrary{fit}
\usetikzlibrary{calc,shapes}
\usetikzlibrary{tikzmark}
\usetikzlibrary{arrows}

\crefname{chapter}{Chapter}{}
\crefname{section}{Section}{Sections}
\crefformat{section}{Section~#2#1#3}

\crefname{table}{Table}{Tables}
\crefname{figure}{Figure}{Figures}
\crefname{algorithm}{Alg.}{Algs.}
\crefname{line}{Line}{Lines}
\crefname{appendix}{App.}{}
\crefname{chapter}{Chapter}{Chapters}

\crefname{thm}{Theorem}{Theorems}
\crefname{prop}{Proposition}{Propositions}
\crefname{definition}{Definition}{Definitions}
\crefname{lemma}{Lemma}{Lemmas}
\crefname{cor}{Corollary}{Corollaries}
\crefname{equation}{Eq.}{Eqs.}

\creflabelformat{equation}{#2\textup{#1}#3}
\crefrangelabelformat{equation}{(#3#1#4--#5#2#6)}
\creflabelformat{figure}{#2\textup{#1}#3}
\crefrangelabelformat{figure}{(#3#1#4--#5#2#6)}

\newcolumntype{C}{>{\centering\arraybackslash}X}

\renewcommand{\ldots}{\ensuremath{{\ldotp\kern-0.2em\ldotp\kern-0.2em\ldotp}}}

\renewcommand{\cdots}{\ensuremath{{\cdotp\kern-0.2em\cdotp\kern-0.2em\cdotp}}}
\renewcommand{\dots}{\ensuremath{{\ldotp\kern-0.2em\ldotp\kern-0.2em\ldotp}}}

\usepackage[plain]{algorithm}
\usepackage{algorithmicx}
\usepackage[noend]{algpseudocode}
\algnewcommand{\parState}[1]{\State%
    \parbox[t]{\dimexpr\linewidth-\algmargin}{\strut\hangindent=\algorithmicindent \hangafter=1 #1\strut}}
\algrenewcommand\algorithmicindent{1.0em}%
\newcommand{\algorithmicdowhile}{\textbf{do}:}
\algdef{SE}[DOWHILE]{Do}{doWhile}{\algorithmicdowhile}[1]{\algorithmicwhile\ #1}%
\newcommand{\algorithmicfunc}[1]{\textbf{def} #1 :}
\algdef{SE}[FUNC]{Func}{EndFunc}[1]{\algorithmicfunc{#1}}{}

\newcommand{\algorithmicclass}[1]{\textbf{class} #1 :}
\algdef{SE}[CLASS]{Class}{EndClass}[1]{\algorithmicclass{#1}}{}

\newif\ifboldnumber

\algrenewcommand\alglinenumber[1]{%
  \footnotesize\ifboldnumber\color{red}\bfseries\fi\global\boldnumberfalse#1:}

\MakeRobust{\Call}   

\newcommand{\rightcomment}[1]{{\color{commentcolor} \(\triangleright\) {\footnotesize\textit{#1}}}}
\algrenewcommand{\algorithmiccomment}[1]{\hfill \rightcomment{#1}}  
\algnewcommand{\LineComment}[1]{\State \rightcomment{#1}}
\algnewcommand{\LinesComment}[1]{\State \rightcomment{\parbox[t]{\linewidth-\leftmargin-\widthof{\(\triangleright\) }}{#1}}}

\renewcommand\algorithmicthen{:}

\makeatletter
\ifthenelse{\equal{\ALG@noend}{t}}%
  {\algtext*{EndFunc}}
  {}%
\makeatother

\makeatletter
\ifthenelse{\equal{\ALG@noend}{t}}%
  {\algtext*{EndClass}}
  {}%
\makeatother

\algnewcommand{\IIf}[1]{\State\algorithmicif\ #1\ \algorithmicthen}
\algnewcommand{\EndIIf}{\unskip}

\definecolor{commentcolor}{rgb}{0.4, 0.22, 0.33}

\usepackage{todonotes}
\makeatletter
\newcommand*\iftodonotes{\if@todonotes@disabled\expandafter\@secondoftwo\else\expandafter\@firstoftwo\fi}  
\makeatother

\title{ASPERA: A Simulated Environment to Evaluate Planning for Complex Action Execution}

\author{
    \textbf{Alexandru Coca\textsuperscript{1}\thanks{Work done while at Apple.}},
    \textbf{Mark Gaynor\textsuperscript{2}},
    \textbf{Zhenxing Zhang\textsuperscript{2}},
    \textbf{Jianpeng Cheng\textsuperscript{3*}},
    \textbf{Bo-Hsiang Tseng\textsuperscript{2}}, \\
    \textbf{Pete Boothroyd\textsuperscript{2}},
    \textbf{Héctor Martinez Alonso\textsuperscript{2}},
    \textbf{Diarmuid Ó Séaghdha\textsuperscript{2}},
    \textbf{Anders Johannsen\textsuperscript{2}}
    \\
    \\
    \textsuperscript{1}Department of Engineering, University of Cambridge,
    \textsuperscript{2}Apple,                                          
    \textsuperscript{3}Meta
    \\ 
    \texttt{ac2123@cam.ac.uk}, \texttt{ajohannsen@apple.com}
}

\begin{document}
\maketitle
\begin{abstract}

This work evaluates the potential of large language models (LLMs) to power digital assistants capable of complex action execution. These assistants rely on pre-trained programming knowledge to execute multi-step goals by composing objects and functions defined in assistant libraries into action execution programs. To achieve this, we develop \aspera{}, a framework comprising an assistant library simulation and a human-assisted LLM data generation engine. Our engine allows developers to guide LLM generation of high-quality tasks consisting of complex user queries, simulation state and corresponding validation programs, tackling data availability and evaluation robustness challenges. Alongside the framework we release \asperaDataset{}, an evaluation dataset of $250$ challenging tasks generated using \aspera{}, which we use to show that program generation grounded in custom assistant libraries is a significant challenge to LLMs compared to dependency-free code generation\footnote{\url{https://github.com/apple/ml-aspera}.}.

\end{abstract}

\section{Introduction}
\label{sec:intro}

Digital assistants, such as Siri and Alexa, provide a conversational interface for users to execute \textit{simple actions} (e.g., \textit{Set a timer for 5 minutes}). To achieve this, developers typically define APIs (intents) and collect data to train specialised parsing models responsible for translating user requests into machine-interpretable, domain-specific languages that can execute these APIs \cite{smcalflow, DBLP:conf/emnlp/ChengAABDFKKLPW20}. Equivalently, action execution in this setting can be modelled as a function call to an intent API implemented by a target application (e.g., \verb|alarm_set_timer(duration=5, unit='min'|))

\begin{figure}[t]
  \includegraphics[width=\columnwidth, trim=0cm 2.2cm 0cm 1cm, clip]{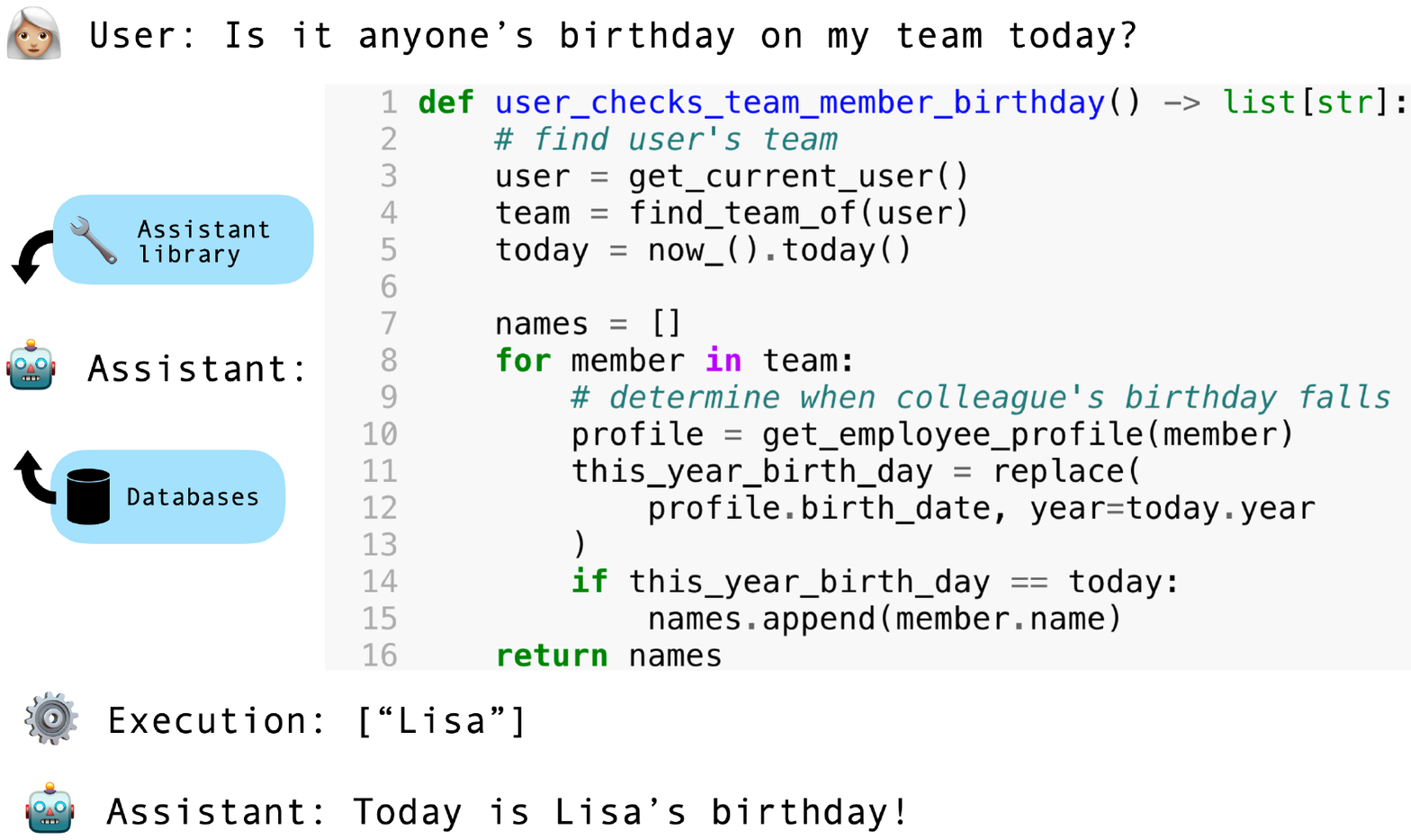}
  \caption{Example of a digital assistant executing a complex action given primitives (e.g., \texttt{now\_}) defined in a custom assistant library and databases containing user's data. The assistant decomposes  the query and calls $5$ APIs (lines 3 - 5, 9 - 10), performing attribute access, passing values by attribute reference (line 11 - 13), in addition to iteration and flow control in a multi-step program to achieve the user's goal. Logical reasoning is required to deduce that the year of the birthday has to be updated to the current year.} 

  \label{fig:plan}
  \vspace{-15pt}
\end{figure}
Function calling supports simple actions, but extension to execute \textit{any} action on the device requires implementation of fine-grained intents and/or specialised parsing functions for an intractably large number of requests. To enable future digital assistants to execute \textit{complex actions} (Figure \ref{fig:plan}), \citet{jhamtani2023natural} propose generation of a program implemented with low-level \textit{primitives} from assistant libraries\footnote{The assistant library is a collection of functions and objects the assistant can use to compose plans which determine or change the user's device state. A \textit{primitive} is any abstraction implemented in the library (e.g., a function or class).}. 

We aim to evaluate the ability of LLMs to generate such programs when (1) the LLM has access to all the relevant information for generation, encoded in the assistant library documentation, or (2) the LLM selects the relevant primitives by exploring the entire assistant library as a first step prior to program generation. To this end, we address two challenges.

\textbf{1. Complex action evaluation instances} comprising diverse, realistic queries annotated with programs requiring compositional use of multiple primitives are required for evaluation. Existing resources do not fully satisfy this requirement. SMCalFlow \cite{smcalflow} contains compositional queries but is annotated with a specialised domain-specific language (DSL) which hinders LLM performance \cite{DBLP:conf/naacl/BoginGCS24}. DeCU \cite{jhamtani2023natural} is a dataset for evaluating plan generation for complex user queries, but evaluates solutions using an LLM judge rather than implementing an executable simulation of the assistant library. In lieu of documentation—which has been shown by recent research to improve LLM performance on many tasks \cite{agentsanbox, optimiseICL,  toolusagedocs}—DeCU provides only in-context examples (ICEs) primarily demonstrating how to parse simple user queries into single-instruction programs. \citet{workbench} and \citet{appworld} develop simulated environments with comprehensively documented APIs, but limit action diversity by grounding queries in task templates. Unlike the approach proposed in this paper, environment simulation and functional correctness validation in these environments relies on human effort and expertise alone.  

\textbf{2. Robust evaluation} of complex action execution capability requires measuring task success, i.e. that the assistant actions satisfy the user goal. \citet{jhamtani2023natural} note this to be an open problem, since functional correctness evaluation requires query-dependent databases and accounting for unwarranted side-effects\footnote{This term describes an unintended action by the agent e.g., setting a meeting with the wrong attendees.}.  \citet{workbench} tackle this by feeding databases to templated executable programs to annotate expected environment states. They propose strict database comparisons to estimate task success, and hence cannot evaluate queries with multiple outcomes and \textit{information-seeking queries}\footnote{Queries where information is provided to the user.}. \citet{appworld} address these limitations, but define environment states and evaluate task success via specialised programs implemented by domain experts for every task. 

\paragraph{Contributions} We propose \aspera{}, a simulated environment supporting evaluation of agents capable of complex action execution with data generation capability. Given an assistant library simulation (\S \ref{sec:assistat-library}), \aspera{} enables a developer and an LLM to interact to generate diverse, high-quality complex user requests and programs which satisfy them. We show that robust task success estimation is possible for both synthesised and human-authored queries by prompting LLMs to generate programs which appropriately initialise the environment state and determine whether the executed action satisfies the user goal (\S\ref{sec:SIP} and \ref{sec:EP}). Using this system, we address the lack of complex actions execution data by generating \asperaDataset{}, a challenging collection of $250$ tasks (\S \ref{sec:dataset}). Evaluation on this dataset shows that (1) generating programs that satisfy complex action requests is a challenge for LLMs even when they are prompted with all the relevant information, despite their ability to generate plausible programs and (2) state-of-the-art (SOTA) LLMs find it difficult to select all the primitives needed for composite tasks, adding a challenge to program generation (\S \ref{sec:results} and \ref{sec:analysis}).

\section{The \aspera{} Framework }

In \aspera{}, a human developer initiates an interactive session in which an LLM is prompted to generate complex user requests grounded in a \texttt{python} library which can implement digital assistant use cases. In subsequent human-LLM interactions, two additional programs which enable success rate evaluation for arbitrary agents are generated. We now discuss how this works in practice.
\begin{table}[t]
  \centering
  \tiny 
  \resizebox{\columnwidth}{!}{
  \begin{tabular}{p{2.5cm}ccc}
    \toprule
    \textbf{Module}           & \textbf{Functions} & \textbf{Classes} & \textbf{Docs length (words)} \\
    \midrule
    time utils                & 22                 & 11               & 986                          \\
    work calendar             & 13                 & 3                & 660                          \\
    company directory         & 10                 & 3                & 236                          \\
    room booking              & 4                  & 2                & 331                          \\
    exceptions                & -                  & 1                & 209                          \\
    \midrule
    \textbf{Total}            & 49                 & 20               & 2,422                           \\
    \bottomrule
  \end{tabular}
  }
  \caption{\label{domain-functions-classes-docs} Assistant library summary statistics. A module corresponds to a \texttt{.py} file. Docs length is the total length of the documentation strings defined inside the module. See \Cref{appendix:aspera-codebase},\texttt{src/aspera/apps\_implementation} and \texttt{src/aspera/apps} in our code release for details.}
  \label{tab:statistics}
  \vspace{-15pt}
\end{table}
\subsection{Assistant library}
\label{sec:assistat-library}
\aspera{} implements an assistant library which simulates a company in which employees in various teams (with a tree-based reporting structure) have meetings with one another under various conditions, managed by a room booking system. The library consists of $7$ databases and $69$ \verb|python| primitives (Table \ref{tab:statistics}). An extensive time utilities library, partially inspired by SMCalFlow \cite{smcalflow}, is implemented to test logical and arithmetic reasoning capabilities.

\begin{figure*}[htpb]
  \centering
  \includegraphics[width=\linewidth, trim=0.2cm 0.2cm 0cm 0cm, clip]{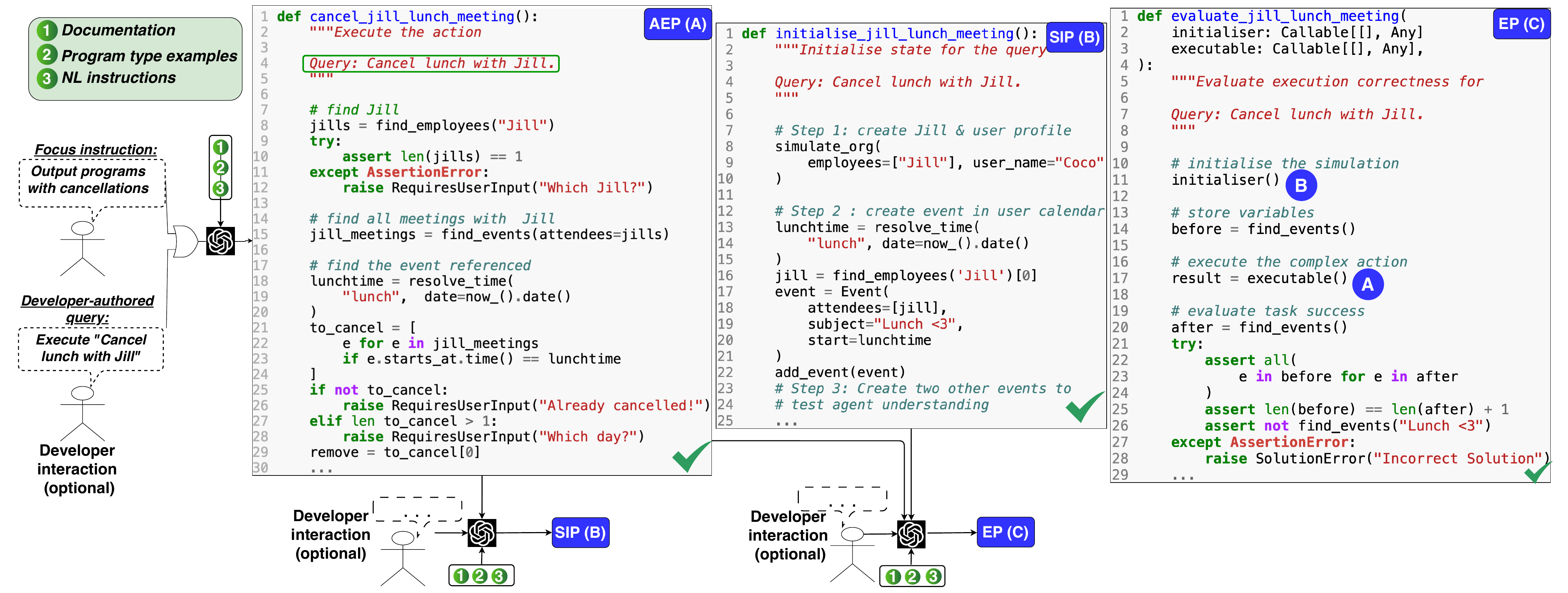}
  \caption{Sample ASPERA task, depicting action execution (A), state initialisation (B) and evaluation (C) programs. The task is generated in an interactive chat session (\S \ref{sec:aspera-interactive}) which is initialised with AEP generation prompts (\Cref{appendix:aep_gen} or \ref{appendix:aep-annotation}). To ground state initialisation, the chat history is extended with SIP generation prompt (\Cref{appendix:sip-gen}), which developers can customise with task-specific instructions. Finally, the chat history is extended with the EP generation prompt ( \Cref{appendix:ep-gen}) which can also be customised by developers via instructions. At each step, the developer can execute and edit the generated programs to ensure data quality. 
  \label{fig:aspera-session}}
\end{figure*}
\subsection{Components of an \aspera{} task}
\label{sec:aspera-task}
A task generated by \aspera{} has four elements: (1) the \textit{user query}, a natural-language request for the assistant to execute an action (e.g., \textit{Cancel my lunch with Jill}); (2) the \textit{action execution program} (AEP), a program which satisfies the user request upon execution; (3) the \textit{state initialisation program} (SIP), which uses the assistant library and simulation tools to set the environment state so that the query can be executed in \texttt{python} (i.e., establishing the existence of an employee named Jill and some meetings scheduled with her); (4) the \textit{evaluation program} (EP) which runs the AEP in the initialised environment and determines its correctness (i.e., checks that the correct meeting has been deleted). \cref{fig:aspera-session} depicts a simple \aspera{} task.

\subsection{\aspera{} task generation}
\label{sec:asperataskgeneration}

Figure \ref{fig:aspera-session} shows that the three programs which comprise a task are generated given: (1) assistant library documentation; (2) ICEs demonstrating the program format; and (3) natural language instructions. The instructions describe the assistant policy, environment assumptions and/or program structure information (depending on the program type to be generated). 

\subsubsection{Query and AEP generation}
\label{sec:query-gen}
The user query can be authored by the human developer or synthesised by the LLM  with the AEP (as part of the AEP docstring -- see prompts in \Cref{appendix:aep_gen} and \ref{appendix:aep-annotation}). By prompting the LLM with the documentation of the assistant library and with suitable examples, diverse and complex AEPs are generated. The complexity of the generated AEPs is characterised by: (1) number of primitives; (2) a variety of compositional patterns (Figure \ref{fig:aspera-session} AEP, lines 8 \& 15, 18 - 20); (3) flow control and iteration (lines 21 - 24) and; (4) complex date-time reasoning (l. 18 - 20). Moreover, by prompting the LLM with exceptions, the AEPs model disambiguation (lines 9 - 12, 27 - 28) and unsatisfiable requests (lines 25 - 26). The AEP examples contain \textit{planning steps}, that outline a possible decomposition of the task (lines 7, 14, 17) to encourage step-by-step thinking and to improve generation quality.

\textbf{Bias mitigation} Since \aspera{} relies on LLMs for data generation, biases inherent to these models may propagate into the generated datasets, potentially limiting task diversity. To address this issue, we employ three techniques. First, inspired by \citet{imaginarium}, we append the history of previously generated queries to the AEP generation prompt, explicitly instructing the model to create novel tasks and thereby reduce repetition bias. Second, we condition task generation on interactively specified \textit{focus instructions}, which developers can use to control query attributes such as complexity, length, or scenario context\footnote{For instance, a focus instruction might be: \textit{"Generate queries requiring coordination of schedules among organisation members."}}. This interactive specification allows direct influence on data diversity (see \Cref{fig:focus-instr} in \Cref{appendix:aep_gen}). Third, queries can be interactively filtered post generation, providing an additional human-driven mechanism to mitigate biases before dataset finalisation (\S\ref{sec:aspera-interactive}).

\subsubsection{SIP generation}
\label{sec:SIP}
After AEP generation, the LLM is prompted (see \Cref{appendix:sip-gen}) to generate an SIP, which initialises the simulation so that the outcomes of an agent's actions can be evaluated. The SIP re-uses the primitives implemented for action execution (Figure \ref{fig:aspera-session} SIP, lines 13 - 22). This obviates the need for handcrafting databases, using templates to define the user query or prompting the LLM with database schemata. While statically defined databases model a single user's behaviour, \aspera{}'s dynamic database generation allows it to model multiple users. To simplify generation of complex environment states (e.g., an organisation reporting structure) the LLM can call \asperaAssistantCodebase{} simulation tools (lines 8 - 10; see \Cref{appendix:eval-tools}).
\begin{table*}[t]
  \centering
  \resizebox{\textwidth}{!}{ 
  \begin{tabular}{cp{15cm}cccc}
    \toprule
    \textbf{Id} & \textbf{Query} & \textbf{Length (words)} & \textbf{Cyclomatic complexity} & \textbf{\# primitives} & \textbf{Max. AST depth} \\
    \midrule
     1 & Assistant, schedule lunch with my entire team tomorrow at noon. & 12 & 1 & 7 & 6\\
     2 & Assistant, schedule lunch with a different team member each day next week at 12:30 PM. & 17 & 3 & 8 & 10 \\
     3 & Assistant, add a 1-hr strategy review with the CFO and the COO one week from today at 2:30. & 23 & 5 & 13 & 9 \\
     4 & Assistant, check my boss' calendar Wednesday to Friday next week, can they meet? & 18 & 7 & 6 & 11 \\
     5 & Assistant, I need to know which of Bill or Bob is busiest next week so I can allocate work. & 21 & 7 & 7 & 14 \\
     6 & Assistant, reorganise my diary on the fifth so that the important meetings come first. & 16 & 9 & 10 & 11\\
     7 & Assistant, cancel the second meeting with Alice tomorrow if she declined. & 13 & 8 & 5 & 10 \\
     8 & Assistant, when in August when everyone from finance is off? & 12 & 10 &  7 & 11\\
     9 & Assistant, set up a status update meeting with my manager every last Friday of the month at 2 PM till the end of the year. Skip his holidays. & 33 & 10 & 16 & 10 \\
     10 & Assistant, edit the attendee list for our fortnightly team planning on Wednesdays at 1 PM to remove Jack and Amy and add the newest sales hire. & 28 & 13 & 11 & 10 \\
    \bottomrule
  \end{tabular}
  }
  \caption{\label{tab:query-sample} \asperaDataset{} sample queries (see \S \ref{sec:dataset}).}
\end{table*}
\begin{figure*}[t]
  \centering
  \includegraphics[width=0.24\linewidth]{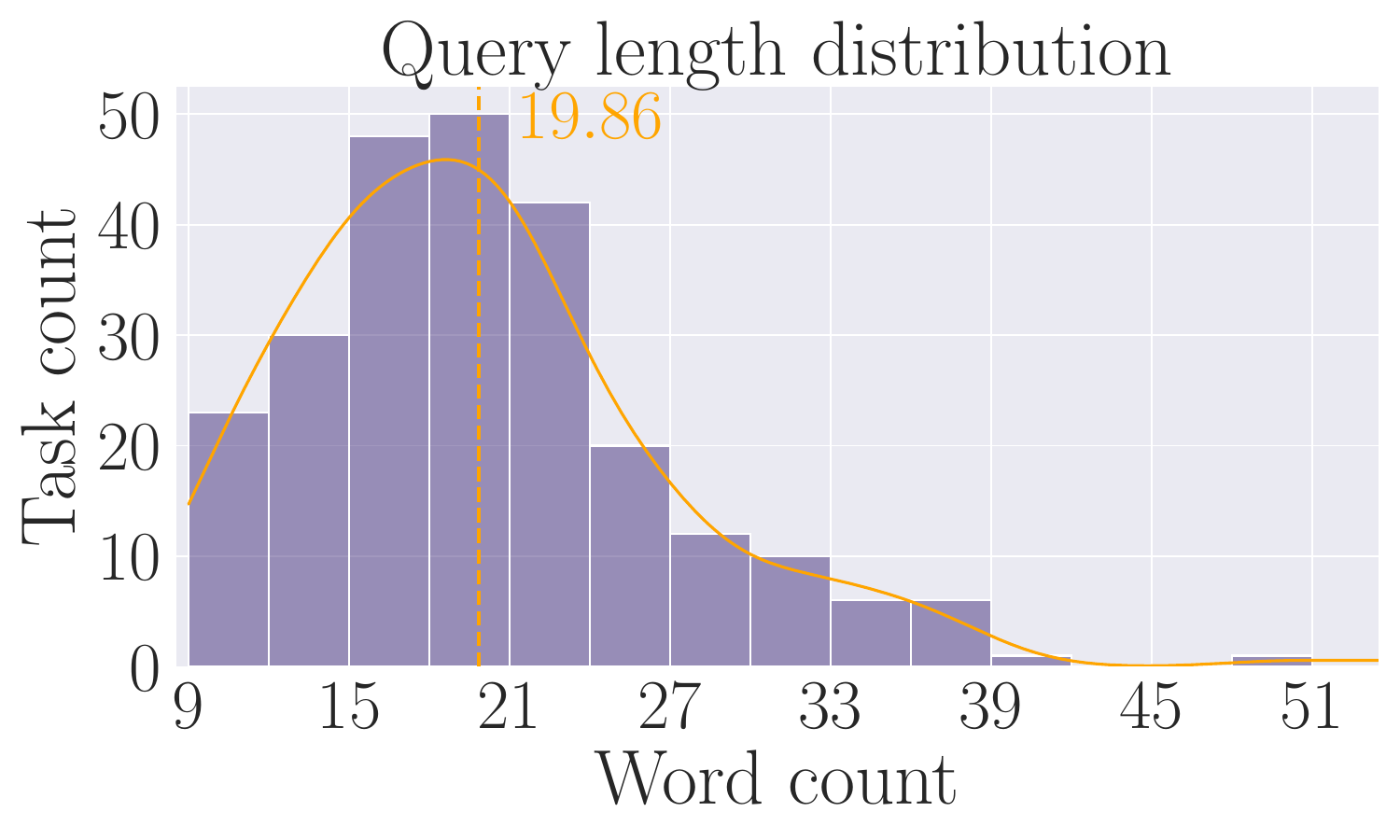} \hfill
  \includegraphics[width=0.24\linewidth]{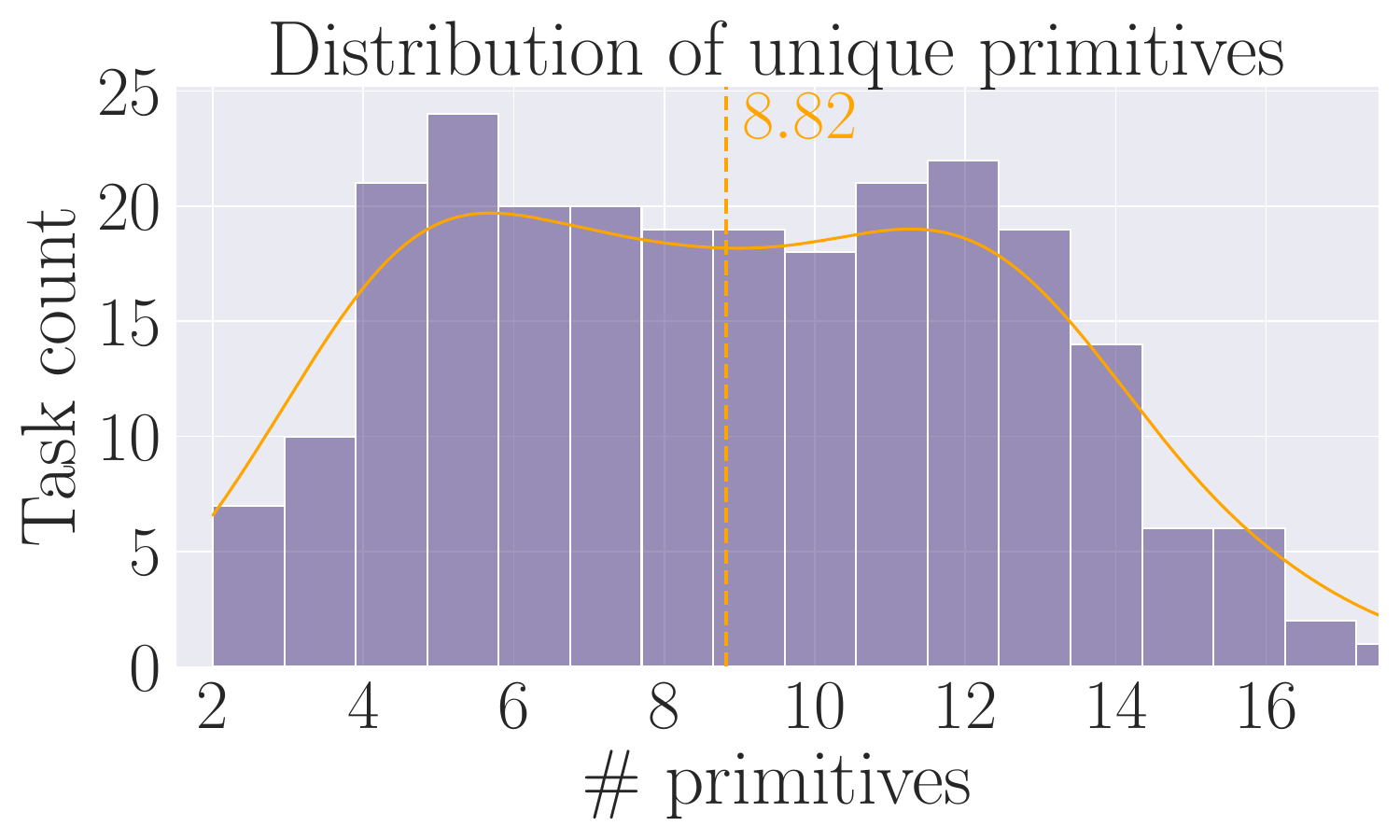} \hfill
  \includegraphics[width=0.24\linewidth]{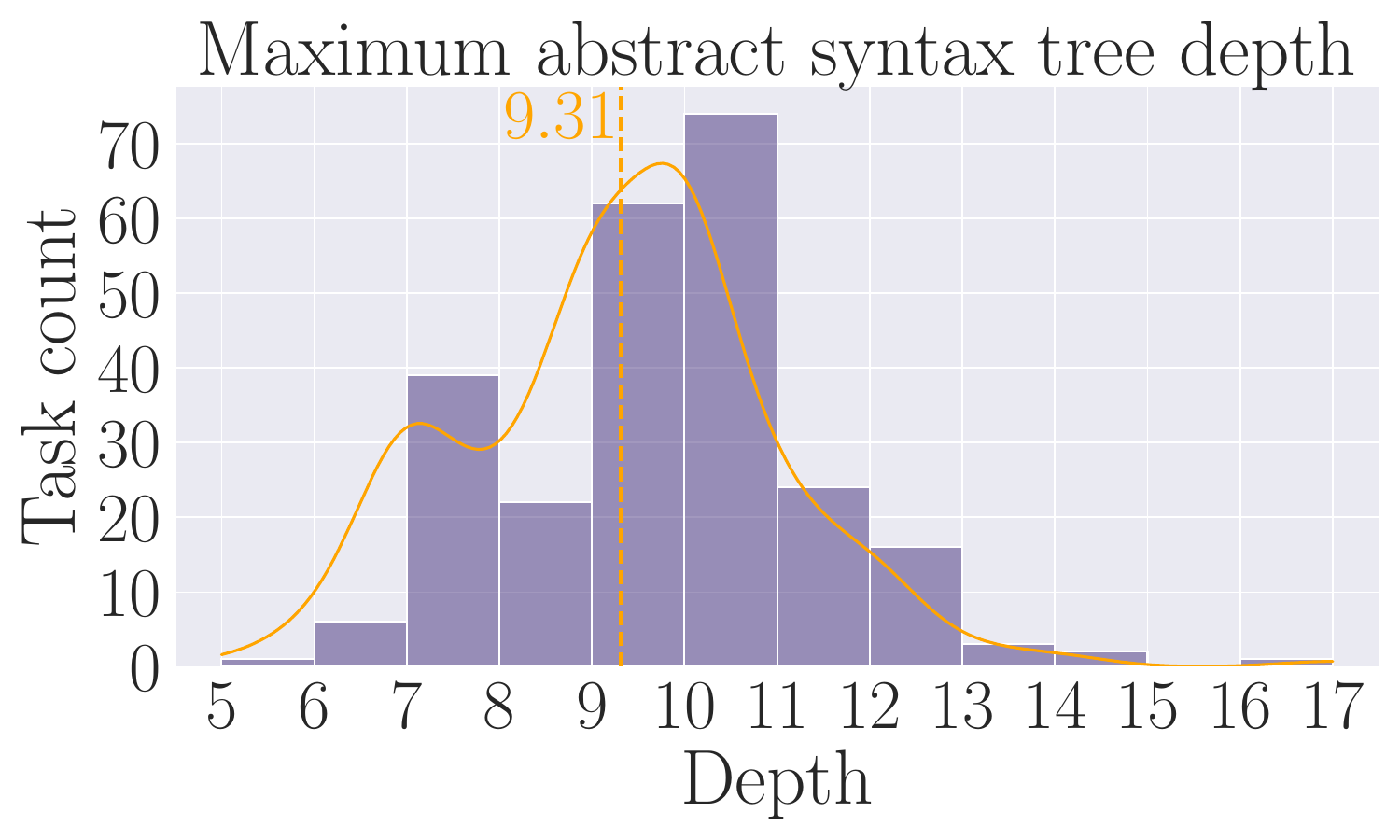}
  \hfill
  \includegraphics[width=0.24\linewidth]{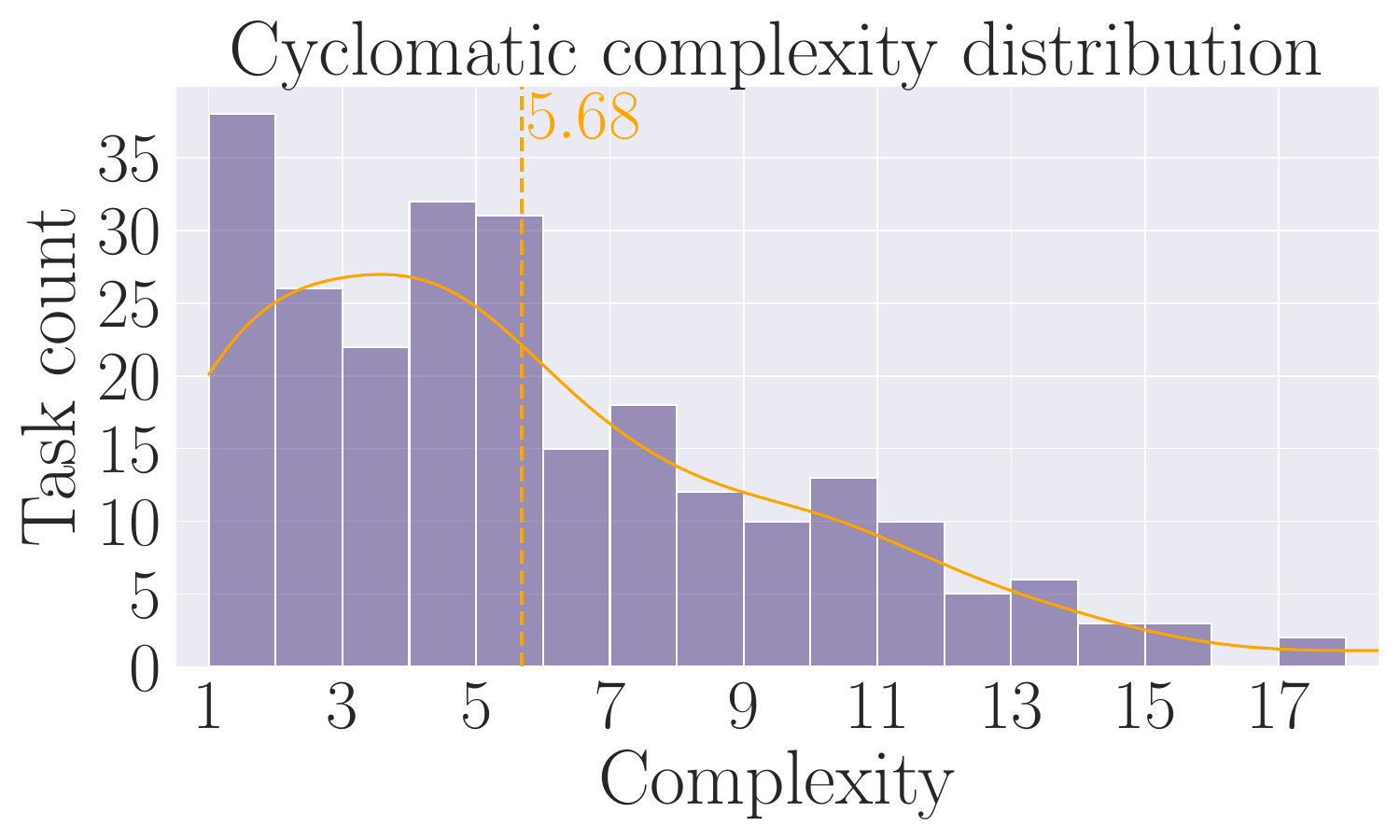}
  \caption{Distributions of key complexity measures in the \asperaDataset{} reference AEPs.}
  \label{fig:dataset_plots}
\vspace{-15pt}
\end{figure*}

\subsubsection{EP generation}
\label{sec:EP}
The final step is to generate an EP, which enables ASPERA to evaluate the functional correctness of an AEP (see \Cref{appendix:ep-gen}). The EP takes as positional arguments the reference SIP and the AEP  (\cref{fig:aspera-session} EP, lines 2 - 4) and executes them in this order (lines 11 \& 17) to initialise the environment and execute the user action. Prior to action execution, one or more variables (line 14) store the initial state relevant to assessing side-effects and user goal completion. After AEP execution, the variables are compared with their expected values in assertion statements (lines 22 - 26). These verify the user goal was met without unexpected side effects. 

The EPs thus implement goal-oriented evaluation even though the environment state is implicit in the queries and SIPs. They generalise database comparison functions implemented in other environments \cite{agentsanbox, workbench} because they can evaluate information-seeking queries by comparing the AEP returned value against its expected value. Finally, evaluation of queries with multiple allowable outcomes\footnote{Multiple outcomes are defined for \textit{When is Bob free next Friday?} since both the upcoming Friday or Friday the following week are valid interpretations of the date mentioned.} is supported in \aspera{} by comparing captured state with a range of accepted values in assertion bodies.

\subsection{Developer-LLM interaction in \aspera{}}
\label{sec:aspera-interactive}
\cref{fig:aspera-session} shows how AEP, SIP, and EP generation is sequential and moderated by a developer. As discussed in \S \ref{sec:query-gen}, the developer can seed the AEP generation with a focus instruction (top left) to ensure diversity or author the query and supervise AEP generation (bottom left). AEPs are generated as \texttt{python} scripts. Developers can add special directives above function signatures to filter low-quality or repetitive examples.

After AEP generation, the chat history is automatically extended with the SIP generation prompt. The developer can optionally instruct the LLM to customise the environment state to be generated, define multiple SIPs or implement new simulation tools the LLM can use to write the SIPs. The interactive loop is repeated to enable EP generation.
At any point, the developer can execute the programs in the simulated environment and edit them (or the queries) accordingly to ensure data quality.
\section{The \asperaDataset{} Dataset}
\label{sec:dataset}

We generate an evaluation dataset of $250$ tasks using GPT-4o\footnote{\texttt{gpt-4o-2024-05-13.}}, given five ICEs for each program type (\S \ref{sec:aspera-task}). $71$ tasks are information-seeking, while the remainder mutate one or more databases. We include both LLM- and human-authored queries. A single SIP and EP are generated for each query, except for conditional queries (Table \ref{tab:query-sample}, line 7) where state initialisation and evaluation are defined to test each AEP branch. Our annotations contain $9$k, $13$k and $17.5$k lines across AEPs, SIPs, and EPs.

\asperaDataset{} AEPs are diverse in their complexity (Figure \ref{fig:dataset_plots}). The distribution of maximum abstract syntax tree (AST) depth indicates AEPs satisfying the queries require compositional use of multiple primitives\footnote{For comparison, the maximum AST depth of an AEP containing a call where all slot values are strings (e.g., \texttt{find\_events(subject="Paper Review")} is $5$.)}; LLMs must interpret extensive documentation across multiple modules and demonstrate strong coding ability to generate AEPs which complete \asperaDataset{} tasks. 

As further shown in \Cref{appendix:dataset-characterisat}, the queries pose challenges ranging from parsing complex time expressions and date/time arithmetic (Table \ref{tab:query-sample}, rows 3, 8 - 10 ) to logical reasoning and interpretation of additional instructions (rows 3 - 5, see \Cref{appendix:challenging-tasks}). Hence, the dataset's diversity stems from task complexity, not paraphrasing.. Representing such complex queries as programs requires iteration and flow-control patterns. This increases a program's \textit{cyclomatic complexity} (CC), defined as the number of independent paths that can be traversed during execution \cite{mccabe1976complexity}. Tasks with higher CC involve non-trivial operations to resolve people, events or dates (Table \ref{sec:dataset}, rows 8, 10), complex rescheduling (row 6) and scheduling events subject to constraints (row 9). Lower CC tasks test fine-grained documentation understanding and programming ability (row 1); occasionally, these tasks require branching to follow instructions which provide relevant information about the environment that does not naturally fit in the documentation (row 3) or describe the assistant policy\footnote{For details, see \cref{fig:guidelines-generation} in \Cref{appendix:aep_gen}.} (\Cref{appendix:policy-prompt}).

\textbf{Quality control} The \aspera{} data generation engine is integrated with the developer's IDE. Consequently, the lead author, who has deep expertise in digital assistants, executed the tasks and used syntax highlighting and auto-completion features to efficiently correct LLM output. Two annotators with similar expertise confirmed the data quality while carrying out the error analysis in \S \ref{sec:analysis}.

\section{\aspera{} Evaluator}
\label{sec:evaluation}
\aspera{} provides an interface which enables arbitrary agents to execute AEPs and observe execution outcome. To support ongoing comparison of the baseline complex action execution capability of LLMs independent of the agent prompt, we provide two implementations of this interface.

\textbf{1. Complete codebase knowledge (CCK)} The agent prompt (\cref{fig:apera-eval-prompts}, \Cref{appendix:agent-prompts}) contains the documentation for the entire assistant library (Table \ref{tab:statistics}) alongside the five AEP example used to generate \asperaDataset{}. The prompt also includes instructions for: an events scheduling policy; information about environment constraints\footnote{These include e.g. company information (e.g., \textit{The leadership team is formed of a CEO, COO and CFO}.).}; and the output format. For information-seeking queries, the type of the object to be returned to the caller is also included in the prompt.

\textbf{2. Primitives selection (PS)} The primitives are not known when the user invokes the assistant. Including the entire assistant library documentation in the prompt (as in the CCK prompt) may be impractical due to context window and latency limitations. In such a case, the assistant must inspect the library to determine which primitives are needed to execute the action requested by the user. To evaluate how well agents perform under these constraints, we provide a simple interface in which AEP generation is conditioned on primitives selected by the LLM prior to generation. This involves an iteration through an extended assistant library\footnote{The extension contains documentation for the \texttt{ai\_assistant}, \texttt{contacts}, \texttt{files}, \texttt{messaging}, \texttt{navigation}, \texttt{user\_device\_settings} modules in addition to those reported in \cref{tab:statistics}, to be implemented in a future release.}. At each step, the agent is prompted with the documentation for an \asperaAssistantCodebase{} module (viz \cref{tab:statistics}) alongside the user request and is asked to issue \verb|import| statements to select relevant primitives or \verb|None| if the module is not relevant for executing the requested action (Figure \ref{fig:ps-prompt-template}, \Cref{appendix:agent-prompts}). On iteration completion, the selected primitives replace the full application library listings in the CCK prompt. 

Unlike the CCK prompt, which includes 5 ICEs, the PS AEP generation prompt contains only one example demonstrating the solution format. Including the CCK examples would have inflated success rates for agents with poor primitive selection recall, as the ICE primitives and their documentation would appear without being explicitly imported.

\textbf{Metrics} We report task success. A task is completed if the generated AEP executes without error and all assertions pass in all reference EPs.

\section{\asperaDataset{} Evaluation}
\label{sec:results}

\textbf{Complete assistant library knowledge (CCK setting)} AEP generation is challenging for both proprietary and open-source LLMs  even when they can directly observe all the knowledge relevant for planning (\cref{tab:cck-results-main}). Despite performing well on standard code generation benchmarks (\citet{chen2021evaluating}, \citet{mbpp}), and their ability to consistently generate syntactically correct AEPs (\cref{tab:cck-results-main}, column 5), the most widely used general-purpose assistants successfully execute only $45.33\%$ (GPT-4o) and $33.73\%$ (Gemini 1.5 Pro \cite{gemini1.5}) of actions.  Task success correlates with model size (\cref{tab:cck-results-main}, r. 9-13). However, the improved task success of o1-mini compared to larger LLMs such as GPT-4o ($+6.1\%$) and Gemini 1.5 Pro ($+17.67\%$) suggests that scaling inference-time compute may be a key enabler of improved complex task understanding and execution capabilities. 

\begin{table}[t]
\large
\resizebox{\columnwidth}{!}{
\begin{tabular}{l l c c c c}
\toprule
\textbf{Model name} & \textbf{Checkpoint} & \textbf{Size} & \textbf{Task success (\%)} & \textbf{Syntax err. (\%)}\\
\midrule
o1 & o1-preview-2024-09-12 & - & 80.13 & -  \\
o1-mini & o1-mini-2024-09-12 & - &  51.40 & 0.13 \\
GPT-4o & gpt-4o-2024-05-13 & - &  45.33 & - \\
GPT-4o-mini & gpt-4o-mini-2024-07-18 & - &  21.07 & -\\
3.5-turbo & gpt-3.5-turbo-0125 & - &  10.80 & 1.20 \\
\midrule
1.5-pro & gemini-1.5-pro-002 & - &  33.73 & 0.40 \\
1.5-flash & gemini-1.5-flash-002 & - &  27.87 & 0.40 \\
1.0-pro & gemini-1.0-pro-002 & - &  12.67  & 0.53 \\
\midrule
Mistral L & Mistral-Large-Instruct-2407 & 123B &  38.00 & - \\
Qwen2.5 & Qwen2.5-72B-Instruct & 72B &  28.80 & - \\
Gemma2 & gemma-2-27b-it & 27B &  14.40 & 0.4 \\
CodeGemma & codegemma-7b-it & 7B &  2.40 & 6.0 \\
\bottomrule
\end{tabular}
}
\caption{CCK \asperaDataset{} task completion rates (5-shot). Proprietary model results are averaged over three runs. Greedy decoding is used for all models except o1, which only supports a temperature of 1. See \Cref{appendix:extended-evals} for further evaluations.}
\label{tab:cck-results-main}
\vspace{-10pt}
\end{table}

\begin{table}[t]
  \centering
  \resizebox{\columnwidth}{!}{
    \begin{tabular}{cccccccc}
      \toprule
      \textbf{Model name} & \textbf{Setting} & \textbf{\# ICE} & \textbf{Micro F1} & \textbf{P} & \textbf{R} & \textbf{Task success (\%)} \\
      \midrule
      \multirow{3}{*}{o1} & CCK & 5 & - & - & - & 80.13 \\
       & CCK & 1 & - & - & - & 72.80 \\
       & PS & 1 & 0.63 & 0.60 & 0.67 & 28.40 \\
      \midrule
      \multirow{3}{*}{GPT-4o} & CCK & 5 & - & - & - & 45.33 \\
       & CCK & 1 & - & - & - & 36.53 \\
       & PS & 1 & 0.56 & 0.56 & 0.55 & 11.46 \\
      \bottomrule
    \end{tabular}
  }
  \caption{PS task success. Rows 1 and  4  are repeated from Table \ref{tab:cck-results-main}, \# ICE denotes the number of AEP examples in the prompt. Precision and recall are computed with respect to the \asperaDataset{} reference AEPs.}  
  \label{tab:ps-results-main}
  \vspace{-15pt}
\end{table}
\begin{table}[t]
    \centering
    \large
    \resizebox{\columnwidth}{!}{
    \begin{tabular}{lccccc}
        \toprule
         \multirow{2}{*}{\textbf{Statistic}} & \multicolumn{5}{c}{\textbf{Model name}} \\
        \cline{2-6}
        & \textbf{GPT-3.5-turbo} & \textbf{GPT-4o-mini} & \textbf{GPT-4o} & \textbf{o1-mini} & \textbf{o1} \\
        \midrule
        Lines of code $\Delta$ to reference AEPs & -12.15 & -7.3 & -5.48 & 3.22 & 8.72 \\ 
         \texttt{RequiresUserInput} usages & 52 & 93 & 170 & 360 & 291 \\
        Average planning steps (viz. Figure \ref{fig:plan}, lines 2\&9) & 4.83 & 5.63 & 5.41 & 6.15 & 9.16 \\
        Helper functions count & 0 & 2 & 11 & 29 & 65 \\
        Average cyclomatic complexity & 2.92 & 3.82 & 4.44 & 5.95 & 6.80 \\
        \bottomrule
    \end{tabular}
    }
    \caption{Key generated AEPs statistics}
    \label{tab:aep_statistics}
\end{table}
\begin{table}[t]
  \centering
  \large
  \resizebox{\columnwidth}{!}{
  \begin{tabular}{lcccc}
    \toprule
    \textbf{Model name} & \textbf{Programs debugged} & \textbf{Programs analysed} & \textbf{Errors labelled} & \textbf{Could recover (\%)} \\
    \midrule
    GPT-4o & 33 & 125 & 41 & 48.39 \\
    GPT-3.5-turbo & 66 & 125 & 100  & 24.62 \\
    \bottomrule
  \end{tabular}
  }
  \caption{\label{tab:execution-err-analysis} Execution error analysis statistics.}
\end{table}
\textbf{Primitive selection (PS setting)} Despite its AEP generation capability when conditioned on the documentation of the entire \asperaAssistantCodebase{} library, o1 retrieves just $67\%$ of the primitives relevant for AEP implementation and achieves a modest $28.4\%$ task completion rate as a result (\cref{tab:ps-results-main}). Hence,  while identifying which primitives are relevant for executing a given action is relatively simple for human developers, we find that SOTA LLMs have limited ability to perform in this setting. 

\section{Analysis and discussion}
\label{sec:analysis}
\subsection{CCK error analysis}
\begin{figure}[t!]
    \centering
    \includegraphics[width=\linewidth]{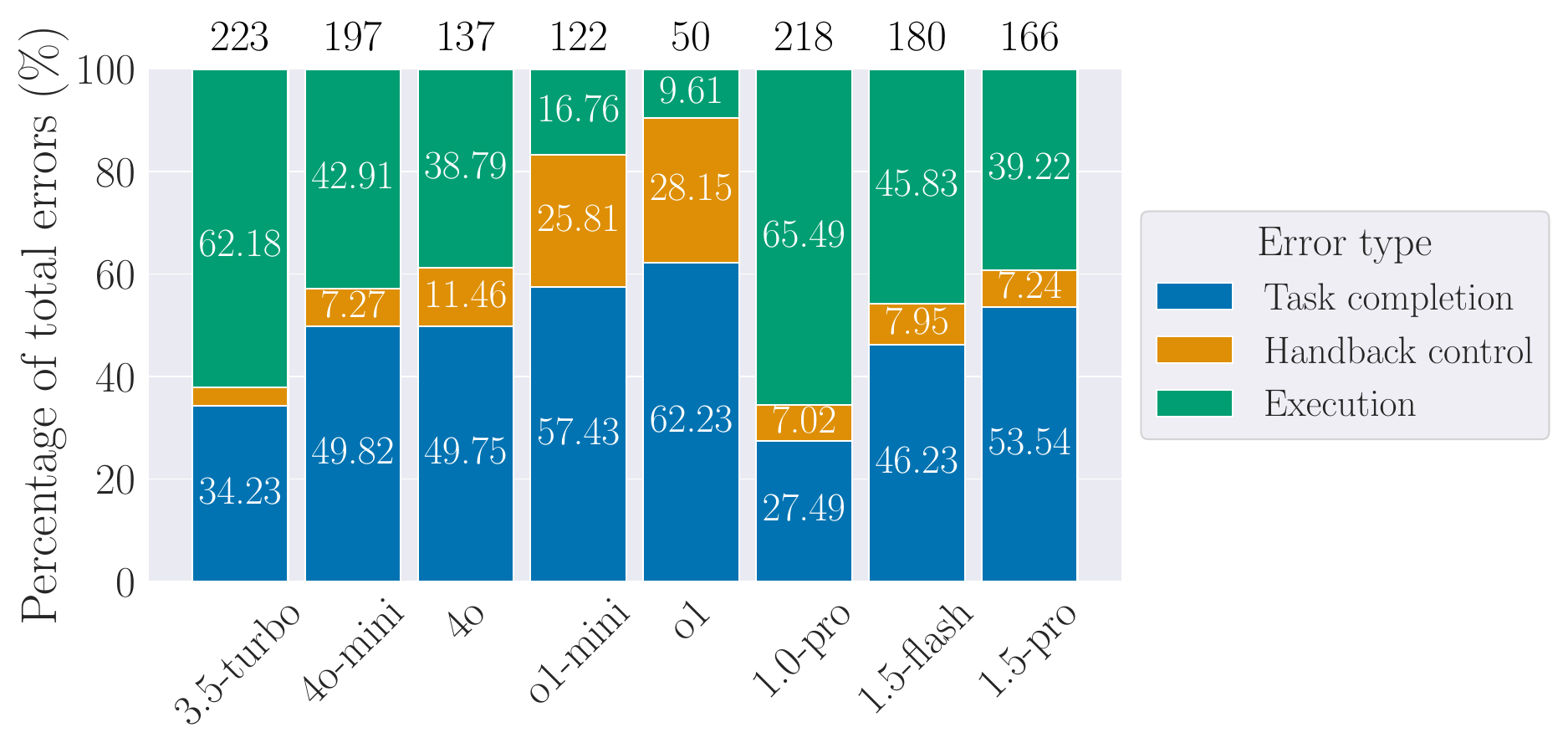}
    \caption{Assistant error types for OpenAI and Gemini model families. Top row displays total error counts.}
    \label{fig:error_breakdown}
\vspace{-15pt}
\end{figure}

We begin with an in-depth analysis of programs generated by agents prompted with the documentation of the entire \asperaAssistantCodebase{} library. A breakdown of the errors observed is presented in Figure \ref{fig:error_breakdown}. We make three key observations.

First, for both OpenAI and Gemini models, more capable variants produce a larger proportion of \textit{task completion errors}, in which programs execute successfully but fail an assertion in evaluation. Such an error indicates that the model can successfully use and combine primitives, but fails to understand some nuance in the user request and therefore takes the wrong action. Table \ref{tab:3.5-short-solutions} (\Cref{appendix:task-completion-errors}) shows concrete examples of this.

Second, less capable models incur relatively more \textit{execution errors}, in which programs are syntactically correct but trigger a runtime exception. An in-depth error analysis of $141$ such errors from GPT-3.5-turbo and GPT-4o (\Cref{appendix:execution-err-analysis}) shows that both models have a tendency to hallucinate in situations where multi-step reasoning is required, generating shorter AEPs compared to the reference annotations (\cref{tab:aep_statistics}, row 1). Additionally, we find that execution errors often occur with task completion errors; in other words, the solution is incorrect even if the execution error is manually fixed (\cref{tab:execution-err-analysis}, column 5). While self-reflecting agents \cite{reflection} could achieve higher task success, our evaluation considers complex action execution in the single trial setting since, in practice, self-debugging iterations increase latency and trial execution might have unintended consequences. 

\begin{table}[t]
\centering
\large
\resizebox{\columnwidth}{!}{
\begin{tabular}{lcccccc}
    \toprule
     \textbf{Subset} & \textbf{CC} & \textbf{AST depth} &  \textbf{o1(\%)} & \textbf{GPT-4o (\%)} & \textbf{Example} \\
    \midrule
    Simple & 1.9 & 7.3 & 100 & 100 & Table \ref{tab:query-sample}, row 1 \\
    Constrained scheduling & 7.1 & 9.6 & 86.67 & 46.67 & Table \ref{tab:query-sample}, row 9 \\
    Complex time expressions & 5.4 & 9.2 & 63.33 & 20.00 & Table \ref{tab:query-sample}, r. 4 \& 10 \\
    Policy / instruction following & 6.0 & 9.2 & 80.00 & 20.00 &  Table \ref{tab:query-sample}, r. 2 \& 3 \\ 
    Advanced problem solving & 9.2 & 10.6 & 56.67 & 26.67&  Table \ref{tab:query-sample}, row 5 \\ 
    \bottomrule
\end{tabular}
}
\caption{\label{tab:subset-complexity} Task success for query subsets. Each subset has $10$ queries, see  \Cref{appendix:problem-categories} for complete listings.}
\vspace{-15pt}
\end{table}
Third, more capable models generate a greater proportion of \textit{handback control errors}. These errors are linked to more frequent use of the \texttt{RequiresUserInput} exception (\cref{tab:aep_statistics}, row 2), used to handle cases in which the assistant cannot complete a task or cannot disambiguate between some entities at runtime. The errors occur when this exception triggers unexpectedly, indicating that the agent has made an incorrect assumption or misidentified an edge case. These errors illustrate which types of queries remain difficult for SOTA models (see \cref{tab:o1-handback-control}, \Cref{appendix:handback-control-errors}).

\vspace{-5pt}
\begin{table*}[h!]
  \centering
  \tiny
  \begin{subtable}{\textwidth}
    \centering
    \resizebox{\textwidth}{!}{
      \begin{tabular}{>{\raggedright}l  m{7cm}  m{8cm}}
        \toprule
        \textbf{Id} & \textbf{Query} & \hspace{0.75cm}\textbf{Error Snippet} \\
        \midrule
        1 & Assistant, schedule our team Christmas party 10 days before Christmas. Should start in the morning and end at 10 PM. &
        \scalebox{1.0}{
          \includegraphics[width=0.45\textwidth, trim=0cm 11cm 1cm 11cm, clip]{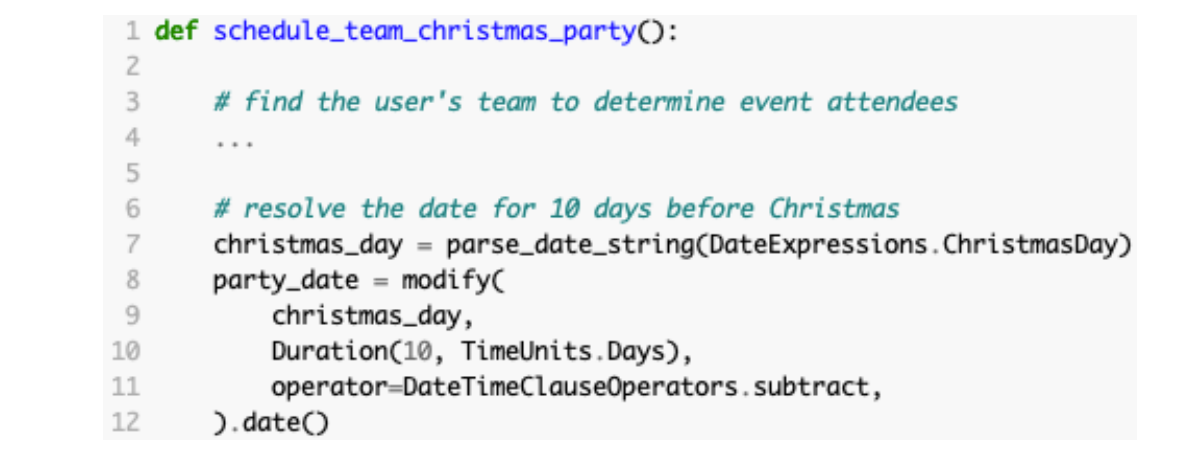}} \\
        \multicolumn{3}{p{\textwidth}}{\textbf{Tool use (datetime):} Line \texttt{9} contains a \texttt{TypeError}, \texttt{modify} only accepts \texttt{datetime} objects. A correct solution requires an additional reasoning step: pass \texttt{christmas\_day} and one of the specified times to the \texttt{combine} library function to get the correct type.} \\
        \midrule
        2 & Assistant, set up a training session for all employees from the Engineering team next Monday from 2 PM to 5 PM. Send out invites and book a conference room that fits 20 people. &
          \includegraphics[width=0.45\textwidth, trim=2.5cm 12.5cm 1cm 12.5cm, clip]{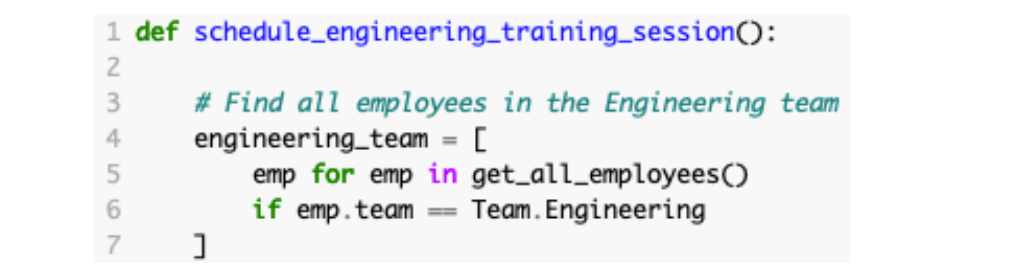} \\
        \multicolumn{3}{p{\textwidth}}{\textbf{Attribute hallucination:} In line \texttt{6}, the \texttt{.team} attribute access raises an error because the \texttt{Employee} objects returned by \texttt{get\_all\_employees} only have \texttt{name} as an attribute. The \texttt{Employee} object should be passed instead to the \texttt{get\_employee\_profile} library function to return an object which has \texttt{team} as an attribute.} \\
        \midrule
        3 & Assistant, put 45 minutes in the calendar, back-to-back, with Engineering and Marketing starting at 10 AM tomorrow... Actually, add a 10-minute buffer between each meeting. &
          \includegraphics[width=0.45\textwidth, trim=0.25cm 12.5cm 1cm 12cm, clip]{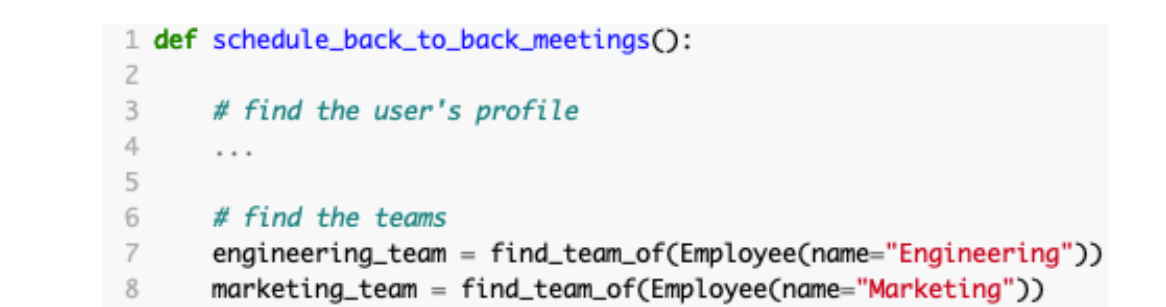} \\
        \multicolumn{3}{p{\textwidth}}{\textbf{No tool use (lazy solution):} The assistant hallucinates lines \texttt{7-8} instead of using relevant APIs to find the engineering team, in spite of documentation that states that \texttt{Employee} objects cannot be instantiated. The functions \texttt{get\_all\_employees}, \texttt{get\_employee\_profile} and the enumeration \texttt{Team.Engineering} should have been composed, similar to snippet in row 2.} \\
        \bottomrule
      \end{tabular}
    }
    \caption{Execution errors examples.}
    \label{tab:execution-errors-main}
  \end{subtable}


  \begin{subtable}{\textwidth}
    \centering
    \resizebox{\textwidth}{!}{
      \begin{tabular}{p{0.1cm}  p{8.5cm}  p{6.5cm}}
        \toprule
        \textbf{Id} & \textbf{Query} & \textbf{Agent action} \\
        \midrule
        1 & Assistant, Ari and James are on holiday next month, who's out for longer? & Sums duration of all vacations, month notwithstanding. \\[0.5ex]
        \multirow{2}{*}{2} & Assistant, add bi-weekly mentorship sessions with the reports of my reports starting next Monday at 2 PM to my calendar. & \multirow{2}{*}{Hallucinates an end date for the  event, scheduling instances only for 6 months.} \\[0.5ex]
        \multirow{2}{*}{3} & Assistant, reschedule the meetings which overlap with the annual review this afternoon to the same time tomorrow. & Creates copies of overlapping events tomorrow, instead of modifying existing events. \\
        \bottomrule
      \end{tabular}
    }
    \caption{Task completion errors examples.}
    \label{tab:task-completion-errors-main}
  \end{subtable}


  \begin{subtable}{\textwidth}
    \centering
    \resizebox{\textwidth}{!}{
      \begin{tabular}{l  p{10cm}  p{6cm}}
        \toprule
        \textbf{Id} & \textbf{Query} & \textbf{Agent action} \\
        \midrule
        \addlinespace[1.0ex]
        \multirow{2}{*}{1} & \multirow{2}{*}{Assistant, find a suitable conference room for a meeting with my team I wanna schedule later today.} & Tries to schedule a meeting, handing back control because of incorrect diary checking. \\
        \addlinespace[0.5ex]
        \multicolumn{3}{p{\textwidth}}{\textbf{Error cause:} Distracted by irrelevant information. The agent is not required to schedule a meeting -- not enough details are provided. It should search for an available room that has sufficient capacity to accommodate the user and their team instead.} \\
        \midrule
        \addlinespace[1.0ex]
        \multirow{2}{*}{2} & Assistant, can you find a time slot in my diary today when I could schedule something with the HR department to discuss my performance review? & Hallucinates a program attempting to find HR team, handing back control because it cannot determine it. \\
        \addlinespace[0.5ex]
        \multicolumn{3}{p{\textwidth}}{\textbf{Error cause:} Distracted by irrelevant information. The HR team is not defined in the simulation. The task requires the agent to find a slot in user's diary.} \\
        \midrule
        \addlinespace[1.0ex]
        3 & Assistant, schedule our team Christmas party 10 days before Christmas. Should start in the morning and end at 10 PM? & Requires the user to provide an alternative date. \\
        \addlinespace[0.5ex]
        \multicolumn{3}{p{\textwidth}}{\textbf{Error cause:} Following policy. The agent follows the instruction \textit{Unless the user explicitly states the date, meetings should not be scheduled on or recur during weekends.} which is irrelevant.} \\
        \bottomrule
      \end{tabular}
    }
    \caption{Handback control errors examples.}
    \label{tab:handback-control-errors-main}
  \end{subtable}
  \caption{Error examples (CCK setting). Further examples are provided in \Cref{tab:error_snippets,tab:3.5-short-solutions,tab:o1-handback-control} in \Cref{appendix:extended-results}.}
  \label{tab:error-examples-main}
  \vspace{-15pt}
\end{table*}

\subsection{Case study}
\Cref{tab:error-examples-main} presents examples of the error categories introduced previously, highlighting the critical need for evaluation protocols that reflect real-world deployment conditions. While execution errors may be recoverable through repeated trials, task-completion and handback control errors are often irreversible, potentially resulting in misinformation to end-users. Crucially, as shown in \Cref{fig:error_breakdown}, these irreversible errors predominate in frontier models. Thus, reductions in these error types provide compelling evidence of enhanced reasoning and reliability in LLMs, underscoring the importance of targeted improvements in model evaluation.
\begin{figure}[t]
  \centering
  \begin{subfigure}[b]{0.8\linewidth}
    \centering
    \includegraphics[width=\linewidth]{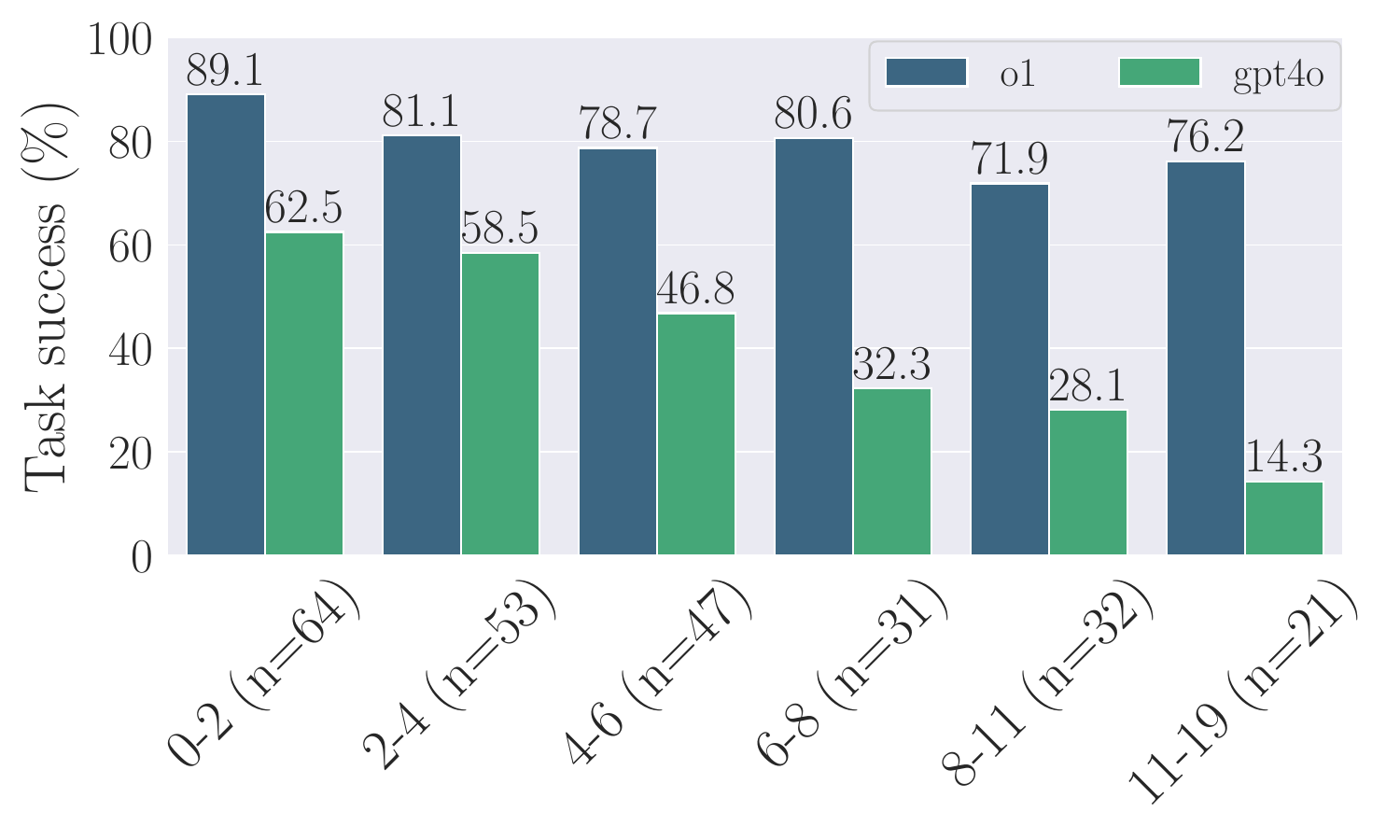}
    \caption{Cyclomatic complexity}
    \label{fig:accuracy_vs_cc}
  \end{subfigure}
  \vspace{-5pt}
  \begin{subfigure}[b]{0.8\linewidth}
    \centering
    \includegraphics[width=\linewidth]{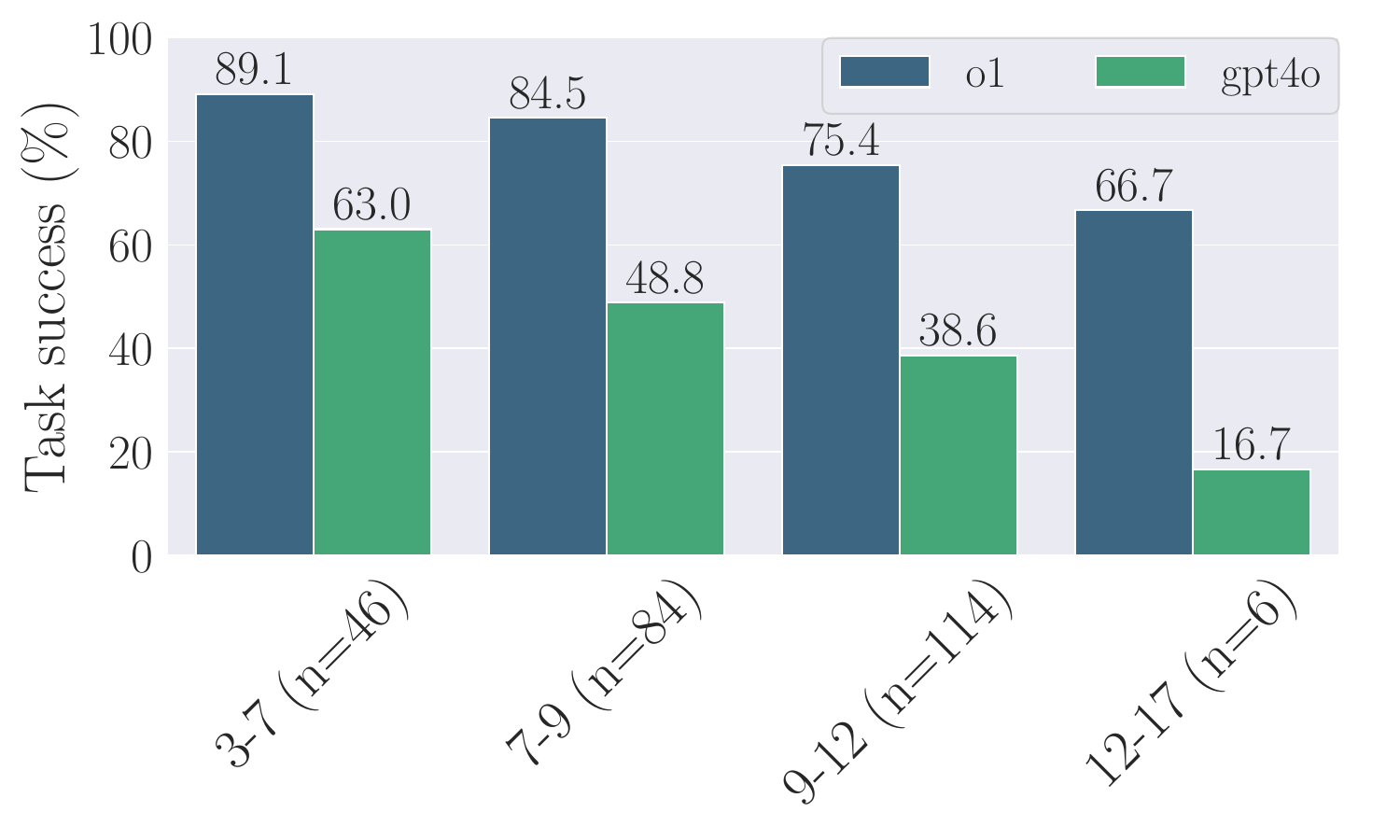}
    \caption{Maximum AST depth}
    \label{fig:accuracy_vs_ast_depth}
  \end{subfigure}
  \caption{Task success as a function of reference AEP complexity ($n$ denotes bucket size).}
  \label{fig:task-success-complexity}
  \vspace{-15pt}
\end{figure}
\subsection{Handling complexity}

\asperaDataset{} requires models to perform various complex compositions of primitives and control flow sequences. \cref{fig:task-success-complexity} shows that o1 can successfully complete a much larger proportion of tasks which require generating complex programs compared to GPT-4o. As seen in Table \ref{tab:aep_statistics}, o1 is more capable in this regard due to its ability to break down the task into fine-grained steps (row 3), make use of helper functions to encapsulate complex functionality (row 4) and to more effectively employ flow control and iteration (row 5). 

To further demonstrate the challenges in \asperaDataset{}, we select $5$ subsets of $10$ queries which test different aspects of assistant understanding and reasoning capabilities. Table \ref{tab:subset-complexity} (row 1) shows that both o1 and GPT-4o can equally handle simple problems (e.g., scheduling an event on a given date, or deleting events) but a large gap is observed in the completion rate of advanced tasks. Compared to o1, GPT-4o struggles with constrained scheduling and resolving challenging relative time expressions (rows 2 \& 3), which require flow control, primitive composition and arithmetic reasoning. The same is true of generating AEPs constrained by additional instructions in the prompt (row 4) and solving very challenging examples from the above categories (row 5).

\subsection{Primitives selection}

The primitives selection setting proved challenging for both models evaluated, as shown in \cref{tab:ps-results-main}.
The LLMs show limited ability to reason about dependent primitives. Using the \verb|work_calendar| module, for example, requires knowledge about properties of the \verb|Event| primitive. We find this relation is not recognised during selection; o1 fails to import both the relevant \texttt{work\_calendar} API and \texttt{Event} in 29 out of 67 occurrences of \verb|find_events|, 16 out of 69 occurrences of \verb|add_event| and 8 out of 19 occurrences of \verb|delete_event|.

The ability of an LLM to use a primitive listed in the prompt is weakly associated with its selection performance for that primitive. In our baseline setting (CCK, 1-shot), o1 achieves a $66\%$ task success rate on queries where the reference AEP uses \verb|add_event|. However, its recall for selecting \verb|add_event| is just $0.41$, with an F1 score of $0.58$ (see \Cref{appendix:primitive-selection-expanded-results}). This suggests that selecting a complete set of fine-grained primitives to execute complex user requests is challenging for LLMs.
\section{Related Work}
\label{sec:related-work}
\textbf{Task-oriented parsing} Parsing natural language queries into DSL programs interpretable by execution engines \cite[\textit{inter alia}]{zelle1996learning,gupta-etal-2018-semantic-parsing} is challenging for program structures unseen in training  \cite{yao-koller-2022-structural}. \citet{DBLP:conf/naacl/BoginGCS24} and \citet{jhamtani2023natural} show that representing targets as \textit{programming languages} improves LLMs' few-shot semantic parsing ability; we build on this, employing program synthesis to collect complex, high-quality, task-oriented queries and to evaluate agents' ability to understand them.

\textbf{Tool-augmented LLMs} An alternative is query synthesis at scale by prompting LLMs with documentation of sampled synthetic- \cite{toolalpaca} or real-world APIs \cite{DBLP:journals/corr/abs-2305-16504,DBLP:journals/corr/abs-2306-06624,DBLP:conf/iclr/QinLYZYLLCTQZHT24} and query examples. Because the relations between the sampled APIs are sparse, the resulting programs are linear sequences of often unrelated API calls. As such, tool-use corpora mostly evaluate LLMs' ability to parse API call sequences rather than complex reasoning with multiple tools. By grounding queries in a library with primitives sharing type relations, we generate more challenging and natural tasks (see \S \ref{sec:ose}) that require multi-step, arithmetic and logical reasoning, building on \citet{taskbench},  who ground queries in handcrafted task relation graphs.

\textbf{LLM Agents} Synthetic data generation at scale comes with quality \cite{qualitymatters} and evaluation \cite{DBLP:conf/acl/GuoCWLQLL0L24} challenges. To tackle the former, human-authoring and manual curation have been increasingly employed \cite[\textit{inter alia}]{DBLP:conf/iclr/HuangSLFWZ000G024, jhamtani2023natural, berkeley-function-calling-leaderboard}. Instead, we propose an interactive data generation engine to ensure data quality and reduce human cost. Like WorkBebnch \cite{workbench} and AppWorld \cite{appworld} we tackle evaluation challenges by executing agent actions in a simulated environment and determining whether they satisfy the user goal. While both WorkBench and AppWorld template user queries and resort to program templates (WorkBench) or high-fidelity task simulators (AppWorld) to annotate environment state, \aspera{} does not constrain the format of the query or of the program grounding it. Like AppWorld we generalise the strict database comparisons of WorkBench, but generate the evaluation programs in LLM interactions as opposed to manually implementing them for every task.

Web agent benchmarks and \asperaDataset{} differ in action space complexity. The former typically define small action sets; for example, WebArena \cite{webarena} has 12 actions with simple descriptions like \texttt{new\_tab} (\textit{Open a new tab}). In contrast, \asperaDataset{} features 69 primitives, spanning high- (e.g., \texttt{delete\_event}) and low-level ones (e.g., \texttt{now\_}). This richer action space demands assistants reason over fine-grained dependencies and documentation to generate complex programs satisfying user requests. Meanwhile, web agents generate short action sequences, one step at a time, to achieve simpler goals\footnote{Compare \href{https://bit.ly/3CUqK1k}{https://bit.ly/3CUqK1k} with \Cref{fig:plan} and the \asperaDataset{} AEPs in \Cref{appendix:challenging-tasks} and our data release.}. See \S \ref{sec:ose} for an in-depth comparison.

\textbf{Code generation} LLM ability is measured by benchmarks \cite{chen2021evaluating,DBLP:journals/corr/abs-2108-07732,DBLP:conf/nips/HendrycksBKMAGB21} which test algorithmic ability via generation of self-contained functions with contextual dependencies limited to standard libraries. To address this, other resources encompass narrow-domain dependencies on external data-science libraries \cite{DBLP:conf/icml/Lai0WZZZYFWY23,DBLP:conf/emnlp/WangZFN23} or a broader set of domains \cite{bigcodebench}. ASPERA focuses on program generation with project-runnable dependencies \cite{DBLP:conf/icse/YuSRZZMLLWX24} of custom primitives in the assistant library, which is very challenging but receives limited coverage in existing resources \cite{DBLP:journals/corr/abs-2404-10155}. Moreover, ASPERA tasks represent high-level user goals requiring the assistant to reason about primitive relevance, while the aforementioned benchmarks test program generation given precise function specifications and knowledge about external libraries acquired during pre-training. Evaluation robustness is guaranteed by execution of human-authored tests for all the  above benchmarks except \citet{bigcodebench} who, like our work, use human-LLM interaction to generate data and robustly evalute general software task competence.  

\section{Conclusion and Future Work}
\label{sec:conclusion}
This work evaluated the ability of LLMs to parse complex natural language queries into executable programs that involve non-trivial primitive composition and flow control. We have addressed key limitations in existing work regarding dataset availability and evaluation by devising an environment where LLMs and human developers interact to collect evaluation data and code for environment state initialisation and execution outcome verification. We found that generating programs which satisfy intricate user queries grounded in custom assistant libraries is challenging for a wide range of SOTA LLMs which are otherwise proficient at code generation. Our initial results also showed that, while SOTA LLMs can compose primitives to execute complex tasks, they struggle to determine when a specific primitive is required given the query alone which is of concern to practical digital assistants. Hence, \asperaDataset{} and the \aspera{} framework enable future study of action execution in the challenging setting where the primitives are not known to the agent and must be retrieved or discovered via environment interaction. Recently, agent-based approaches have been proposed to address such challenges for software engineering \cite{DBLP:conf/nips/YangJWLYNP24,DBLP:conf/iclr/0001LSXTZPSLSTL25}. Extending these approaches for complex query parsing is a promising direction for future work.

\section{Limitations}

\textbf{Dataset size} \asperaDataset{} is comparable in size to other popular code generation benchmarks such as HumanEval \cite{chen2021evaluating}, NumpyEval  \cite{DBLP:conf/ijcai/ZanCYLKGWCL22} ,  PandasEval  \cite{DBLP:conf/ijcai/ZanCYLKGWCL22} and TorchDataEval \cite{DBLP:conf/emnlp/ZanCLGWL22}, but likely not sufficiently large for fine-tuning LLMs for digital assistant applications. Future work could focus on scaling the size of our data using the \aspera{} data generation engine or by LLM-assisted paraphrasing of existing queries and refactoring of SIPs and EPs, similar to \citet{bigcodebench}. This would enable future work to study robustness of finetuned digital assistant models under non-trivial, semantics-preserving transformations of the assistant library (e.g., refactoring). 

\textbf{Limited domain coverage} The \aspera{} assistant library supports parsing of complex time expressions and a simple simulation of a corporate calendar. Furthermore, the assistant library provides documentation for $6$ domains (see \S \ref{sec:evaluation}, footnote 12). With more time investment, these domains could be simulated, along with any additional simulation and evaluation tools necessary to generate the environment state. The expansion could focus on evaluating requests which span multiple applications (e.g., \textit{How long will it take me to drive to my next meeting this afternoon?}) which are not supported in the current release.

We note that, while the simulation and the current set of evaluation and simulation tools were developed offline by one of the authors with GPT-4o assistance, future releases could explore using LLMs for assisting the developer with auxiliary tool implementation during the \aspera{} interactive session. We anticipate that the human effort required to scale to new domains depends on the LLMs available for data generation, the complexity of the domain considered and the complexity of the scenarios 
developers wish to simulate.

\textbf{Dataset bias} As discussed in Sections \ref{sec:query-gen} and \ref{sec:aspera-interactive}, we mitigate dataset bias during \asperaDataset{} generation through three strategies: incorporating query history into AEP prompts, conditioning generation on focus instructions and filtering repeated or redundant examples. However, these safeguards have limitations. Large language models may not consistently follow instructions, and filtering becomes increasingly challenging as dataset size grows. As a result, the degree of bias in \aspera{}-generated datasets ultimately depends on both the underlying LLM and the extent of human oversight during data curation.

\textbf{Multi-turn interactions} In keeping with works focused on multiple tool use and LLM agents, our work considers a user which issues a complex request in a single-turn interaction. In practice, it is desirable that the digital assistant can handle complex requests at any point in a conversation. Moreover, multi-turn interaction is necessary when the assistant cannot perform entity disambiguation or has failed to solve the task. Future work could exploit the error handling sequences in the reference \asperaDataset{} AEPs to generate dialogues where complex action execution requires user interaction, similar to recent work by \citet{agentsanbox}.

\textbf{Human supervision requirement} As discussed in \S \ref{sec:dataset}, \aspera{} currently relies on human supervision and (optional) interactive prompting to ensure the generation of high-quality and diverse data. Even state-of-the-art models such as GPT-4o (release \texttt{gpt-4o-2024-05-13}) did not reliably produce data suitable for robust capability evaluation without supervision. However, while future improvements in LLM capabilities are expected to reduce this human oversight requirement, the current framework already supports capturing developer interventions and on-the-fly annotation of natural language error descriptions. Over time, this interaction naturally creates a growing corpus of domain-specific errors and corresponding corrections, which can then be leveraged to further fine-tune and improve the performance of the data-generation models themselves.

\textbf{Interactive code generation} Humans write code in an interactive manner \cite{DBLP:conf/nips/YangPNY23}, occasionally relying on execution feedback to correct errors, resolve ambiguities and decompose tasks iteratively. The majority of existing code generation benchmarks, including the current work, consider a non-interactive instruction-to-code sequence transduction process which has the potential for error propagation and a disconnect between the generated code and its execution environment. While the \aspera{} environment supports interactive code generation grounded in environment feedback and observations, we have focused on evaluating LLMs' fine-grained understanding and ability to compositionally use multiple primitives and curated the tasks such that that they are solvable without interaction. In doing so, we have increased the difficulty of certain types of tasks (e.g., scheduling subject to constraints, tasks involving re-scheduling and diary re-organisation). Important baselines to be considered in future work are incremental program generation following ReACT \cite{REACT} framework as well as more agents specialised for software engineering discussed in \S \ref{sec:conclusion}.

\textbf{Primitive selection} While our primitive selection (PS) baseline partially emulates how human developers interact with unfamiliar codebases, it remains relatively simple. As discussed in \S \ref{sec:evaluation}, the agent is not given prior knowledge of available primitives at deployment, making this setting a more realistic and robust measure of an LLM’s ability to execute complex tasks. Future work should explore more sophisticated strategies for hierarchical codebase exploration and incremental action execution program generation.

\textbf{Efficiency} Meeting real-world latency and cost constraints requires agents capable of executing complex queries within tight input and output budgets. In \aspera{}, agents are prompted with \texttt{python}-style function signatures, type annotations and docstrings. Future work should investigate more compact and expressive representations of documentation such as compressed module-level summaries containing key function signatures and type relations. Additionally, for large assistant libraries, concurrent exploration by multiple agents could improve efficiency. These directions hold promise for scaling LLMs to broad virtual assistant deployment scenarios.

\textbf{Scenario-based evaluation} We have designed \aspera{} such that each task can have multiple SIPs and corresponding EPs to support creating contrast sets \cite{DBLP:conf/emnlp/0001ABBBCDDEGGH20} for each task and comprehensively evaluate that agent actions satisfy the user goal regardless of the initial state. However,  unlike in domains such as customer resource management \cite{workbench} or online ordering \cite{appworld} where the user may not know the state of the environment, we assume that the user has complete knowledge of the state of their calendar. Consequently, scenario-based evaluation is very limited in \asperaDataset{} and concerns only queries involving the calendars of other actors in the environment (e.g., other employees) or the room booking system. Moreover, we do not generate states where entities are ambiguous (e.g., two employees share the same surname and the user attempts to schedule a meeting with one of them without further identifying them). Future work could thus extend the SIP generation to support scenario-based evaluation.

\section*{Acknowledgments}
Alexandru Coca's doctoral studies are supported by the EPSRC grant EP/R513180/1. He would like to acknowledge the significant contribution of Prof Bill Byrne from the University of Cambridge to the presentation of this work. The authors also thank Jian Zhang, Rohit Gupta, Ran Zmigrod, and Stephen Pulman from Apple for feedback and improvement suggestions on earlier versions of this work. We also thank the anonymous reviewers and the area chairs for their helpful and constructive improvement suggestions on our original draft. 

\bibliography{custom}

\newpage
\appendix

\clearpage
\onecolumn
\section{\aspera{} dataset generation prompts}
\label{appendix:aspera-dataset-generation-prompts}

\subsection{Joint query and AEP generation}
\label{appendix:aep_gen}

{\small
\begin{tcolorbox}[breakable, colback=lightgray!20, colframe=white, boxrule=0pt, rounded corners, boxsep=0.5mm, width=\linewidth]
\begin{verbatim}
My team needs your help with generation a wide variety of complex programs that can be implemented with 
our application backend. We care to generate only programs that would be generated by our large language 
model when interacting with our application via a voice interface.

Here is our application code.

```python
{{ code }}
```

Here are some examples of high quality programs that we wrote to help you understand the task.

```python
{{ query_solution_examples }}
```

Guidelines:

1. Please limit yourself to generating programs involving complex combinations of the members of our 
codebase. It is not helpful to assume scenarios that our application cannot implement or assume unknown 
details about method implementations - focus on the interfaces and read our documentation carefully.

2. Diversity is key. Focus on user requests that can be parsed to a fairly complex program implemented 
with the codebase above. Just put yourself in the shoes of the user wanting to get a lot done with our 
application. Some ways to achieve diversity may be:
    - imagine scenarios using for loops
    - imagine scenarios based on user conditions
    - imagine scenarios requiring filtering operations
      - imagine many scenarios where multiple dataclasses and their methods are required to support a 
      complex user goal
      - scenarios imagined should always be compositional (ie always have diverse combinations of object 
      attributes and methods operating on them)

3. To reiterate, diversity (2) should not come at the expense of imagining scenarios our codebase cannot 
support (1). We will discuss how to improve our codebase in the future.

### Program structure guidelines ###
The examples above follow {{ guidelines.generation_labelling | length }} structure guidelines listed 
below. Do the same, clearly stating when you follow them in your comments, as demonstrated above.
{% for instruction in guidelines.generation_labelling %}
{{ loop.index }}. {{ instruction }}
{%- endfor %}
\end{verbatim}
\end{tcolorbox}
\captionof{figure}{System turn. In the above the field \texttt{code} is replaced with the documentation of the assistant library and \texttt{query\_solution\_examples} is replaced with $5$ AEP examples. See \Cref{fig:guidelines-generation} for guidelines definition.}
\label{fig:joint-query-gen}
\newpage
\begin{tcolorbox}[breakable, colback=lightgray!20, colframe=white, boxrule=0pt, rounded corners, boxsep=0.5mm, width=\linewidth]
\begin{verbatim}
You have done a stellar job generating some brilliant programs and user queries already. To remind you
of work you completed and keep things brief, we only show the queries extracted from the docstrings of
programs you generated:

{% for q in queries %}
{{ loop.index }}. {{ q }}
{%- endfor %}

Now we have to generate more programs representing complex user utterances. Crucially, these should 
represent a complex set of new user queries, where the user tries to complete different tasks from 
the ones you generated above. *Do not merely paraphrase the queries you already generated* when 
synthesizing programs - think of new and original complex user tasks that our application backend 
supports.

{% if focus -%}
{{ focus }}
{% endif -%}

Let us generate {{ n_programs }} programs.

```python"""
\end{verbatim}
\end{tcolorbox}
\captionof{figure}{User turn. To encourage diversity, we optionally include the history of the queries generated in the prompt, similar to  \citet{imaginarium}. If \texttt{n\_programs} is set to values greater than $1$, multiple programs are generated. The \texttt{focus} field can be changed after each round of interaction, to encourage diversity of generated queries and programs.}
\label{fig:focus-instr}
\bigskip 
\begin{tcolorbox}[breakable, colback=lightgray!20, colframe=white, boxrule=0pt, rounded corners, boxsep=0.5mm, width=\linewidth]
  \begin{itemize}[leftmargin=1mm, itemsep=-1mm]
    \item \texttt{Employee names are generally assumed unique, so you may use find\_employee(name)[0] for resolving a name to an Employee object. Use this sparingly; even though there may be multiple employees with the same name, the user query might give additional information which resolves the ambiguity (eg specify the meeting time). If you decided to make this assumption add a 'by structure guideline \#1' comment.}
    \item \texttt{Work meetings can start after 9:06 AM and should end before 5:10 PM. They don't happen at the weekend unless the user explicitly mentions so.}
    \item \texttt{Type annotate the return for programs which have a return type which is not None.}
    \item \texttt{Do not call functions with default optional values.}          
  \end{itemize}
\end{tcolorbox}
\captionof{figure}{Guidelines iterated over to populate \texttt{\{\{instruction\}\}} 
fields in the loop in \Cref{fig:joint-query-gen}. The first guideline enforces a unique entity name environment constraint, which grounds 0-indexing \texttt{find\_employee} results. We make this design decision to decrease task difficulty for our initial release, but note the LLM is instructed to mark this assumption with \texttt{\# by structure guideline 1} to support future LLM-based annotations of AEPs which handle disambiguation. The second guideline encodes a simple events scheduling policy to be followed when explicit constraints are not provided by the user and when rescheduling events. The third guideline prompts for return type annotation for information-seeking queries and the final guideline encourages concise coding.}
\label{fig:guidelines-generation}
}
\clearpage
\subsection{AEP generation given human-authored request}
\label{appendix:aep-annotation}
\begin{figure*}[htbp]
  \centering
  \tiny
      \begin{subfigure}[t]{\textwidth}
    \begin{tcolorbox}[colback=lightgray!20, colframe=white, boxrule=0pt, rounded corners, boxsep=0.5mm]
    \begin{verbatim}
You are an expert programmer working with my team which is specialising in developing AI assistants. Your current task is to translate a series 
of complex user requests into executable `python` programs using our application backend below:

```python
{{ code }}
```

Here are some examples your colleagues shared with you to help you generate your response in a style that is compatible with our infrastructure:

```python
{{ query_solution_examples }}
```

{% if guidelines.generation_labelling -%}
### Program structure guidelines ###
The examples above follow the {{ guidelines.generation_labelling | length }} structure guidelines listed below. Do the same, clearly stating when you 
follow them in your comments, as demonstrated above.
{% for instruction in guidelines.generation_labelling %}
{{ loop.index }}. {{ instruction }}
{%- endfor %}
{%- else %}
{%- endif %}
    \end{verbatim}
    \end{tcolorbox}
    \caption{System turn. The \texttt{code} and \texttt{query\_solution\_examples} fields are populated with the assistant library documentation and $5$ AEP examples like in the joint AEP and query generation prompt depicted in \cref{fig:joint-query-gen}.}
    \end{subfigure}
\\
  \begin{subfigure}[t]{\textwidth}
  \begin{tcolorbox}[colback=lightgray!20, colframe=white, boxrule=0pt, rounded corners, boxsep=0.5mm]
\begin{verbatim}
Now it's your turn. Please translate the queries below into `python` programs using the examples above to guide your response format. The response should be 
inside a Python markdown block.
{% for q in queries %}
{{ loop.index }}. {{ q }}
{%- endfor %}

```python"""
    \end{verbatim}
    \end{tcolorbox}
    \caption{User turn. The framework supports AEP generation for query batches.}
    \end{subfigure}
\caption{Prompt template used for AEP generation given a human-authored request. (\cref{sec:query-gen})}
\label{fig:aep-annotation}
\end{figure*}

\clearpage
\subsection{SIP generation}
\label{appendix:sip-gen}
\begin{figure}[htbp]
  \centering
  \tiny
      \begin{subfigure}[t]{\textwidth}
    \begin{tcolorbox}[colback=lightgray!20, colframe=white, boxrule=0pt, rounded corners, boxsep=0.5mm]
    \begin{verbatim}
For testing purposes, we need to generate the underlying runtime  state of the user device. Your task is to carefully analyse 
`{{ plan_name }}` along with the application code above and assist our testing team in setting up the runtime environment such 
that `{{ plan_name }}` can be executed and its outputs verified. To do so, you will need to generate a `python` function named `{{ setup_function_name }}`.

We have implemented additional tooling you may find helpful for completing this task:

```python
{{ setup_code }}
```

You may use additional knowledge and create your own functions if needed - custom functions should be defined inside the 
`{{ setup_function_name }}` function. Note how we import modules in the standard python library locally inside the 
`{{ setup_function_name }}` and how our application code does not need to be imported (we automatically do so when we run the code).

Here are some comprehensive examples your testing team colleagues shared to help you generate a high quality program that sets up 
the runtime environment correctly.

```python
{{ runtime_setup_examples }}
```

{% if guidelines.runtime_setup -%}
### Runtime environment setup guidelines ###
The examples above follow the {{ guidelines.runtime_setup | length }} setup guidelines listed below. Do the same, clearly stating when 
you follow them in your comments, as demonstrated above.
{% for instruction in guidelines.runtime_setup %}
{{ loop.index }}. {{ instruction }}
{%- endfor %}
{%- else %}
{%- endif %}

Let's now write `{{ setup_function_name }}`, our developers wrote some TODOs to get you started.

```python
def setup_env_{{plan_name}}():
    """Simulate the environment for the query:

    {{ query }}

    Note this means to create any persons, contacts, emails, events and everything that should exist
    in the user's virtual context when they make the query. You **should not** create new entities that are 
    implied in the user request that the assistant has created in the `{{plan_name}}` function.
    """
'''
    {{ TODOs }}
    \end{verbatim}
    \end{tcolorbox}
    \caption{User turn for SIP generation. This turn is added to the chat history which contains the AEP generation system and user turns and assistant turn with the generated AEPs. \texttt{plan\_name} is the name of the AEP function for which the state is to be initialised and the \texttt{setup\_function\_name} is the name of the SIP to be generated. \texttt{setup\_code} is replacted by the documentation for additional tools the LLM can call to simulate complex environment state. One example is \texttt{simulate\_org} in \cref{fig:aspera-session} (program B, l. 9 - 11) which allows the LLM to simulate an organisation with a complex reporting structure by parametrising the simulation. The \texttt{runtime\_setup\_examples} field shows $5$ SIP examples, which initialise the state for the $5$ AEP examples in the chat history. Guidelines, shown in \cref{fig:runtime-setup-guidelines}, state simulation assumptions. The LLM is prompted to mark these assumptions in comments to enable LLM-assisted refactoring of the SIPs. The \texttt{query} field is replaced by the user query. The \texttt{TODOs} fields marks instruction the developer may optionally specify. These are formatted on separate lines following \texttt{\#TODO:} tags.}
    \end{subfigure}
    \\
  \begin{subfigure}[t]{\textwidth}
  \begin{tcolorbox}[colback=lightgray!20, colframe=white, boxrule=0pt, rounded corners, boxsep=0.5mm]
    \begin{itemize}[leftmargin=1mm, itemsep=-1mm]
    \item \texttt{Dates should be grounded using the tools in the time\_utils library. When doing so, add a 'setup guideline \#1' comment.}
    \item \texttt{Work meetings can start after 9:06 AM and should end before 5:10 PM. When doing so, add a 'setup guideline \#2' comment.}
    \item \texttt{Events assumed to occur in the future should start after the date and time specified by time\_utils.now\_(), whereas events in the past should finish before time\_utils.now\_(). When doing so, add a 'setup guideline \#3' comment.}
    \item \texttt{Employee names are assumed unique, so you may use find\_employee(name)[0] for resolving a name to an Employee object. When doing so, add a 'setup guideline \#4' comment.}
    \item \texttt{Ensure you follow all the TODOs with appropriate steps, but don't be afraid to do additional steps if you think it necessary - our developers may not write detailed enough TODOs.}
    \end{itemize}
    \end{tcolorbox}
    \caption{Guidelines used to populate \texttt{\{\{instruction\}\}} in the bottom loop of (a).}
    \label{fig:runtime-setup-guidelines}
    \end{subfigure}
\caption{Prompt template used for runtime setup program generation (Figure \ref{fig:aspera-session}, B).}
\label{fig:runtime-setup-prompt}
\end{figure}

\clearpage
\subsection{EP generation}
\label{appendix:ep-gen}
\begin{figure}[h!]
  \centering
  \tiny
    \begin{subfigure}[t]{\textwidth}
    \begin{tcolorbox}[colback=lightgray!20, colframe=white, boxrule=0pt, rounded corners, boxsep=0.5mm]
    \begin{verbatim}
We need some test code to check that `{{ plan_name }}` executes correctly on the user device. After a careful analysis of `{{ plan_name }}` and 
`{{ setup_function_name }}` (defined below), your task is to write a function `{{ test_function_name }}` to do so.

We have implemented additional tooling you may find helpful for completing this task:

```python
{{ setup_code }}
```

```python
{{ testing_code }}
```

You may use additional knowledge and create your own functions if needed - custom functions should be defined inside the `{{ test_function_name }}` 
function. Note how we import modules in the standard python library locally inside the s`{{ test_function_name }}` and how our application code does not
need to be imported (we automatically do so when we run the code).

Here are some comprehensive examples your testing team colleagues wrote:

```python
{{ evaluation_examples }}
```
{% if guidelines.evaluation -%}
### Testing guidelines ###
The examples above follow the {{ guidelines.evaluation | length }} setup guidelines listed below. Do the same, clearly stating when you 
follow them in your comments, as demonstrated above.
{% for instruction in guidelines.evaluation %}
{{ loop.index }}. {{ instruction }}
{%- endfor %}
{%- else %}
{%- endif %}

 Here is the code that sets up the runtime environment for `{{ plan_name }}` execution:

 ```python
 {{ runtime_setup_program }}
 ```

Write `{{ test_function_name }}`:

```python
def evaluate_{plan_name}(
    query: str, executable: Callable[[], Any], setup_function: Callable[[], Any]
):
    """Validate that `executable` program for the query

    {{ query }}

    has the expected effect on the runtime environment.

    Parameters
    ----------
    query
        The query to validate
        executable
        The query execution function, `{plan_name}`
    setup_function
        `{setup_function_name}` function.
    """
'''
    \end{verbatim}
    \end{tcolorbox}
    \caption{User turn template for EP generation. This turn is added to the chat history, which contains at this point the user and system turns for AEP and SIP generation.  \texttt{plan\_name},  \texttt{setup\_function\_name} and \texttt{test\_function\_name} are formatted with the AEP, SIP and EP function names, respectively. \texttt{setup\_code} is defined in \cref{fig:runtime-setup-prompt} and \texttt{testing\_code} is replaced by documentation of other tools the LLM can use to verify AEP correctness (see  \Cref{appendix:eval-tools}). The \texttt{evaluation\_examples} field is replaced by $5$ EP examples, which demonstrated how to evaluate the correctness of the AEPs in the interaction history given the SIPs examples. Guidelines, shown in \cref{fig:eval-guidelines}, provide relevant assumptions for writing correct and concise test code ( \Cref{appendix:eval-tools}). The LLM is prompted to mark these assumptions in comments to enable LLM-assisted refactoring of the EPs. The \texttt{runtime\_setup\_program} is the SIP, and \texttt{test\_function\_name} is name of the EP to be generated.}
    \end{subfigure}
    \\
  \begin{subfigure}[t]{\textwidth}
  \begin{tcolorbox}[colback=lightgray!20, colframe=white, boxrule=0pt, rounded corners, boxsep=0.5mm]
    \begin{itemize}[leftmargin=1mm, itemsep=-1mm]
    \item \texttt{fields of type list[Employee] of events returned by find\_events are sorted alphabetically according to the name attribute. Sort attendees lists you create accordingly. When doing so, add a 'testing guideline \#1' comment"}
    \item \texttt{For queries that have a return type, consider a range of possible alternative return types that could have been returned instead by the executable and check the result correctness in those cases too. Add a '\#testing guideline \#2' comment in this case.}
    \item \texttt{When checking events requested by the user were created, never test equality of the 'subject' attribute because variations in the meeting name can affect test robustness.}
    \item \texttt{When add\_event is called without an ends\_at parameter, a default duration of 16 minutes is assumed when writing the event to the underlying database. Check that the events for which end time or duration is not specified satisfy this constraint.}
    \item \texttt{SolutionError message is always 'Incorrect Solution'.}
    \item \texttt{Where possible, use the information in the runtime environment setup function below to simplify testing code.}
    \end{itemize}
    \end{tcolorbox}
    \caption{Guidelines.}
    \label{fig:eval-guidelines}
    \end{subfigure}
\caption{Prompt template used for evaluation program generation (Figure \ref{fig:aspera-session}, C).}
\label{fig:eval-program-prompt}
\end{figure}
\clearpage
\twocolumn
\subsection{Auxiliary \aspera{} tools}
\label{appendix:eval-tools}
\aspera{} defines auxiliary tools designed to aid SIP and EP generation (\cref{tab:sim-eval-tools}). These can be implemented by the developer interactively\footnote{The developer is prompted to implement simulation tools after AEP generation and evaluation tools after SIP generation. The implemented tools are displayed in the subsequent SIP/EP generation prompts.} or (before task generation begins). 
\begin{table}[htpb]
\large
\resizebox{\columnwidth}{!}{
\large
\begin{tabular}{l l p{6cm}}
\toprule
\multicolumn{3}{c}{\textbf{Simulation Tools}} \\ 
\hline
\textbf{Module} & \textbf{Tool} & \textbf{Functionality} \\
\texttt{work\_calendar} & \texttt{simulate\_user\_calendar} & Adds a set of LLM-generated events to the user's calendar.\\
& \texttt{simulate\_employee\_calendar} & Adds a set of LLM-generated events to the calendar of a given employee. \\
\texttt{company\_directory} & \texttt{simulate\_org\_structure} & Build an organisation structure given, employee names, team membership, user name and user role. Reporting relationships and employee profiles are simulated by \aspera{}. \\
& \texttt{simulate\_vacation\_schedule} & Simulate the vacation schedule of a given employee. \\
& \texttt{UserRole} & Enum listing key company roles such as CEO and COO. \\
\texttt{room\_booking} & \texttt{simulate\_conference\_room} & Add a conference room to the conference room database. \\
\hline
\addlinespace[0.5em]
\multicolumn{3}{c}{\textbf{Evaluation Tools}} \\ 
\hline
\textbf{Module} & \textbf{Tool} & \textbf{Functionality} \\
\texttt{time\_utils} & \texttt{repetition\_schedule} &     Create a recurrence schedule for a meeting or reminder. \\
\texttt{work\_calendar} & \texttt{assert\_user\_calendar\_shared} & Check that a calendar has been shared between a list of employees. \\
\bottomrule
\end{tabular}
}

\caption{\asperaAssistantCodebase{} auxiliary tools.} \label{tab:sim-eval-tools}
\end{table}
\paragraph{Simulation tools}  Simulation tools are included in SIP generation prompts to allow the LLM to create entities stored in environment databases. These tools differ in their implementation complexity. Some tools (e.g., \texttt{simulate\_user\_calendar}) simply write LLM-defined entities to the environment databases whereas others can be used to invoke more advanced simulations implemented by developers (possibly with LLM assistance) in \aspera{} (e.g., \texttt{simulate\_org\_structure}). The LLM uses information in the query and the AEP to parametrise the simulation and generates complex entities as a result.
\begin{figure}[h] 
  \centering
  \vspace{-11em}
  \includegraphics[width=\linewidth]{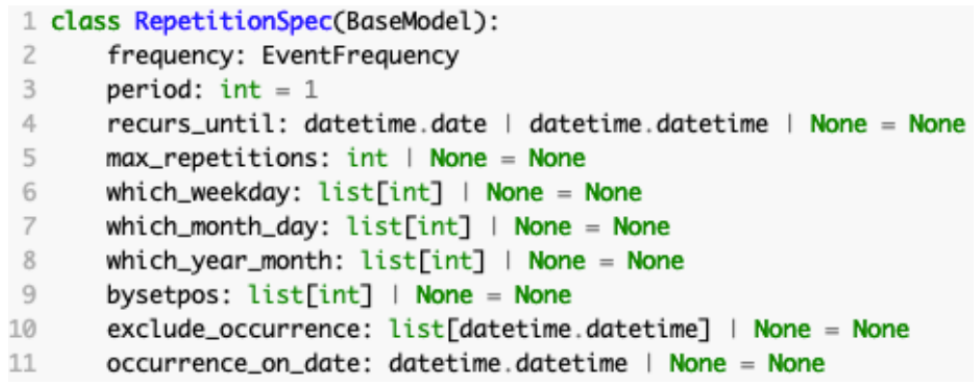}
  \vspace{-11em}
  \caption{Definition of \texttt{RepetitionSpec}, an object used for generating recurring event instances. Documentation omitted for brevity.}
  \label{fig:repetition-spec}
  \vspace{-10pt} 
\end{figure}
\paragraph{Evaluation tools} EP generation prompts include evaluation tools to support robust evaluation and access to environment state that is not possible with the tools the assistant uses to compose AEPs. To understand why this is necessary, consider the query \textit{Remind me to check arxiv on Wednesdays.} To execute this action, the assistant must create an \texttt{Event} instance and set the \texttt{repeats} property to a correctly parametrised recurrence rule (a \texttt{RepetitionSpec} instance, shown in Figure \ref{fig:repetition-spec}). Because the recurrence always inherits the parent event parameters, setting \texttt{which\_weekday=[2]} in this case is optional. More generally, complex recurrences admit multiple parametrisations which are difficult to enumerate for developers. For this reason, we include the \texttt{repetition\_schedule} tool in the prompt so that the LLM can use it to compare the event instances it returns rather than comparing generator object properties. This ensures robust comparison independent of \texttt{RepetitionSpec} parametrisation. 

\clearpage
\newpage
\onecolumn
\section{Assistant library}
\label{appendix:aspera-codebase}

\begin{table*}[htpb]
\resizebox{\textwidth}{!}{
\begin{tabular}{l l l l l l}
\toprule
& \textbf{time\_utils} & \textbf{work\_calendar} & \textbf{company\_directory} & \textbf{room\_booking} \\
\midrule
\textbf{Functions}   &         &     &    &  &   \\
\hline
& \texttt{now\_}   &  \texttt{get\_default\_preparation\_time}       & \texttt{get\_current\_user}     & \texttt{find\_available\_time\_slots}\textsuperscript{$\textdagger$}      \\
& \texttt{get\_weekday}   & \texttt{add\_event}        & \texttt{find\_employee}    & \texttt{room\_booking\_default\_time\_window}      \\
& \texttt{this\_week\_dates}textsuperscript{*}  & \texttt{find\_events}        & \texttt{find\_team\_of}    & \texttt{search\_conference\_room}      \\
& \texttt{get\_weekday\_ordinal}textsuperscript{*}   & \texttt{find\_past\_events}        & \texttt{find\_reports\_of}    & \texttt{summarise\_availability}textsuperscript{*}      \\
& \texttt{parse\_time\_string}textsuperscript{*}   &  \texttt{get\_calendar}       &\texttt{find\_manager\_of}    &       \\
& \texttt{time\_by\_hm}textsuperscript{*}   & \texttt{delete\_event}        & \texttt{get\_assistant}    &       \\
& \texttt{date\_by\_mdy}textsuperscript{*}   &  \texttt{get\_search\_settings}       &  \texttt{get\_vacation\_schedule}   &       \\
& \texttt{get\_next\_dow}textsuperscript{*}   & \texttt{find\_available\_slots}textsuperscript{$\textdagger$}        & \texttt{get\_employee\_profile}    &       \\
& \texttt{get\_prev\_dow}textsuperscript{*}   & \texttt{share\_calendar}        &  \texttt{get\_all\_employees}   &       \\
& \texttt{parse\_duration\_to\_calendar}textsuperscript{*}   &  \texttt{summarise\_calendar}       & \texttt{get\_office\_location}    &       \\
& \texttt{parse\_durations\_to\_date\_interval}textsuperscript{*}   &  \texttt{provide\_event\_details}       &     &       \\
& \texttt{parse\_date\_string}textsuperscript{*}   &         &     &       \\
& \texttt{sum\_time\_units}textsuperscript{*}   &         &     &       \\
& \texttt{compare\_with\_fixed\_duration}   &         &     &       \\
& \texttt{modify}textsuperscript{*}   &         &     &       \\
& \texttt{combine}   &         &     &       \\
& \texttt{intervals\_overlap}textsuperscript{*}   &         &     &       \\
& \texttt{replace}textsuperscript{*}   &         &     &       \\
\textbf{Objects}   &         &     &       \\
\hline
& \texttt{Duration}   &  \texttt{Event}textsuperscript{$\textdagger$}       & \texttt{EmployeeDetails}    & \texttt{ConferenceRoom}      \\
& \texttt{TimeInterval}   & \texttt{CalendarSearchSettings}        & \texttt{Employee}    & \texttt{RoomAvailability}      \\
& \texttt{DateRange}   &         &     &       \\
& \texttt{RepetitionSpec}   &         &     &       \\
\textbf{Enums}   &         &     &       \\
\hline
& \texttt{TimeExpressions}   &  \texttt{ShowAsStatus}       & \texttt{Team}    &       \\
& \texttt{DateRanges}   &         &     &       \\
& \texttt{DateExpressions}   &         &     &       \\
& \texttt{TimeUnits}   &         &     &       \\
& \texttt{DateTimeClauseOperators}   &         &     &       \\
& \texttt{ComparisonResult}   &         &     &       \\
& \texttt{EventFrequency}   &         &     &       \\

\bottomrule
\end{tabular}
}
\caption{The \asperaAssistantCodebase{} assistant library defines 62 primitives across 4 domains, implemented by a single developer with GPT-4o assistance. Primitives marked with \textsuperscript{*} were implemented interactively with the LLM using the ChatGPT graphical user interface. For each primitive, the LLM was prompted with the docstring describing the primitive functionality, and its output subsequently refined until the specification was correctly implemented, if necessary. Unit tests were generated in addition to developer-authored tests to verify complex functionality. Primitives marked with \textdagger were implemented with partial LLM assistance, where the developer described the functionality to be implemented to the LLM, but substantially refactored and enhanced the code. The LLM was also used for generating unit tests for \textdagger primitives. }
\end{table*}

\newpage
\onecolumn

\section{Dataset characterisation}
\label{appendix:dataset-characterisat}

\subsection{Examples of challenging tasks}
\label{appendix:challenging-tasks}

\begin{figure*}[h!]
    \centering
    \begin{framed}
    \begin{subfigure}[t]{0.48\textwidth}
        \centering
        \fbox{\includegraphics[width=\linewidth, trim=0cm 0cm 0cm 0cm]{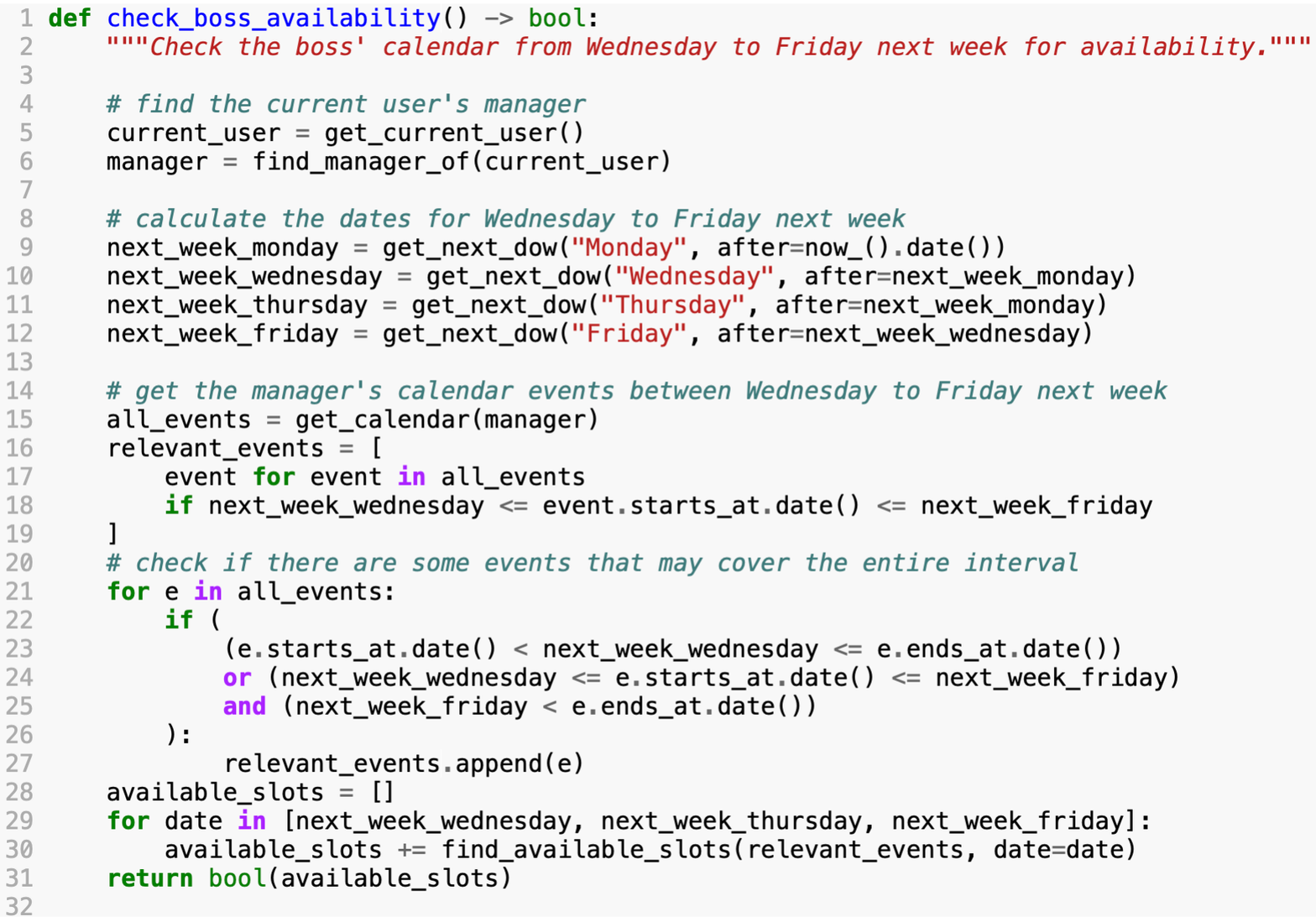}}
        \vspace{-0.2cm}
        \caption{\textit{Assistant, check my boss' calendar Wednesday to Friday next week, are they available for a meeting?} Solving this query involves reasoning about time and having the common sense to account for events spanning multiple days.}
        \label{fig:110}
        \vspace{0.5cm} 
        \fbox{\includegraphics[width=\linewidth, trim=0cm 1cm 0cm 1cm]{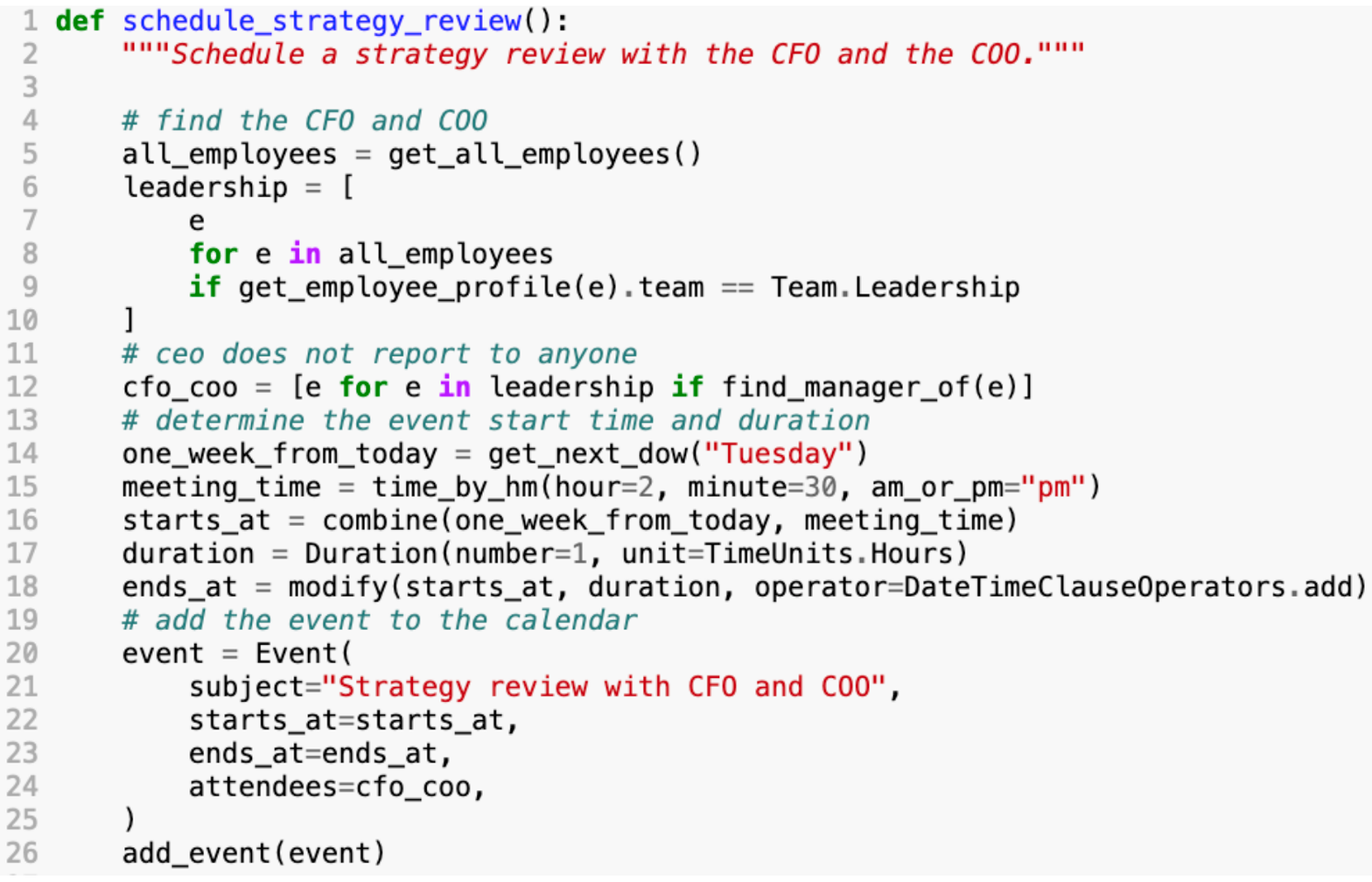}}
        \vspace{-0.2cm}
        \caption{\textit{Assistant, add a strategy review with the CFO and the COO one week from today at 2:30 PM, for 1 hr}. Solving this query involves clever tool use to find the leadership team while taking care to exclude the CEO.}
        \label{fig:207}
    \end{subfigure}
    \hfill
    \begin{subfigure}[t]{0.48\textwidth}
        \vspace{-5cm}
        \centering
        \fbox{\includegraphics[width=\linewidth, trim=0cm 0cm 0cm 0cm]{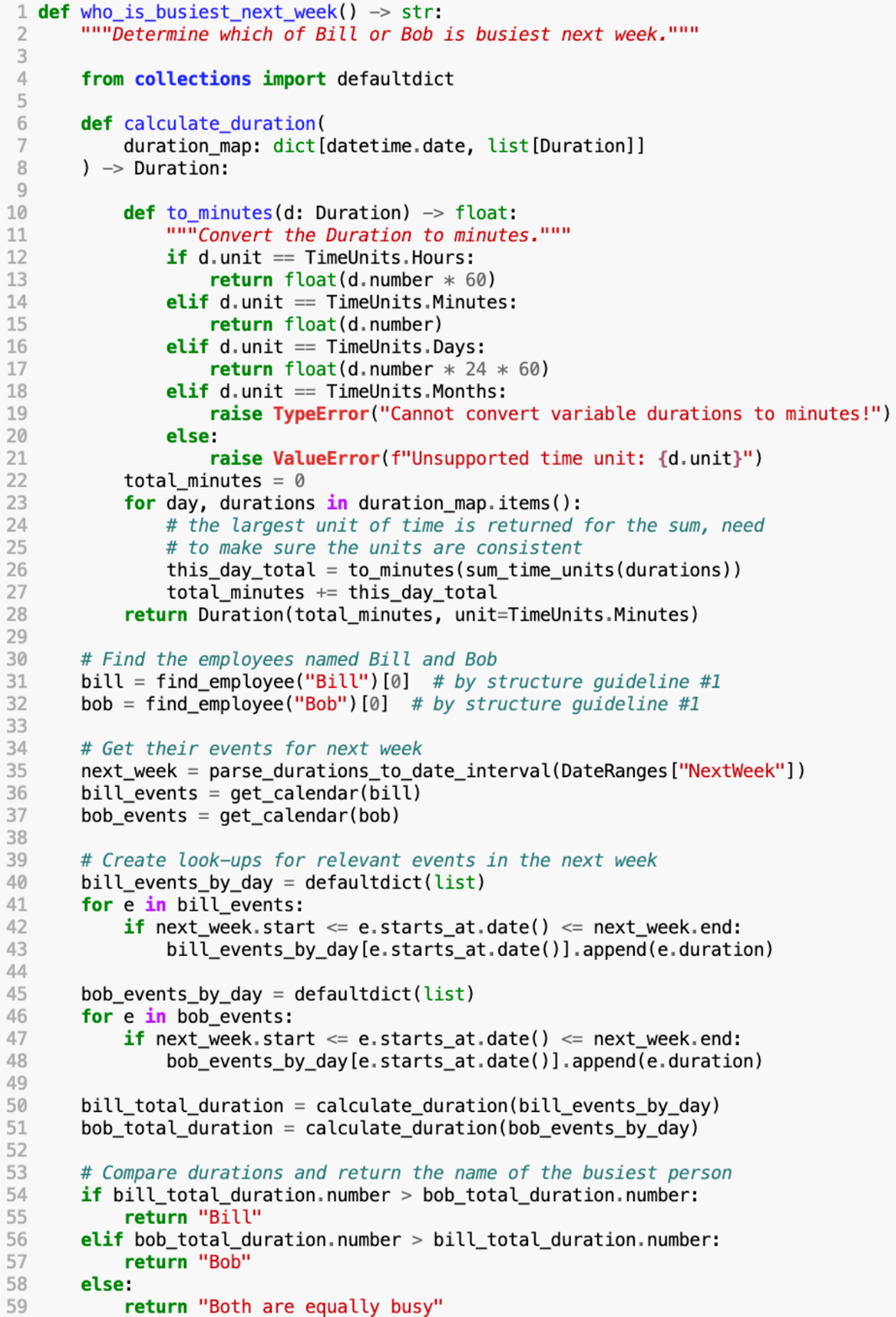}
        \vspace{-0.2cm}}
        \caption{\textit{Assistant, I need to know which of Bill or Bob is busiest next week so I can allocate work.} Here, summing the event duration involves careful unit conversion to provide the correct answer.}
        \label{fig:168}
    \end{subfigure}
    \caption{Challenging queries from lines 3 -5 of Table \ref{tab:query-sample} as particularly challenging. Figures \ref{fig:110}, \ref{fig:168} and \ref{fig:207} show the sample solutions for these queries respectively, with explanations of their difficulty.}
    \label{fig:combined_sample_solutions}
    \end{framed}
\end{figure*}

\newpage
\subsection{Further corpus descriptive statistics}
\label{appendix:corpus-descriptive-stats}

\begin{figure*}[htpb]
  \centering
  \begin{minipage}{0.48\textwidth}
    \centering
    \includegraphics[width=\linewidth]{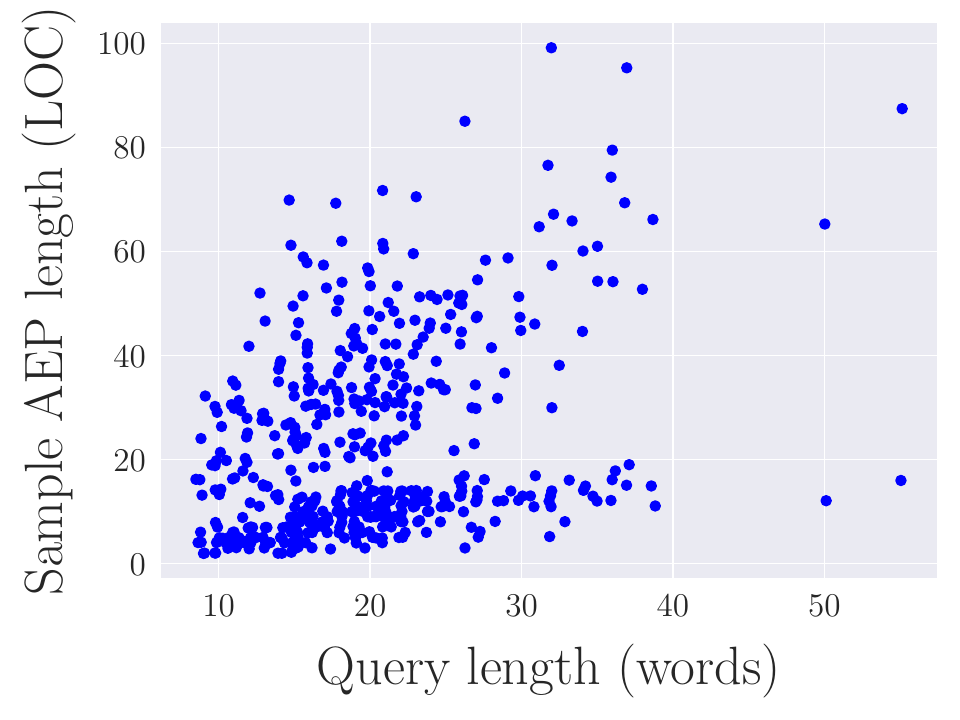}
    \caption{\asperaDataset{} AEP query length vs program length.}
    \label{fig:query-len-vs-program-len}
  \end{minipage}\hfill
  \begin{minipage}{0.48\textwidth}
    \centering
    \includegraphics[width=\linewidth]{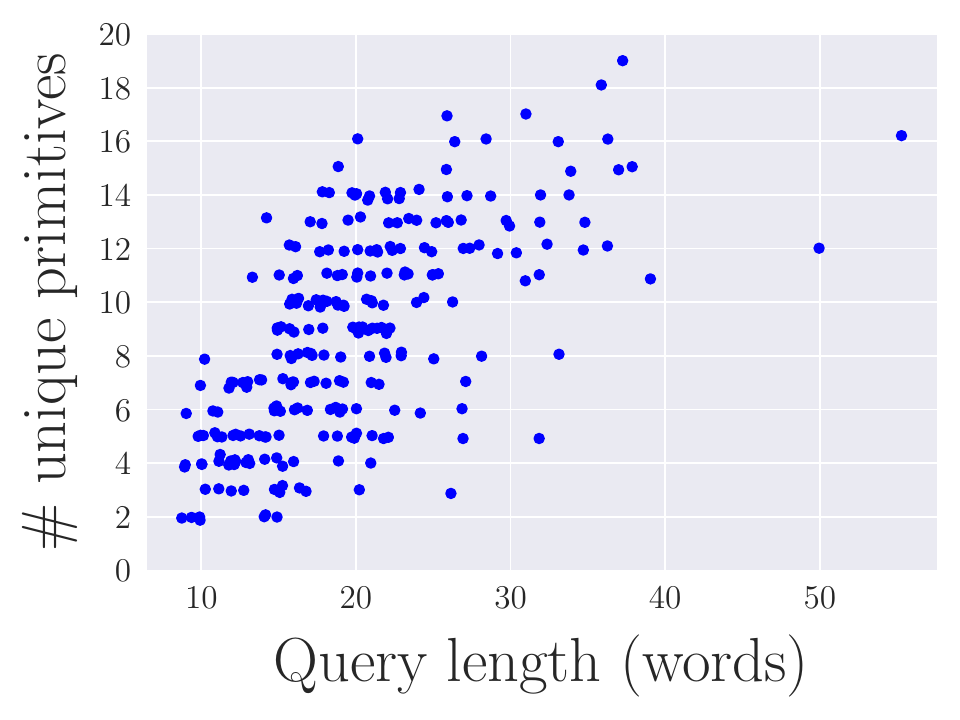}
    \caption{\asperaDataset{} AEP query length vs number of unique primitives.}
    \label{fig:query-len-vs-uniq-primitives}
  \end{minipage}
  \vspace{1em} 
\end{figure*}

\begin{figure*}[htpb]
  \centering
  \begin{minipage}{0.48\textwidth}
    \centering
    \includegraphics[width=\linewidth]{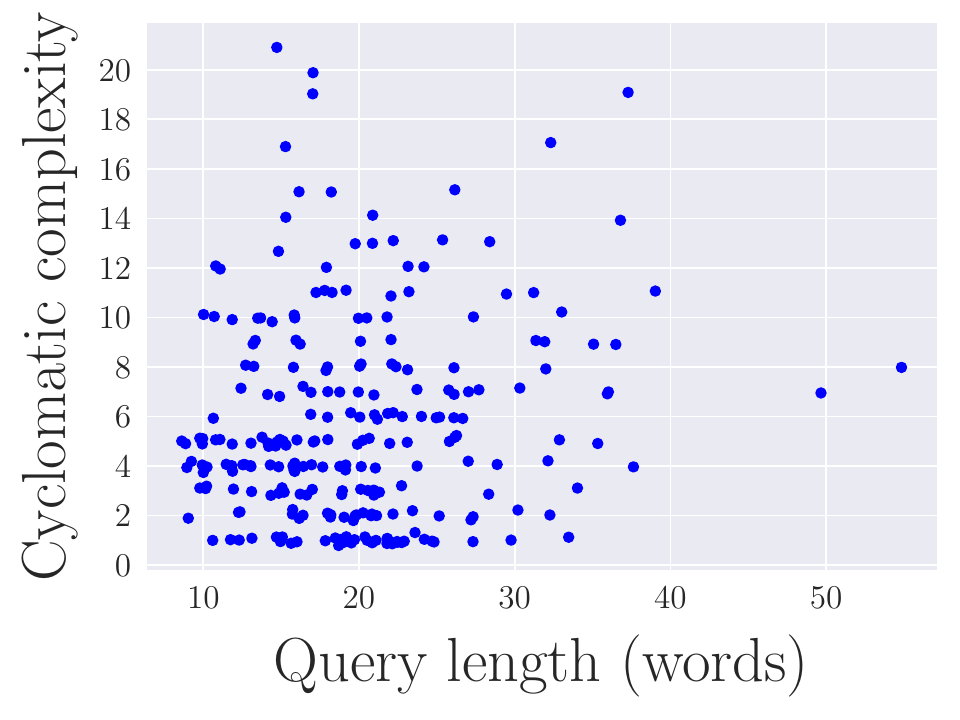}
    \caption{\asperaDataset{} AEP query length vs cyclomatic complexity.}
    \label{fig:query-len-vs-cyc-compl}
  \end{minipage}\hfill
  \begin{minipage}{0.48\textwidth}
      \centering
    \includegraphics[width=\linewidth]{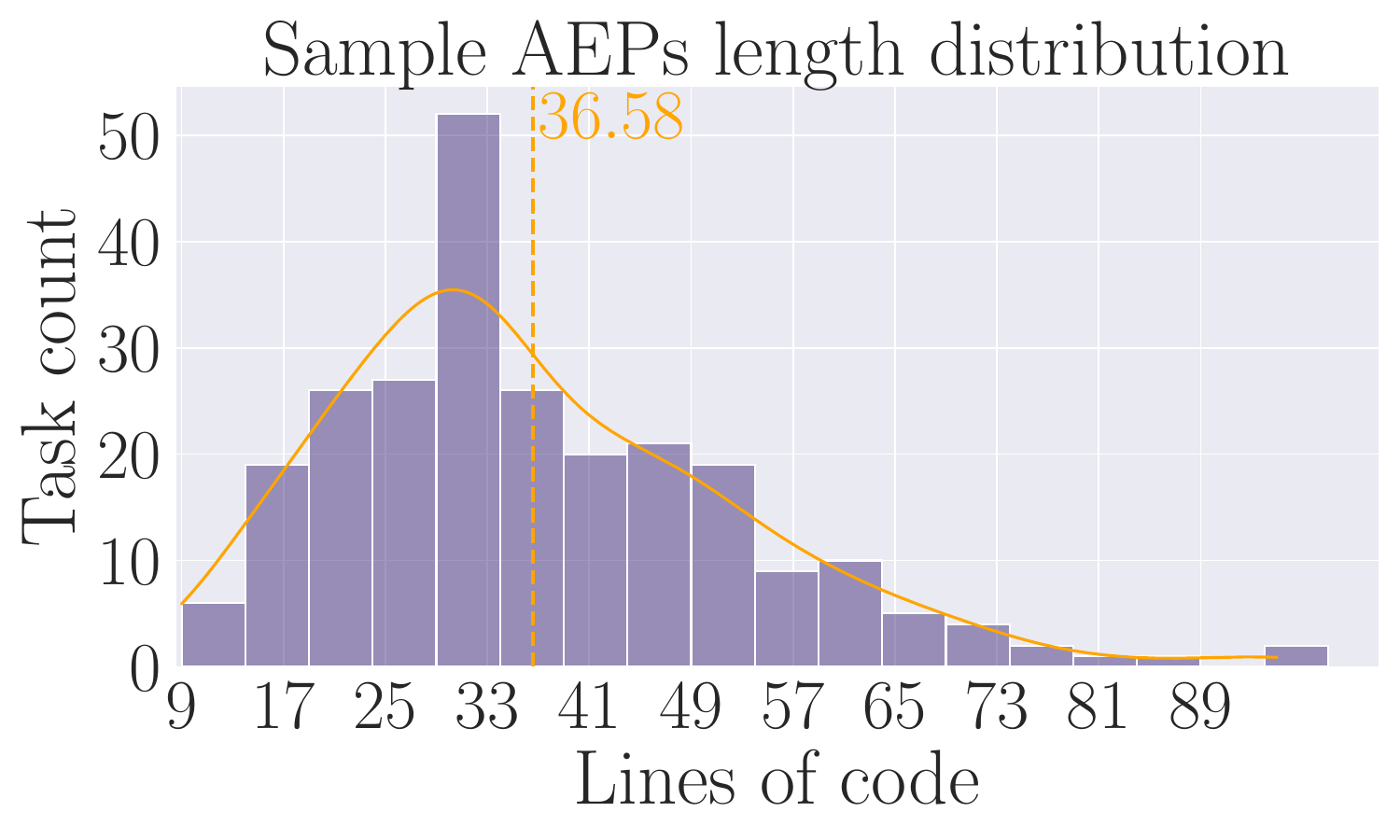}
    \caption{\asperaDataset{} AEP length distribution.}
    \label{fig:loc-distribution}
  \end{minipage}
  \label{fig:scatters-and-loc-dist}
\end{figure*}

Here, we present some further descriptive statistics of \asperaDataset{}. Tables \ref{tab:cyclomatic-complexity-examples} and \ref{tab:num-primitives-examples} show some example queries organised according to their complexity, whereas Figures \ref{fig:query-len-vs-program-len} to \ref{fig:loc-distribution} show how key program complexity measures vary with query length and the distribution of \asperaDataset{} reference AEPs.

\begin{table*}[h]
  \small
  \begin{tabular}{p{10cm}cc}
    \hline
    \textbf{Query} & \textbf{Cyclomatic complexity} & \textbf{$\sigma$ from mean} \\
    \hline
    Assistant, can you tell me when are my manager and skip manager both available on Friday? & \multirow{2}{*}{1.00} & \multirow{2}{*}{-1.14} \\
    Assistant, schedule an urgent meeting with my manager now. & 1.00 & -1.14 \\
    Assistant, schedule a project meeting with my team next Wednesday at 2 PM and block 30 minutes right before for preparation. &  \multirow{2}{*}{1.00} & \multirow{2}{*}{-1.14} \\
    Assistant, schedule a project update meeting with my manager before 3 PM tomorrow. &  \multirow{2}{*}{5.00} & \multirow{2}{*}{-0.16} \\
    Assistant, schedule a meeting in the afternoon with my engineering colleagues, avoiding any engineering management. &  \multirow{2}{*}{6.00} &  \multirow{2}{*}{+0.08} \\
    Assistant, remove my second holiday notification from the calendar, something came up. &  \multirow{2}{*}{7.00} & \multirow{2}{*}{+0.32} \\
    Assistant, send out a meeting invite to the entire team for a company update next Monday at 2 PM, but exclude those who are on vacation. &  \multirow{2}{*}{7.00} & \multirow{2}{*}{+0.32} \\
    Assistant, see if my boss' boss and Jane have accepted my meeting request for tomorrow. If anybody declined, reschedule to take place later but at the earliest available time for everyone, I'm free all day. &  \multirow{3}{*}{19.00} & \multirow{3}{*}{+3.24} \\
    Assistant, tell me which days is Sally in office in the third week of August? & 20.00 & +3.48 \\
    Assistant, is there a time in August where everyone from finance is off? & 21.00 & +3.72 \\
    \bottomrule
  \end{tabular}
  \caption{\label{tab:cyclomatic-complexity-examples} Sampling of queries according to cyclomatic complexity of sample solution.}
\end{table*}

\begin{table*}[h]
  \centering
  \small
  \begin{tabular}{p{10cm}cc}
      \hline
    \textbf{Query} & \textbf{\# unique primitives} & \textbf{$\sigma$ from mean} \\
    \hline
    Assistant, how many meetings with Jianpeng are in my calendar at the moment? & 2 & -1.79 \\
    Assistant, cancel everything but the important meetings. & 2 & -1.79 \\
    Assistant, find the names of our assistants please. & 2 & -1.79 \\
    Assistant, schedule a meeting with my manager tomorrow at 10 AM if I have no other meetings then. & \multirow{2}{*}{9} & \multirow{2}{*}{+0.04} \\
    Assistant, provide a summary of my manager's calendar for the next two weeks. & 9 & +0.04 \\
    Assistant, invite the entire sales department to a meeting today from 3 to 5. & 9 & +0.04 \\
    Assistant, schedule a team meeting next Monday at 10 AM, and book a conference room for it. Also, schedule a follow-up meeting one week later at the same time and book the same room. & \multirow{3}{*}{18} & \multirow{3}{*}{+2.39} \\
    Assistant, can you schedule a 30 mins recurring weekly meeting with the engineering team on Fridays at 3 PM for the next two months? If there are clashes, tell me their dates, don't double book. & \multirow{3}{*}{19} & \multirow{3}{*}{+2.65} \\
        \bottomrule
  \end{tabular}
  \caption{\label{tab:num-primitives-examples} Sampling of queries according to number of unique primitives in sample solution.}
\end{table*}

\newpage

\clearpage
\twocolumn
\vspace*{-1.5cm}

\subsection{\aspera{} policy}
\label{appendix:policy-prompt}
\begin{figure}[!htbp]
  \centering
    \begin{subfigure}[b]{\columnwidth}
    \centering
    \includegraphics[width=\linewidth, trim=0mm 0mm 0mm 0mm, clip]{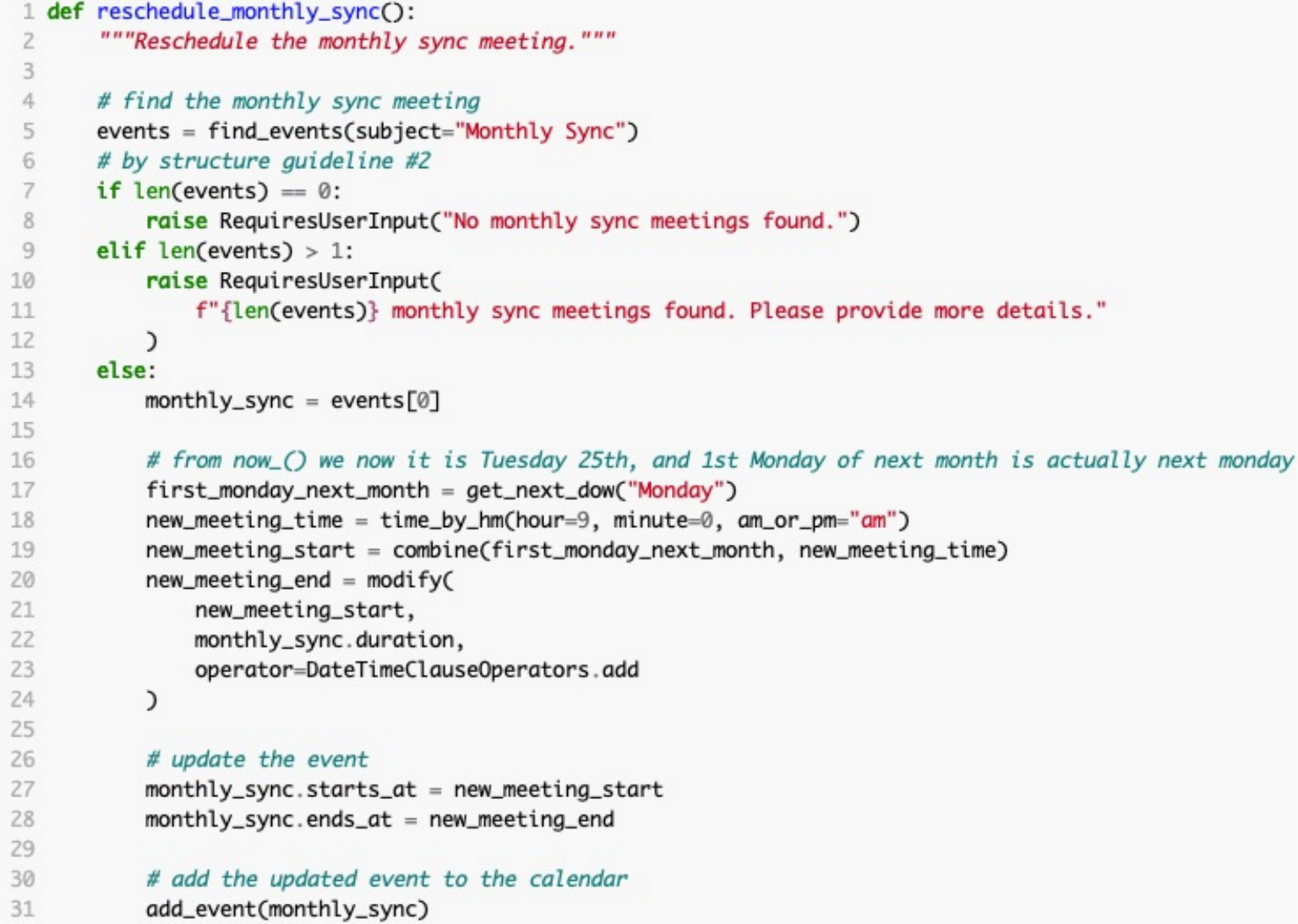}
    \caption{LLM-generated policy for error handling and disambiguation.}
    \label{fig:disambiguation}
  \end{subfigure}

  \vspace{1em}  
  \begin{subfigure}[b]{\columnwidth}
    \centering
    \includegraphics[width=\linewidth]{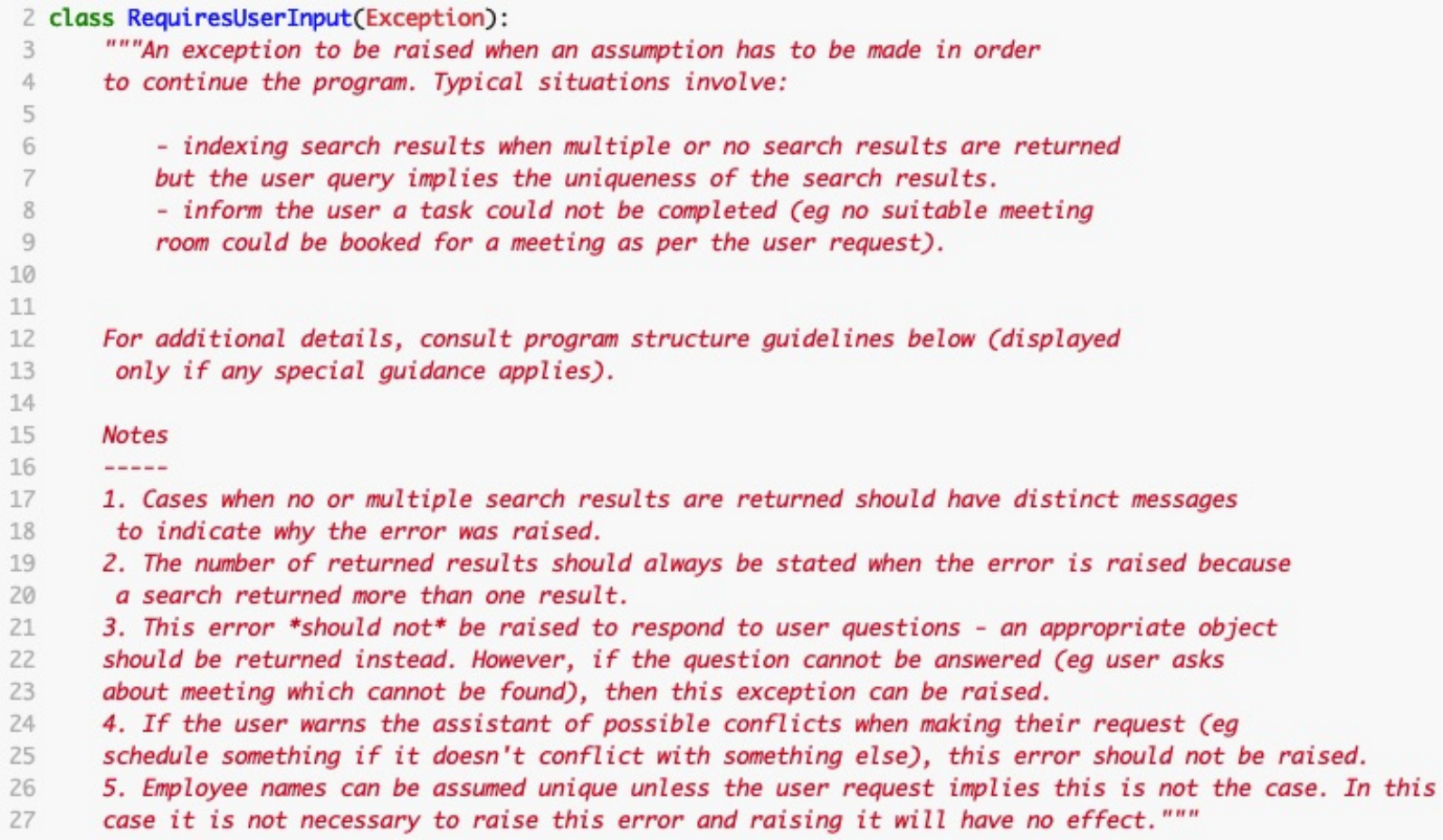}
    \caption{\texttt{RequiresUserInput} documentation}
    \label{fig:req_user_input}
  \end{subfigure}

  \caption{\aspera{} employs exceptions to generate reference AEPs following a simple policy:
    the assistant raises \texttt{RequiresUserInput} if a task cannot be completed due
    to environment constraints or if the user must disambiguate. We observe $144$
    \texttt{RequiresUserInput} usages across $78$ programs. Additionally, top guidelines
    in Figure~\ref{fig:guidelines-generation} enforce a simple scheduling policy.}
  \label{fig:stacked_in_one_column}
\end{figure}

\asperaDataset{} programs follow a policy for interrupting execution to interact with the user: the \verb|RequiresUserInput| exception is raised if the entities mentioned by the user cannot be retrieved from the databases\footnote{Given the complexity of our tasks, we always simulate these entities; we leave adversarial user behaviour (e.g., the user deliberately requests to update an event that is not in the calendar) to future work.} or the task cannot be completed (e.g., a room is unavailable), as shown in \Cref{fig:disambiguation}.

\clearpage
\onecolumn

\section{\aspera{} evaluator prompt templates}
\label{appendix:agent-prompts}

\begin{figure}[H]
  \centering
  \tiny
    \begin{subfigure}[t]{\textwidth}
    \begin{tcolorbox}[colback=lightgray!20, colframe=white, boxrule=0pt, rounded corners, boxsep=0.5mm]
    \begin{verbatim}
You are an expert programmer working with my team which is specialising in developing AI assistants. Your current task is to translate a complex user 
request into a `python` program using our application backend below:

```python
{{ code }}
```

Here are some examples your colleagues shared with you to help you to understand the solution format and some assumptions about our application backend.

```python
{{ query_solution_examples }}
```

The examples above follow the {{ guidelines.generation_labelling | length }} structure guidelines listed below. You must adhere to these when writing your 
solution.
{% for instruction in guidelines.generation_labelling %}
{{ loop.index }}. {{ instruction }}
{%- endfor %}
    \end{verbatim}
    \end{tcolorbox}
    \vspace{-0.2cm}
    \caption{Prompt template for AEP generation, shared by CCK and PS agents.  The syntax (\texttt{\{\{ variable \}\}, \{\% loop \%\})} follows standard templating convention, where placeholders represent dynamically inserted content and loops iterate over a list of instructions. See guidelines below.}
    \label{fig:agent-prompt-template}
    \end{subfigure}
  \\
  \begin{subfigure}[t]{\textwidth}
      \begin{tcolorbox}[colback=lightgray!20, colframe=white, boxrule=0pt, rounded corners, boxsep=0.5mm]
    \begin{itemize}[leftmargin=1mm, itemsep=-1mm]
        \item \texttt{Unless the user explicitly states, meetings should not be scheduled on or recur during weekends.}
        \item \texttt{Work meetings can only happen during the times prescribed in the time\_utils library unless the user explicitly states otherwise.}
         \item \texttt{The leadership team is formed of a CEO, COO, CFO. Department heads report to either the COO or the CFO.}
        \item \texttt{Use the tools in the time\_utils library to reason about time. Hence, current date and time on the user device should be found using the tools and documentation in this library and not the datetime library.}
        \item \texttt{Information-seeking queries should return an appropriate object to the caller; avoid simply printing the information inside your solution.}
        \item \texttt{If you need to format dates in a string, use strftime('\%Y-\%m-\%d'). For datetime objects use strftime('\%Y-\%m-\%d \%H:\%M:\%S').}
        \item \texttt{Make sure to escape \textbackslash n characters.}
        \item \texttt{Type annotate the return for programs which have a return type which is not None}
        \item \texttt{Only the first Python markdown block will be executed, so if you wish to use helper functions, these should be defined locally inside your solution.}
        \item \texttt{Only import modules from the standard library that you need for your programs (eg import collections). Imports from our application backend will be automatically done when we execute the program you generate.}
    \end{itemize}
    \end{tcolorbox}
  \vspace{-0.2cm}
  \caption{The first two guidelines implement a simple events schedule policy. The third provides additional information about the environment, required to solve a range of queries involving the organisation leadership. The remainder of the guidelines are concerned with various aspects of the AEP structure such as time grounding, return type, function nesting and importing. These guidelines were designed to minimise execution errors due to mismatches between the simulation environment and model behaviour following detailed error analyses on initial agent development iterations.}
  \label{fig:agent-guidelines}
  \end{subfigure}
 \caption{\aspera{} AEP generation prompt template.}
 \label{fig:apera-eval-prompts}
\end{figure}
\begin{figure}[htpb]
  \centering
  \tiny
\subsection{Primitive selection prompt}

  \begin{subfigure}[t]{\textwidth}
      \begin{tcolorbox}[colback=lightgray!20, colframe=white, boxrule=0pt, rounded corners, boxsep=0.5mm]

    \begin{verbatim}
You are a programmer using a Python library of personal assistant tools in order to write a program that executes a user query. You will be shown signatures 
from a Python module and a query, and will be asked to formulate Python import statements importing any tools that might be relevant to writing a program that 
executes the user query.

When writing the program, you will be asked to follow the {{ guidelines | length }} structure guidelines listed below.
{% for instruction in guidelines %}
{{ loop.index }}. {{ instruction }}
{%- endfor %}
Use this additional information to guide your import decisions.

Module:
{{ module }}
Query: {{ query }}

Think carefully, and output the relevant Python import statements, or None. Any code you write must be included in a Python markdown block (ie start with a 
"```python" sequence and end with "```"). If there are no relevant tools in the current module being shown, simply output None.
    \end{verbatim}
    \end{tcolorbox}
    \vspace{-0.2cm}
    \caption{Primitives selection prompt template.}
    \label{fig:ps-prompt-template}
  \end{subfigure}
  \\
  \begin{subfigure}[t]{\textwidth}
          \begin{tcolorbox}[colback=lightgray!20, colframe=white, boxrule=0pt, rounded corners, boxsep=0.5mm]
    \begin{itemize}[leftmargin=1mm, itemsep=-1mm]
      \item \texttt{Use the tools in the time\_utils library to reason about time. Hence, current date and time on the user device should be found using the tools and documentation in this library and not the datetime library.}
    \item \texttt{Work meetings can only happen during the times prescribed in the time\_utils library unless the user explicitly states otherwise.}
      \item \texttt{The leadership team is formed of a CEO, COO, CFO. Department heads report to either the COO or the CFO. Appropriate tools will have to be imported and combined to resolve these employees to Employee objects required by all APIs.}
    \end{itemize}
    \end{tcolorbox}
  \vspace{-0.2cm}
  \caption{Guidelines presented to the agent during at each primitive selection iteration step. These are a subset of the guidelines defined for the CCK prompt in \cref{fig:agent-guidelines}, including only the instructions which can influence primitive selection.}
  \label{fig:primitives-selection-guidelines}
  \end{subfigure}
  \caption{Primitive selection prompt}
\end{figure} 

\newpage
\twocolumn
 
\section{Analysis supplementary material}
\label{appendix:extended-results}

\subsection{Execution errors}
\label{appendix:execution-err-analysis}
\begin{figure}[h!]
  \centering
  \begin{subfigure}{\linewidth}
    \centering
    \includegraphics[width=\linewidth]{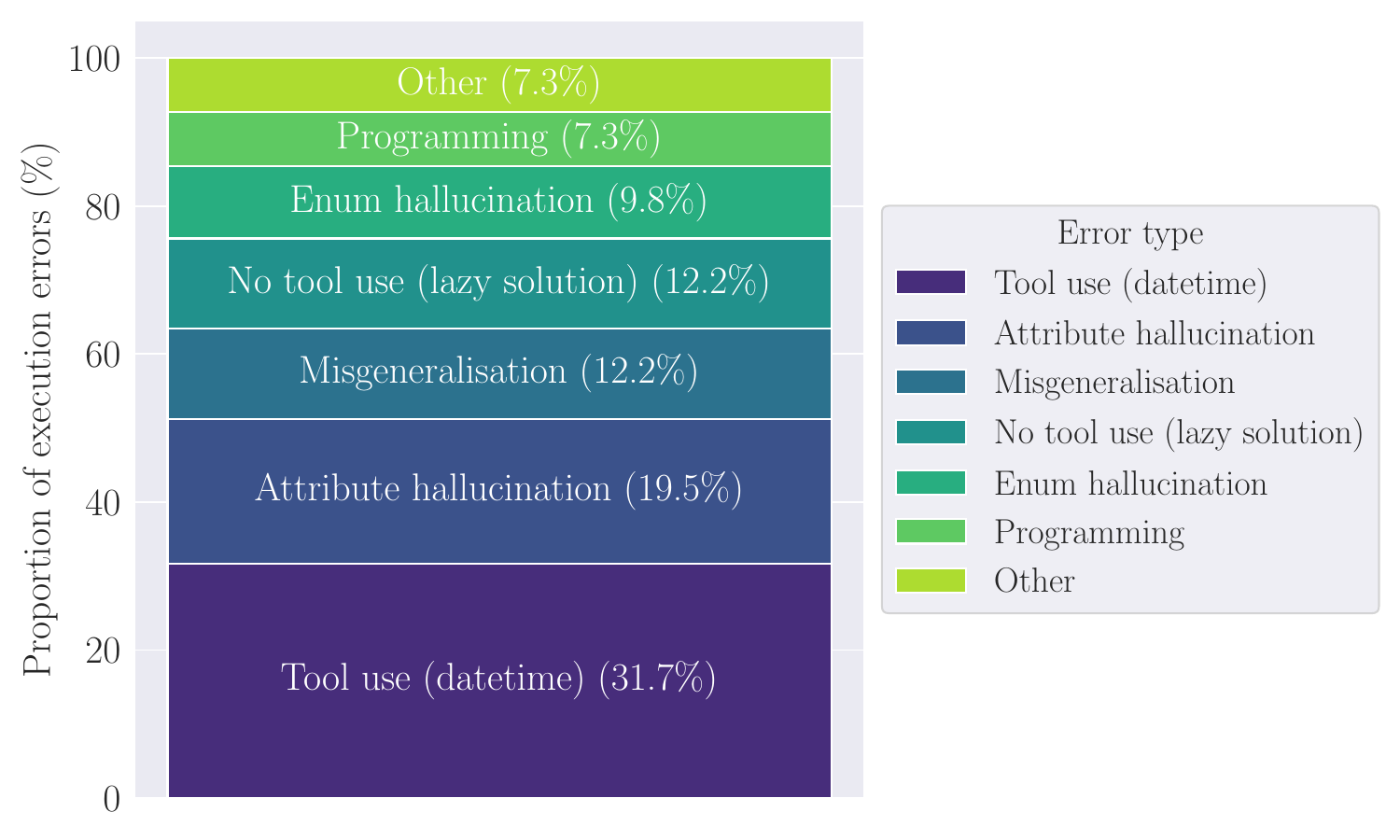}
    \caption{GPT-4o}
    \label{fig:gpt4o-execution}
  \end{subfigure}
  \\
  \begin{subfigure}{\linewidth}
    \centering
    \includegraphics[width=\linewidth]{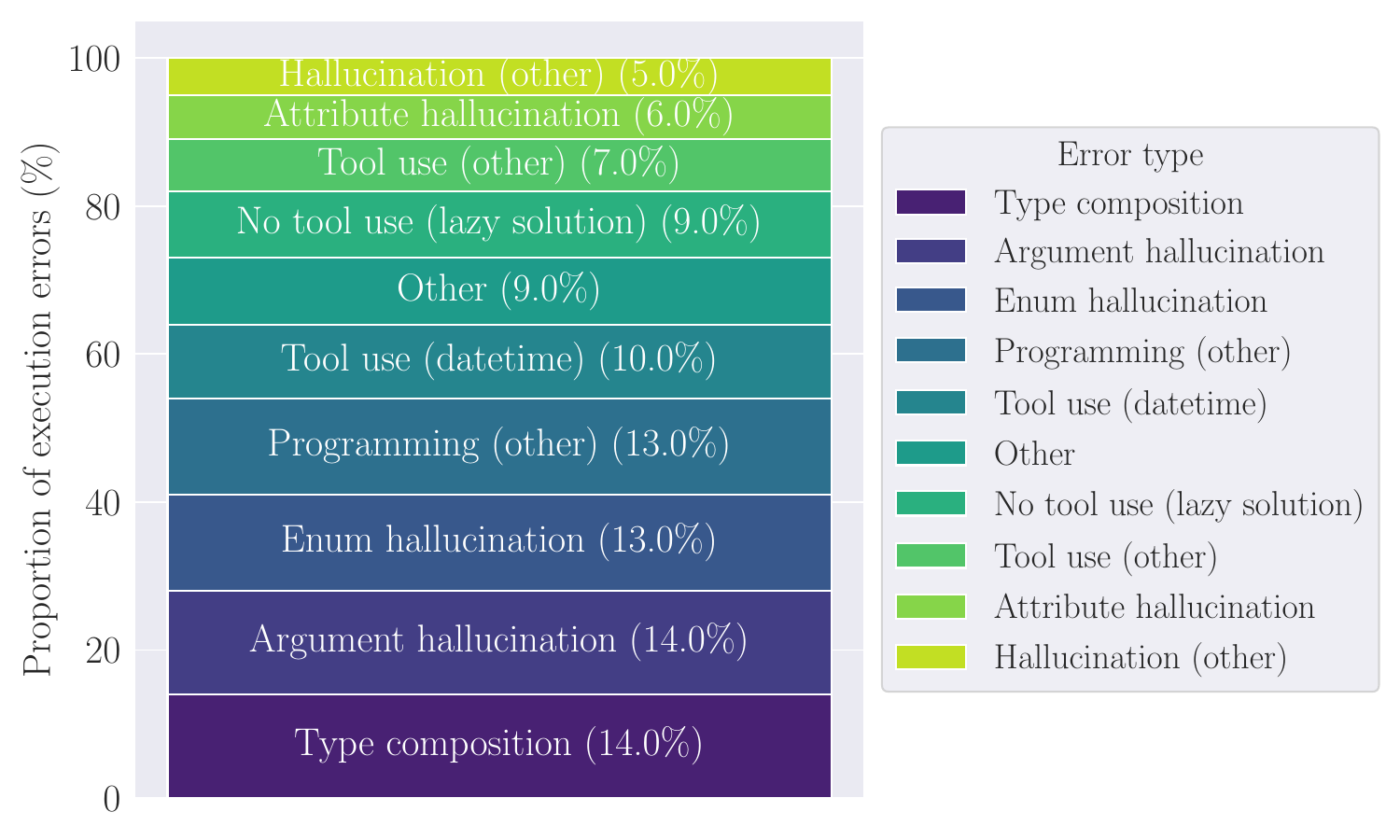}
    \caption{GPT-3.5-turbo}
    \label{fig:gpt35-turbo-execution}
  \end{subfigure}
  \caption{Execution error classification for the first $125$ \asperaDataset{} queries.}
  \label{fig:execution-error-analysis}
\end{figure}

We debug the AEPs generated by the best GPT-4o  and GPT-3.5-turbo runs\footnote{Success rate of $46\%$ and $11.2\%$, respectively.} for the first $125$ queries in our corpus ($50\%$ of the data), analysing a total of $141$ execution errors (Table \ref{tab:execution-err-analysis}) which we classify into several categories depicted in Figure \ref{fig:execution-error-analysis} and for which representative examples are shown and explained in Table \ref{tab:error_snippets}. We find execution errors occur because the LLMs hallucinate in preference to performing additional problem solving steps. While GPT-4o fails to appropriately combine the primitive to perform non-trivial compositions for date and time reasoning (row 1, Table \ref{tab:error_snippets}) or simple arithmetic reasoning (row 5), GPT-3.5-turbo additionally fails to appropriately exploit type relations to compose primitives (row 6) and often hallucinates API arguments (row 7), demonstrating very limited ability to program according to a complex set of constraints defined by an assistant library. 
\begin{table*}[t]
    \vspace{-4cm}
  \centering
  \tiny
  \resizebox{\textwidth}{!}{
  \begin{tabular}{>{\raggedright}l  m{7cm}  m{10cm}}
    \hline
    \textbf{Id} & \textbf{Query} & \hspace{0.75cm} \textbf{Error Snippet} \\
    \hline
    1 & Assistant, schedule our team Christmas party 10 days before Christmas. Should start in the morning and end at 10 PM. &
      \includegraphics[width=0.48\textwidth, trim=0cm 11cm 1cm 11cm, clip]{figures/results/appendix/tool_use_datetime.pdf} \\
    \multicolumn{3}{p{\textwidth}}{\textbf{Tool use (datetime):} Line \texttt{9} contains a \texttt{TypeError}, \texttt{modify} only accepts \texttt{datetime} objects. A correct solution requires an additional reasoning step: 
    pass \texttt{christmas\_day} and one of specified times to the \texttt{combine} library function to get the correct type.}\\
    \hline
    2 & Assistant, set up a training session for all employees from the Engineering team next Monday from 2 PM to 5 PM. Send out invites and book a conference room that fits 20 people. &
      \includegraphics[width=0.48\textwidth,  trim=2.5cm 12.5cm 1cm 12.5cm, clip]{figures/results/appendix/attribute_hallucination.pdf} \\
    \multicolumn{3}{p{\textwidth}}{\textbf{Attribute hallucination:} In, line \texttt{6} the \texttt{.team} attribute access raises an error because the \texttt{Employee} objects returned by \texttt{get\_all\_employees} only have \texttt{name} as attribute. The \texttt{Employee} object should be passed instead to the \texttt{get\_employee\_profile} library function to return an object which has \texttt{team} as an attribute.} \\
    \hline
    3 & Assistant, can you schedule a 30 mins recurring weekly meeting with the engineering team on Fridays at 3 PM for the next two months? If there are clashes, tell me their dates, don't double book. &
  \includegraphics[width=0.48\textwidth,  trim=1cm 12.5cm 1cm 12.5cm, clip]{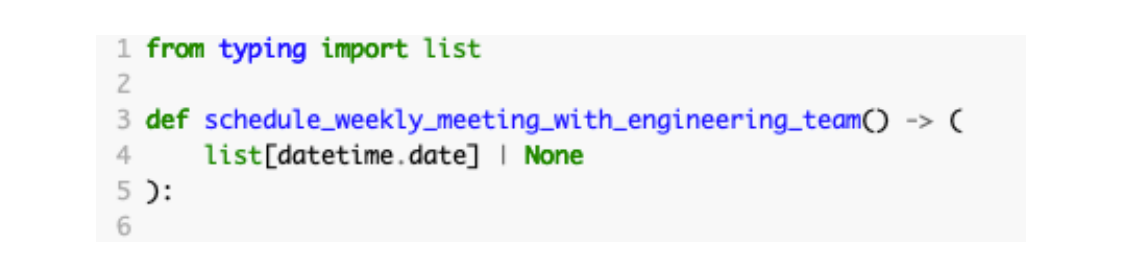} \\
    \multicolumn{3}{p{\textwidth}}{\textbf{Misgeneralisation:} The assistant triggers an import error in line 1. The pretraining data contains \texttt{from typing import List}, a common idiom for static typing prior to PEP 585 (2019). When prompted to return an object of type \texttt{list[datetime.date] | None}, the model does not make this distinction and misgeneralises by generating line \texttt{1}.} \\
    \hline
   4 & Assistant, put 45 minutes in the calendar, back-to-back, with Engineering and Marketing starting at 10 AM tomorrow... Actually, add a 10-minute buffer between each meeting. &
  \includegraphics[width=0.48\textwidth,  trim=0.25cm 12.5cm 1cm 12cm, clip]{figures/results/appendix/lazy_solution.pdf} \\
    \multicolumn{3}{p{\textwidth}}{\textbf{No tool use (lazy solution):} The assistant hallucinates lines \texttt{7-8} instead of using relevant APIs to find the engineering team, in spite of documentation that states that \texttt{Employee} objects cannot be instantiated. The functions \texttt{get\_all\_employees}, \texttt{get\_employee\_profile} and the enumeration \texttt{Team.Engineering} should have been composed, similar to snippet in row 2.} \\
    \hline
  5 & Assistant, mark my vacation from next Tuesday for 2 weeks and cancel all my meetings during this period. &
  \includegraphics[width=0.48\textwidth,  trim=2.5cm 11.5cm 1cm 11.5cm, clip]{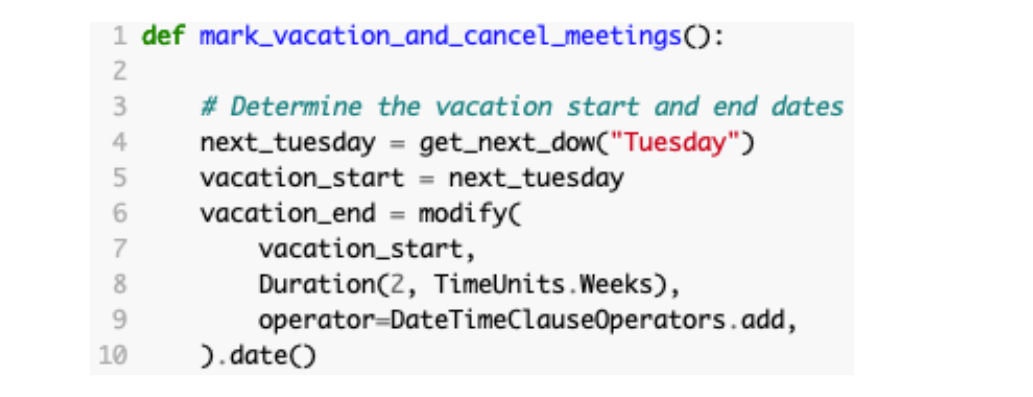} \\
  \multicolumn{3}{p{\textwidth}}{\textbf{Enum hallucination:} The assistant uses the enum value \texttt{TimeUnits.Weeks} (line 8), which is undefined. The library deliberately defines the \texttt{TimeUnits} members as "Hours", "Minutes", "Days", "Months" so that assistants have to perform simple unit conversions.} \\
    \hline\hline
  6 & Assistant, notify me of overlapping meetings this week. & \hspace{0.3cm}  \includegraphics[width=0.48\textwidth,  trim=0cm 10cm 3cm 8cm, clip]{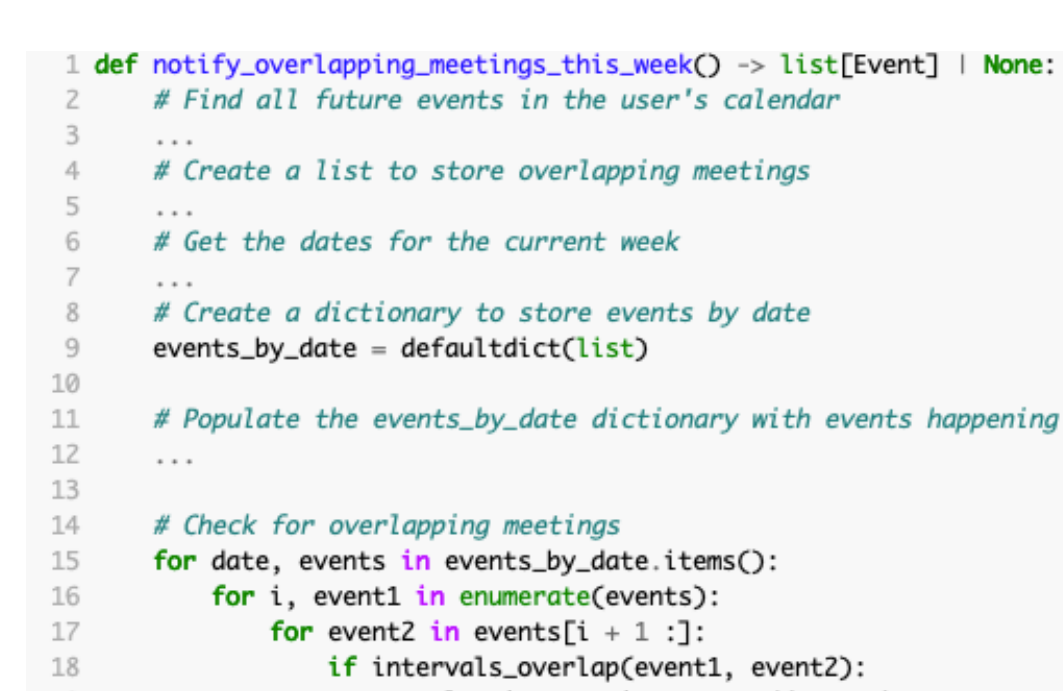} \\
    \multicolumn{3}{p{\textwidth}}{\textbf{Type composition:} The assistant calls \texttt{intervals\_overlap} with \texttt{Event} instead of \texttt{TimeInterval} types (line 18). The latter must be instantiated from the event properties. } \\ \hline
  7 & Assistant, block time for preparation before important meetings. & \hspace{0.65 cm} \includegraphics[width=0.48\textwidth,  trim=0.5cm 13.5cm 0cm 13cm, clip]{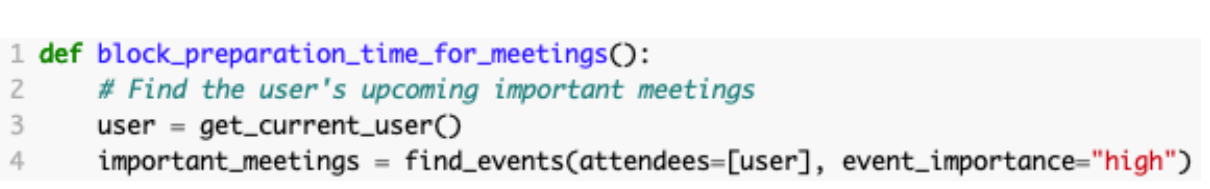} \\ \multicolumn{3}{p{\textwidth}}{\textbf{Argument hallucination:} The assistant calls \texttt{find\_event} with \texttt{event\_importance} keyword (line 4). Valid \texttt{find\_event} arguments are \texttt{attendees} and \texttt{subject}.} \\ \hline
  \end{tabular}
  }
  \caption{\label{tab:error_snippets} Sample execution errors.}
\end{table*}

\newpage
\onecolumn

\subsection{Task completion error examples}
\label{appendix:task-completion-errors}

\begin{table*}[htpb]
  \centering
  \tiny
  \resizebox{\textwidth}{!}{
  \begin{tabular}{p{0.1cm} p{8.5cm} p{6.5cm}}
    \toprule
    \textbf{Id} & \textbf{Query} & \textbf{Agent action} \\
    \midrule

    1 & Assistant, Ari and James are on holiday next month, who's out for longer? & Sums duration of all vacations, month notwithstanding. \\ \addlinespace[0.5ex]
    \multirow{2}{*}{2} & \multirow{2}{*}{Assistant, reorganise my diary on the fifth so that the important meetings come first.} & Sets the importance of the first low-priority meeting to "high" and all other events to "normal", without any further updates. \\
    \addlinespace[0.5ex]
    3 & Assistant, is there a time in August where everyone from finance is off? & Returns \texttt{True} for the first employee whose vacation starts in August. \\
    \hline
     \multirow{2}{*}{4} & \multirow{2}{*}{Assistant, book a conference room for the meeting with sales tomorrow at 2 PM.}  & Assumes the user is part of the sales team, scheduling a meeting with the wrong attendees as a result. \\
    5 & Assistant, add bi-weekly mentorship sessions with the reports of my reports starting next Monday at 2 PM to my calendar. & Hallucinates an end date for the recurrent event, scheduling instances only for six months. \\ 
    \multirow{2}{*}{2} & \multirow{2}{*}{Assistant, add a reminder 1 hour before all important meetings, with the meeting title in the subject.} & Disregards \texttt{add\_event} documentation according to which the user should not be a member of attendees lists for events in their own calendar. \\  
    \hline
    7 & Assistant, schedule by-monthly team training sessions on the first Monday at 10 am for hires who joined since the 1st of May, alternating between the Engineering and Sales and Marketing. & Cannot correctly resolve the meeting start dates scheduling two meetings which start at the same time in the first Monday of the current month, which has already passed. \\
    8 & Assistant, cancel all my meetings Wednesday next week and mark me out of office & Cancels meetings on Wednesday in the current week instead \\
    9 & Assistant, how many employees called John are in my team? & Exact matches the \texttt{name} attribute instead of calling 
    \texttt{find\_employee('John') and filtering to ensure returned employees are in user's team} \\
    \hline
    \multirow{3}{*}{10} & \multirow{3}{*}{Assistant, what date did Joris and Pete meet last week?} & Wrong information provided to the user because the model is looking for a meeting involving Joris and Pete in user's calendar as opposed to checking either Joris' or Pete's calendar. \\ 
    11 & Assistant, reschedule the meetings which overlap with the annual review
    this afternoon to the same time tomorrow. & Adds copies of overlapping events tomorrow, instead of modifying existing events. \\
    12 & Assistant, schedule a 30 mins meeting with Frank from finance at 10 AM in any available meeting room. & Schedules a meeting in the wrong room, choosing the first room returned by the room search API without first checking availability for the entire duration specified by the user. \\
    \hline
    \multirow{2}{*}{13} & \multirow{2}{*}{Assistant, can you find a room that can accommodate 20 people for a meeting on Thursday afternoon?} & Incorrectly processes serch results, returning rooms that are not available during the stated interval \\ 
    14 & Assistant, who in our team has not booked any vacations yet? & Includes the user in the list of returned names, not expected since the user was asking about other team members, not themselves. \\ 
    15 & Assistant, reschedule all meetings from today to next Monday. & Reschedules all the meetings happening until next Monday to next Monday instead of rescheduling today's meetings. \\ 
    \bottomrule
  \end{tabular}
}
  \caption{\label{tab:3.5-short-solutions} Sample task completion errors for gpt3.5-turbo (rows 1-3), gpt-4o-mini (4 - 6), gpt-4o (7 - 9 ), o1-mini (10 - 12) and o1 (13 - 15).}
\end{table*}

\subsection{Handback control error examples}
\label{appendix:handback-control-errors}
\begin{table*}[htpb]
  \centering
  \tiny
  \resizebox{\textwidth}{!}{
  \begin{tabular}{l p{10cm} p{6cm}}
    \toprule
    \textbf{Id} & \textbf{Query} & \textbf{Agent action} \\
    \midrule
    \addlinespace[1.0ex]
    \multirow{2}{*}{1} & \multirow{2}{*}{Assistant, find a suitable conference room for a meeting with my team I wanna schedule later today.} & Tries to schedule a meeting, handing back control because of incorrect diary checking.\\
    \multicolumn{3}{p{\textwidth}}{\textbf{Error cause:} Distracted by irrelevant information. The agent is not required to schedule a meeting, not enough details are provided. Instead, it should have searched for a room that is available and has sufficient capacity to accommodate the user and their team.}\\
    \midrule
    \addlinespace[1.0ex]
    2 & Assistant, can you find a time slot in my diary today when I could schedule something with the HR department to discuss my performance review?& Hallucinates a program attempting to find HR team, handing back control because it cannot determine it. \\
    \addlinespace[1.0ex]
    \multicolumn{3}{p{\textwidth}}{\textbf{Error cause:} Distracted by irrelevant info. The HR team is not defined in the simulation. The task requires the agent to find a slot in user's diary.}\\
    \midrule
    \addlinespace[1.0ex]
    3 & Assistant, schedule our team Christmas party 10 days before Christmas. Should start in the morning and end at 10 PM? & Requires the user to provide an alternative date. \\        
    \addlinespace[1.0ex]
    \multicolumn{3}{p{\textwidth}}{\textbf{Error cause:} Following policy. The agent follows the instruction \textit{Unless the user explicitly states the date, meetings should not be scheduled on or recur during weekends.} which is irrelevant.}\\
    \midrule
    \addlinespace[1.0ex]
    4 & Assistant, schedule a follow-up meeting two weeks after my last one-on-one with my manager. & Hand back control because it cannot find the 1:1 meeting. \\
    \addlinespace[1.0ex]
    \multicolumn{3}{p{\textwidth}}{\textbf{Error cause:} Documentation comprehension. The agent fails to follow a note according to which the user should not be specified as an attendee during search by convention. The note is included in \texttt{find\_events} docs and referenced in \texttt{find\_past\_event} documentation.}\\
    \midrule
    \addlinespace[1.0ex]
    \multirow{2}{*}{5} & \multirow{2}{*}{Assistant, move back my meeting with John from sales and Jane by one hour.} & Hands back control because it determines two employees named John are part of the sales team. \\
    \multicolumn{3}{p{\textwidth}}{\textbf{Error cause:} Unwarranted disambiguation. The event can uniquely determined by checking the calendar.}\\
    \bottomrule
  \end{tabular}
  }
  \caption{\label{tab:o1-handback-control} Examples of queries where o1 mistakenly hands back control to the user.}
\end{table*}

\clearpage

\subsection{Problem categories}
\label{appendix:problem-categories}

In \cref{tab:subset-complexity}, we report task success for five problem categories. Table \ref{tab:problem-category-examples} lists the queries which  were used to estimate the performance per problem category. For each query, a model predicts three AEPs with different random seeds, so $30$ task completion outcomes are considered when estimating subset performance. 

\begin{table*}[htpb]
    \centering
    \tiny
    \resizebox{\textwidth}{!}{ 
    \begin{tabular}{p{\textwidth}}      
    \toprule
      \textbf{Simple} \\
      \midrule
    Assistant, how many meetings with Jianpeng are in my calendar at the moment? \\
    Assistant, plan a weekend trip to the beach with my work colleagues Alice and Bob starting Saturday morning. \\
    Assistant, schedule lunch with my entire team tomorrow at noon. \\
    Assistant, schedule a 3-hour workshop with my team next Monday starting at 1 PM. \\
    Assistant, schedule a meeting with my manager at lunch tomorrow. \\
    Assistant, schedule an urgent meeting with my manager now. \\
    Assistant, share my calendar with my assistant. \\
    Assistant, cancel everything but the important meetings. \\
    Assistant, schedule a team event next Tuesday at 4 PM for 2 hours at the bowling alley. \\
    Assistant, cancel my meeting with Pete and move my meeting with Jianpeng in that slot instead \\
    \\
    \midrule
    \textbf{Constrained scheduling} \\
    \midrule
    Assistant, schedule a project update meeting with my manager when I'm free, before 3 PM tomorrow. \\
    Assistant, schedule a project update meeting with my manager when we're both free, before 3 PM tomorrow. \\
    Assistant, set a 3 to 4 meeting in room z with any team members available then. \\
    Assistant, set a 30 mins meeting with Jianpeng at the earliest time when we are both free today. \\
    Assistant, reschedule today's meetings to Monday - keep the same time. If you detect clashes the rescheduled meetings should start as soon as possible after the end of existing events. No overlaps! \\
    Assistant, find an available slot for a 30-minute meeting with my team two weeks from now. \\
    Assistant, is it possible to schedule a team meeting tomorrow 10 am to 11:30 am or is any colleague from my team busy? \\
    Assistant, check my boss' calendar Wednesday to Friday next week, are they available for a meeting? \\
    Assistant, set up a status update meeting with my manager every last Friday of the month at 2 PM till the end of the year. Skip the ones on his holidays. \\
    Assistant, my manager just told me of a clash with our 1:1 tomorrow, reschedule it to the latest free slot we're available. \\
    \\
    \midrule
    \textbf{Complex time expressions} \\
    \midrule
    Assistant, show me the last time I met with Alice. \\
    Assistant, schedule a 45-minute team follow-up call two weeks after tomorrow's project deadline, keeping the start time. \\
    Assistant, schedule our team Christmas party 10 days before Christmas. Should start in the morning and end at 10 PM. \\
    Assistant, schedule a 1-hour meeting with my manager, then a 45-minute meeting with my team, followed by a 30-minute meeting with the sales team. Add a 15-minute buffer between each meeting starting tomorrow at 9 AM. \\
    Assistant, put 45 minutes in the calendar, back-to-back, with Engineering and Marketing starting at 10 AM tomorrow... Actually, add a 10-minute buffer between each meeting. \\
    Assistant, find an available conference room for my next meeting and schedule it there. \\
    Assistant, book me out of office for the last two hours of the working day the day before my vacation in October. \\
    Assistant, schedule a 1-hour review meeting with my sales team next Monday at 10, then one with finance right after that, and one with engineering after a 30 mins break. \\
    Assistant, block the last hour of the working day for a catch-up with my team the day before any of their vacations start. \\
    Assistant, change our weekly team meeting to happen on Thursday instead, with a update to say 'friday is a no-meeting day'? \\
    \\
  \midrule
  \textbf{Policy / instruction following} \\
  \midrule
    Assistant, schedule a meeting with my team every day next week at 3 PM. \\
    Assistant, plan an off-site event with my team this weekend at Central Park starting at 10 AM. \\
    Assistant, schedule lunch with a different team member each day next week at 12:30 PM. \\
    Assistant, block 90 mins of focus time every morning at 8 AM for the next two weeks. \\
    Assistant, I've got an urgent task that needs 3 hours starting at 1 PM tomorrow. Reschedule my existing meetings to fit this in, but try to keep the same day. \\
    Assistant, schedule a meeting with my team late afternoon tomorrow. Mark Alice optional. \\
    Assistant, reorganise my diary on the fifth so that the important meetings come first. \\
    Assistant, add a strategy review with the CFO and the COO one week from today at 2:30 PM, for 1 hr. \\
    Assistant, set 30 minutes tomorrow late afternoon with the department heads from engineering, finance and marketing. \\
    Assistant, add a reminder 1 hour before all important meetings, with the meeting title in the subject. \\
    \midrule
      \textbf{Advanced problem solving} \\
    \midrule
    Assistant, find a suitable conference room for a meeting with my team I wanna schedule later today. \\
    Assistant, see if my boss' boss and Jane have accepted my meeting request for tomorrow. If anybody declined, reschedule to take place later but at the earliest available time for everyone, I'm free all day. \\
    Assistant, schedule a meeting in the afternoon with my engineering colleagues, avoiding any engineering management. \\
    Assistant, find an available conference room for my next meeting and schedule it there. \\
    Assistant, block 2 hours of free time for holiday preparation after dinner on the last working day before my next vacation. \\
    Assistant, I will need to schedule an important retrospective sometime next week, how many rooms accommodating between 8 and 12 people do we have? \\
    Assistant, add a finance manager to my meeting with the marketing manager. \\
    Assistant, who in finance is yet to book a holiday this year? \\
    Assistant, Ari and James are on holiday next month, who's out for longer? \\
    Assistant, what's ratio  of Diarmuid to Anders holidays from the start of the year till the second of July? \\
  \\
    \bottomrule
    \end{tabular}
}
\caption{Listing of queries for which task success is reported in \cref{tab:subset-complexity}.}
\label{tab:problem-category-examples}
\end{table*}
\clearpage
\twocolumn

\subsection{Primitive selection}
\label{appendix:primitive-selection-expanded-results}

Below, we report primitive selection results broken down for three key \asperaAssistantCodebase{} modules. "Task success" represents the task success rate for queries whose sample solution made use of the primitive in question. The final row shows the global precision, global recall, micro F1 and mean task success across primitives in the module.

\begin{table}[htbp]
  \tiny
  \resizebox{\columnwidth}{!}{
  \begin{tabular}{p{2.5cm}cccc}
    \multicolumn{5}{c}{\texttt{work\_calendar}} \\
    \hline
    \textbf{Primitive} & \textbf{Precision} & \textbf{Recall} & \textbf{F1} & \textbf{CCK task success (1-shot)} \\
    \hline
    find\_past\_events & 0.83 & 0.91 & 0.87 & 0.73 \\
    RepetitionSpec & 0.76 & 0.94 & 0.84 & 0.68 \\
    find\_events & 0.97 & 0.71 & 0.82 & 0.73 \\
    summarise\_calendar & 1.00 & 0.67 & 0.80 & 0.67 \\
    get\_default\_preparation\_time & 0.67 & 1.00 & 0.80 & 0.00 \\
    Event & 0.73 & 0.78 & 0.76 & 0.64 \\
    delete\_event & 0.79 & 0.65 & 0.71 & 0.78 \\
    find\_available\_slots & 0.73 & 0.64 & 0.68 & 0.76 \\
    get\_calendar & 0.73 & 0.55 & 0.63 & 0.69 \\
    add\_event & 0.97 & 0.41 & 0.58 & 0.66 \\
    CalendarSearchSettings & 0.29 & 0.50 & 0.36 & 0.75 \\
    ShowAsStatus & 0.33 & 0.12 & 0.18 & 0.56 \\
    get\_search\_settings & 0.33 & 0.09 & 0.14 & 0.73 \\
    \hline
    \textbf{Overall} & 0.62 & 0.61 & 0.61 & 0.66 \\
    \hline
  \end{tabular}
  }\\
  \vskip 3mm
  \resizebox{\columnwidth}{!}{
  \begin{tabular}{p{2.5cm}cccc}
   \multicolumn{5}{c}{\texttt{company\_directory}} \\
    \hline
    \textbf{Primitive} & \textbf{Precision} & \textbf{Recall} & \textbf{F1} & \textbf{CCK task success (1-shot)} \\
    \hline
    get\_all\_employees & 0.98 & 0.83 & 0.90 & 0.71 \\
    get\_employee\_profile & 0.87 & 0.92 & 0.90 & 0.71 \\
    get\_current\_user & 0.89 & 0.89 & 0.89 & 0.72 \\
    Team & 0.97 & 0.79 & 0.87 & 0.67 \\
    find\_reports\_of & 0.95 & 0.77 & 0.85 & 0.81 \\
    find\_employee & 0.92 & 0.77 & 0.84 & 0.74 \\
    find\_team\_of & 0.98 & 0.68 & 0.81 & 0.74 \\
    get\_vacation\_schedule & 0.73 & 0.83 & 0.77 & 0.76 \\
    get\_assistant & 1.00 & 0.60 & 0.75 & 1.00 \\
    find\_manager\_of & 0.86 & 0.63 & 0.73 & 0.65 \\
    Employee & 0.05 & 0.50 & 0.09 & 0.75 \\
    \hline
    \textbf{Overall} & 0.66 & 0.74 & 0.70 & 0.74 \\
    \hline
  \end{tabular}
  }\\
  \vskip 3mm
  \resizebox{\columnwidth}{!}{
  \begin{tabular}{p{2.5cm}cccc}
     \multicolumn{5}{c}{\texttt{room\_booking}} \\
    \hline
    \textbf{Primitive} & \textbf{Precision} & \textbf{Recall} & \textbf{F1} & \textbf{CCK task success (1-shot)} \\
    \hline
    room\_booking\_default\_time\_window & 0.75 & 1.00 & 0.86 & 1.00 \\
    find\_available\_time\_slots & 0.50 & 0.50 & 0.50 & 0.50 \\
    search\_conference\_room & 0.30 & 0.89 & 0.45 & 0.72 \\
    summarise\_availability & 0.06 & 1.00 & 0.12 & 0.67 \\
    \hline
    \textbf{Overall} & 0.27 & 0.42 & 0.33 & 0.72 \\
    \hline
  \end{tabular}
  }\\
  \label{tab:primitive-selection-expanded-work-calendar}
    \caption{Primitive selection results broken down for three \asperaAssistantCodebase{} modules. The final column shows o1's task success in the CCK setting for the subset of queries whose sample solution made use of the primitive in question. This can be thought of as a proxy for how well the model is able to make use of this tool, in contrast to how well it is able to select it.}
\end{table}

\renewcommand{\arraystretch}{1.15}
\setlength{\tabcolsep}{1pt}

\begin{table*}[ht]
  \centering\small
  \begin{tabularx}{\textwidth}{l
      >{\centering\arraybackslash}p{3.2cm}
      >{\centering\arraybackslash}p{1.4cm}
      >{\centering\arraybackslash}p{1.9cm}
      >{\centering\arraybackslash}p{1.9cm}
      >{\centering\arraybackslash}p{1.6cm}
      >{\centering\arraybackslash}p{1.6cm}
      >{\centering\arraybackslash}p{1.6cm}}
    \toprule
    \multirow{2}{*}{\textbf{Model}} &
    \multirow{2}{*}{\textbf{Checkpoint}} &
    \multirow{2}{*}{\makecell{\textbf{Task}\\\textbf{success}\\(\%)}} &
    \multicolumn{2}{c}{\textbf{Task success (lenient)}} &
    \multirow{2}{*}{\makecell{\textbf{Solution}\\\textbf{err.\ rate}}} &
    \multirow{2}{*}{\makecell{\textbf{Execution}\\\textbf{err.\ rate}}} &
    \multirow{2}{*}{\makecell{\textbf{Handback}\\\textbf{control}\\\textbf{err.\ rate}}} \\
    \cmidrule(lr){4-5}
      & & & \makecell{\textbf{ASPERA}\\\textbf{Imports}} & \makecell{\textbf{Future}\\\textbf{Imports}} & & & \\ \midrule
    \rowcolor{LightGray} o1                & o1-preview-2024-09-12           & 80.13 & ---    & ---    & 62.23 & 9.61  & 28.15 \\
    o3                & o3-2025-04-16                & 75.07 & 77.73  & 81.33  & 34.70 & 40.74 & 24.56 \\
    o3-mini           & o3-mini-2025-01-31           & 64.27 & 67.30  & ---    & 59.52 & 19.32 & 21.17 \\
    GPT-4o (May~25)   & gpt-4o-2024-11-20            & 52.00 & ---    & ---    & 64.16 & 21.11 & 14.74 \\
    \rowcolor{LightGray} o1-mini           & o1-mini-2024-09-12              & 51.40 & ---    & ---    & 57.43 & 16.76 & 25.81 \\
    \rowcolor{LightGray} GPT-4o            & gpt-4o-2024-05-13               & 45.33 & ---    & ---    & 49.75 & 38.79 & 11.46 \\
    \rowcolor{LightGray} GPT-4o-mini       & gpt-4o-mini-2024-07-18           & 21.07 & ---    & ---    & 49.82 & 42.91 & 7.27 \\
    \rowcolor{LightGray} gpt-3.5-turbo     & gpt-3.5-turbo-0125               & 10.80 & ---    & ---    & 34.23 & 2.96  & 62.81 \\ \midrule
    2.5-flash         & gemini-2.5-flash-preview-05-20 & 59.33 & 69.60  & ---    & 39.02 & 49.83 & 11.15 \\
    2.0-flash         & gemini-2.0-flash-001           & 50.67 & 52.27  & ---    & 70.81 & 17.57 & 11.62 \\
    \rowcolor{LightGray} 1.5-pro           & gemini-1.5-pro-002              & 33.73 & ---    & ---    & 53.54 & 39.22 & 7.24  \\
    \rowcolor{LightGray} 1.5-flash         & gemini-1.5-flash-002            & 27.87 & ---    & ---    & 46.23 & 45.83 & 7.95  \\
    \rowcolor{LightGray} 1.0-pro           & gemini-1.0-pro-002              & 12.67 & ---    & ---    & 27.49 & 65.49 & 7.02  \\
    \bottomrule
  \end{tabularx}
  \caption{CCK task-success evaluation for OpenAI and Gemini model families. ``---'' indicates import errors do not affect these models. \textbf{Shaded rows} repeat results reported in \Cref{tab:cck-results-main} and \Cref{fig:error_breakdown} to facilitate comparisons.}
  \label{tab:results}
\end{table*}

\section{Evaluation supplementary material}
\label{appendix:extended-evals}
Given the rapid evolution of LLM capabilities, we additionally evaluate several models from the OpenAI and Gemini families released after our manuscript submission in December 2024. \Cref{tab:results} summarizes the task success rates achieved by these models in the CCK setting.

\paragraph{OpenAI}
The GPT-4o release evaluated at the time of submission (May 2025) exhibits a $6.67\%$ increase in task success compared to the variant we assessed in this paper (September 2024), matching the performance of o1-mini. Meanwhile, the o3-mini model demonstrates a substantial $12.27\%$ improvement in task success over the latest GPT-4o, attributed to its superior reasoning capabilities. However, notably, our evaluation shows that o3 underperforms compared to o1 by a $5.06\%$ margin.

The high execution error rates of o3 and o3-mini prompted further analysis of their generated AEPs. This revealed that a considerable number of errors stemmed from improper module imports. Specifically, ASPERA imports were mistakenly included by these models despite explicit instructions (as detailed in \Cref{fig:agent-guidelines}) to only import standard library modules, as necessary, because ASPERA-specific imports are automatically handled at runtime. Additionally, o3 occasionally introduced unnecessary future imports (e.g., \texttt{from \_\_future\_\_ import annotations}), even after the prompt explicitly specified the Python version for AEP implementation. This behaviour did not changed when the AEP generation prompt was updated to include the \texttt{python} version the code should target.

While these errors indicate diminished instruction-following capabilities and redundant code generation tendencies, the primary purpose of \asperaDataset{} is to evaluate an agent's capability to perform complex, multi-step reasoning tasks. Upon manually correcting import-related errors and re-executing the programs, task success increased by $3.05\%$ for o3-mini and $6.26\%$ for o3. The results show that despite discounting instruction adherence errors, o3 does not demonstrate substantially better performance compared to o1 in executing \asperaDataset{} queries.

\paragraph{Gemini}
Similar issues with disregarding import handling instructions were observed in the latest Gemini models. For the 2.0-flash model, this behavior resulted in a modest $1.6\%$ difference in task success, whereas for 2.5-flash, the discrepancy was significantly larger at $10.27\%$.

Overall, our findings indicate improvements in models' abilities to handle complex queries. However, particularly challenging queries that demand advanced reasoning and creative tool use (such as query 3 in \Cref{tab:query-sample}) continue to elude all evaluated models. Systematic analysis of these challenging cases can highlight specific weaknesses, guiding the development of increasingly demanding benchmarks as model capabilities evolve -- a direction we leave open for future research.
\newpage
\section{Comparison with other benchmarks}
\label{appendix:literature-dataset-examples}
This appendix extends \S \ref{sec:related-work}, providing further comparison between \asperaDataset{} and existing benchmarks in tool use (\S \ref{sec:mtu}), LLM agent evaluation (\S \ref{sec:ose}), and code generation (\S \ref{sec:codegen}). While our focus is on \asperaDataset{}, it is important to note that developers can extend \aspera{} to new domains, and use our data generation engine to create high-quality datasets alongside robust evaluation programs—a key contribution of our work.
\subsection{Multiple tool use datasets}
\label{sec:mtu}
Evaluating complex action execution in digital assistants requires datasets grounded in realistic, multi-step queries that integrate multiple tools (\S \ref{sec:intro}). Several benchmarks focus on multi-tool queries, but each has limitations which make them unsuitable for assessing complex action execution capabilities in digital assistants. We consider popular datasets, referring the reader to \citet{toolusereview} for an in-depth review of tool-use datasets.
\begin{figure*}[t]
  \centering
  \includegraphics[width=\textwidth]{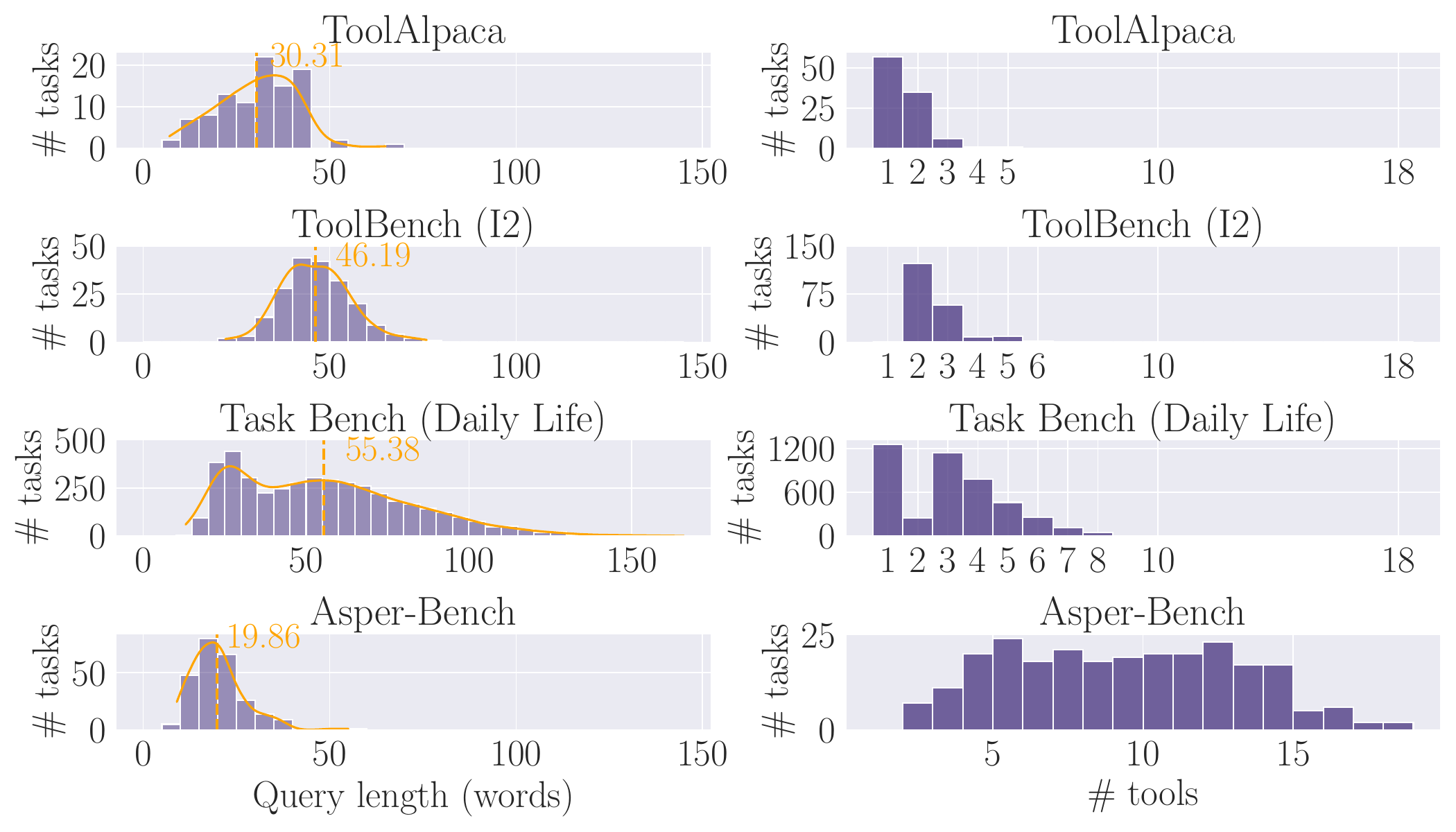}
  \caption{Comparison of query length and action distribution between ToolAlpaca \cite{toolalpaca}, ToolBench \cite{DBLP:conf/iclr/QinLYZYLLCTQZHT24}, TaskBench \cite{taskbench} and \asperaDataset{} (generated with the \aspera{} framework, \S\ref{sec:dataset}).}
  \label{fig:dataset-compare}
\end{figure*}
\begin{table*}[h!]
\centering\small
\begin{tabular}{lclp{11cm}}
    \toprule
     \textbf{ID} & \textbf{Number of tools}  & \textbf{Corpus} & \textbf{Query}\\
    \midrule
    \multirow{3}{*}{1} & \multirow{3}{*}{1} & \multirow{7}{*}{ToolAlpaca} & I'm sending a package to my friend in New York, but I'm not sure if I have the correct address. Can you check if this address is valid and deliverable? Here's the address: 123 Main St, Apt 4B, New York, NY, 10001. \\
    \multirow{4}{*}{2} & \multirow{4}{*}{3} & & There's an update to our data source, the connection string has changed. Can you modify the existing data source called "SalesDB"? After you've done that, I'd also like to add a new chart to our "Sales Overview" dashboard, called "Top Selling Products". Don't forget to update the dashboard name to "Complete Sales Overview". \\
    \midrule
    \multirow{4}{*}{3} & \multirow{4}{*}{2} & \multirow{8}{*}{ToolBench} & I'm working on a personal project and I need to gather a large number of random anime images. Can you provide me with around 5000 random anime images from the Random anime img API? Additionally, I would like to create profile images for my project using the Image Service API. \\
    \multirow{4}{*}{4} & \multirow{4}{*}{3} & & I'm planning a family road trip and I want to create a playlist of MP3 songs. Can you convert the audio from the YouTube videos with the ids 'UxxajLWwzqY', 'abc123', 'xyz456' to MP3 format? Please provide the download links and make sure the converted files are free of any profanity. \\
    \midrule
    \multirow{3}{*}{8} & \multirow{3}{*}{3} & \multirow{11}{*}{\parbox{2cm}{TaskBench \\ (daily life)}} & I'm enrolled in a Data Science Conference happening on May 15, 2023. Could you help me manage the logistics? Let's start by scheduling a flight from Los Angeles to New York for the conference day and ensure I'm reminded of the meeting at 2 PM. \\[2em]
    \multirow{8}{*}{9} & \multirow{8}{*}{7} & & I'm planning on applying for a software development job soon and I also need to purchase a new Smartphone from Amazon for my everyday tasks. Could you assist me with these tasks? I also need to prepare for the interview, so I would appreciate it if you could help me record notes on a few topics such as data structures, problem solving, and algorithm design. Plus, I want to record an audio file named 'example.wav'. After purchasing the Smartphone, could you make sure it is delivered to my home address and send me an SMS on 1234567890 to confirm its arrival? By the way, could you install the Zoom application on my computer to facilitate video conferencing. \\
    \bottomrule
\end{tabular}
\caption{\label{tab:literature-dataset-examples} Samples from multiple tool use corpora.}
\end{table*}

\paragraph{ToolAlpaca}  \citet{toolalpaca} seed ChatGPT3.5 with crawled API names and intended use information to synthesise documentation along with user queries, agent actions and simulated environment response. The resulting instructions primarily involve a small number of API calls (\cref{fig:dataset-compare}), making them better suited for evaluating argument parsing rather than complex multi-tool reasoning. The quality of queries deteriorates as more tools are incorporated. As observed in prior work \cite{qualitymatters}, API call annotations frequently contain hallucinations or missing arguments, limiting the dataset's suitability for our setting\footnote{In \cref{tab:literature-dataset-examples}, row 2, the user specifies the "SalesDB" connection string has changed and has to be modified, but does not specify the new value.} .

\paragraph{ToolBench} \cite{DBLP:conf/iclr/QinLYZYLLCTQZHT24} improves upon ToolAlpaca by incorporating real-world API documentation instead of synthesising it. While this allows for more diverse tool-use scenarios, multi-tool queries remain sparse in the evaluation set (\cref{fig:dataset-compare}), restricting opportunities to assess complex tool interactions.

Our ToolBench analysis revealed that direct synthesis using real-world API documentation affects query naturalness in several ways. For example, the API names are directly mentioned in the query (\cref{tab:literature-dataset-examples}, row 3), a problem which increasingly affects coherence as the number of APIs invoked in the query increases(\cref{tab:literature-dataset-examples}, row 4). This arises because the sampled APIs are designed for a wide variety of high-level tasks (e.g., video download, web crawling, weather report, etc) amongst which relationships are sparse and which cannot be naturally combined to synthesise natural complex tasks. \citet{qualitymatters} conduct an in-depth study focused on the impact of query synthesis from API documentations on query quality.

\paragraph{TaskBench} \citet{taskbench} study LLMs for task automation. The authors recognise that task complexity is not only dependent on the number of tools, but also on the relationships between them, which the documentation- and template-based synthesis approaches of \citet{DBLP:conf/iclr/QinLYZYLLCTQZHT24} and \citet{toolalpaca} do not model. To address this, they ground query generation in a graph which encodes tool dependencies. In their framework, single-node graphs model simple tasks and more general chain and directed-acyclic graphs (DAGs) ground tasks with higher complexity. \cref{tab:literature-dataset-examples} shows an example of a query grounded in DAG tool graph (row 8). These are more complex and natural compared to ToolBench queries as API parameters are shared across tasks, and, more generally, API inputs can depend on the output of previous tasks. However, such relationships become sparse at the number of tools increases, and, as result, queries grounded in chain graphs with multiple nodes (row 9) are unnatural and pose limited additional challenges to LLMs compared to parsing single API calls.

\subsection{LLM agent benchmarks}
\label{sec:ose}
A growing body of research focuses on developing autonomous LLM-based agents, primarily for web-based tasks \cite{webarena, web1, web2, web3, web4, webvisual}, home automation \cite{webvisual2}, mobile development \cite{androiddev}, open-ended computer tasks \cite{osagent}, and workplace assistance \cite{workflowllm, officebench}. Some of these environments are designed to support reinforcement learning research \cite{androiddev}, while others benchmark visually grounded agents \cite{webvisual, webvisual2}. In contrast, \asperaDataset{} targets digital assistant capabilities (e.g., Alexa, Siri), emphasising the parsing of complex user actions into executable programs rather than long-horizon planning, a central theme in most LLM agent benchmarks.
\noindent
\begin{table}[t]
  \vspace{0.1cm}
  \centering
  \tiny
  \resizebox{\linewidth}{!}{%
    \begin{tabular}{lcccc}
      \toprule
      \textbf{Benchmark} & \textbf{Dynamic}      & \textbf{Realistic}    & \textbf{Diverse} &      \textbf{Functional}\\
                         & \textbf{Interaction?} & \textbf{Environment?} & \textbf{Human Tasks?} & \textbf{Correctness?}\\
      \midrule
      Mind2Web \cite{web4}       & \textcolor{red}{\ding{55}} & \textcolor{green}{\ding{51}} & \textcolor{green}{\ding{51}} & \textcolor{red}{\ding{55}} \\
      Form/QAWeb \cite{web1}     & \textcolor{red}{\ding{55}} & \textcolor{green}{\ding{51}} & \textcolor{green}{\ding{51}} &  \textcolor{red}{\ding{55}} \\
      MiniWoB++ \cite{web2}      &  \textcolor{green}{\ding{51}} & \textcolor{red}{\ding{55}} & \textcolor{red}{\ding{55}} & \textcolor{green}{\ding{51}} \\
      WebShop \cite{web3}        &  \textcolor{green}{\ding{51}} & \textcolor{red}{\ding{55}} & \textcolor{red}{\ding{55}} & \textcolor{green}{\ding{51}} \\
      ALFRED \cite{webvisual}    & \textcolor{green}{\ding{51}} & \textcolor{red}{\ding{55}} & \textcolor{red}{\ding{55}} & \textcolor{green}{\ding{51}} \\
      VirtualHome \cite{webvisual2} & \textcolor{red}{\ding{55}} & \textcolor{red}{\ding{55}} & \textcolor{green}{\ding{51}} & \textcolor{red}{\ding{55}} \\
      AndroidEnv \cite{androiddev} & \textcolor{green}{\ding{51}} & \textcolor{green}{\ding{51}} & \textcolor{red}{\ding{55}} & \textcolor{red}{\ding{55}} \\
      WebArena \cite{webarena} & \textcolor{green}{\ding{51}} & \textcolor{green}{\ding{51}} & \textcolor{green}{\ding{51}} & \textcolor{green}{\ding{51}} \\
      WorkflowLLM \cite{workflowllm} & \textcolor{red}{\ding{55}} & \textcolor{red}{\ding{55}} & \textcolor{red}{\ding{55}} & \textcolor{red}{\ding{55}} \\
      OfficeBench \cite{workflowllm} & \textcolor{green}{\ding{51}} &  \textbf{\textcolor{black}{N/A}}& \textcolor{green}{\ding{51}} & \textcolor{green}{\ding{51}} \\
      \midrule
      \textbf{\asperaDataset{}} (ours) & \textcolor{green}{\ding{51}} & \textbf{\textcolor{black}{N/A}} & \textcolor{green}{\ding{51}} & \textcolor{green}{\ding{51}} \\
      \bottomrule
    \end{tabular}%
  }
  \captionof{table}{Comparison of \asperaDataset{} with agent benchmarks.}
  \label{tab:comparison}
\end{table}

One key distinction between \asperaDataset{} and existing agent benchmarks lies in action space complexity. Web-based agent benchmarks typically define small, discrete action sets; for example, WebArena \cite{webarena} includes just 12 actions, with simple textual descriptions such as \texttt{new\_tab} (\textit{Open a new tab}). In contrast, \asperaDataset{} features 69 actions, including both high-level commands like \texttt{delete\_event} and low-level primitives such as \texttt{get\_next\_dow}\footnote{A function for computing the next occurrence of a specified weekday.}. This richer action space supports nuanced execution, requiring reasoning over fine-grained dependencies instead of following step-by-step workflows.. As such \asperaDataset{} is a code generation benchmark which tests language understanding, logical reasoning and short-term planning capability \textit{when the agents are grounded in fine-grained dependencies, unseen during pre-training} whereas other benchmarks provide a complementary view of LLMs' long-term planning capability.

\Cref{tab:comparison} compares \asperaDataset{} with other benchmarks across four key dimensions: whether the environment allows dynamic interaction, whether tasks are grounded in realistic environments, whether datasets include diverse human-verified tasks, and whether the benchmark ensures functional correctness through execution. Unlike most of these, \aspera{} supports dynamic interaction\footnote{For example, the agents have access to the stack trace. See \texttt{src/aspera/evaluator.py::get\_solution\_feedback} in our code release.} and functional correctness evaluation. Moreover, \asperaDataset{} is a diverse collection of human-verified tasks. \aspera{} uniquely enables developers to generate high-quality tasks and evaluation code through LLM interaction to support benchmarking on custom use cases.

The simulation fidelity depends on task complexity and is more readily achieved for web benchmarks which rely on widely used open-source technologies. In contrast, more complex environments such as OfficeBench \cite{officebench} and digital assistants are difficult to simulate since they rely on proprietary, commercial technologies. In \aspera{}, we tackle this by implementing a simplified but fine-grained \texttt{python} simulation of a fictitious corporate calendar-management application which supports our objective of evaluating LLMs' capability of complex action execution via program synthesis.

\subsection{Other code generation benchmarks}
\label{sec:codegen}
As discussed in \S \ref{sec:related-work}, \asperaDataset{} is a code-generation benchmark which tests LLMs complex action execution capability given \textit{custom, project-runnable dependencies}. This significantly more challenging setting is uncommon, as it requires a custom simulation environment \cite{DBLP:journals/corr/abs-2404-10155}. In contrast, function generation focusing on competitive programming \cite{evalplus} requires only standard library dependencies, whereas more general software capability benchmarks \cite{bigcodebench} assess program generation based on widely used dependencies seen in training (e.g., \texttt{numpy} apis). Consequently, \asperaDataset{} complements existing benchmarks by assessing program generation under custom dependencies. While prior benchmarks (e.g., \cite{evalplus}) are increasingly saturated by strong LLMs, \S \ref{sec:results} shows \asperaDataset{} remains challenging. However, \asperaDataset{} performance correlates with model ability on popular LLM code benchmarks (\cref{tab:aspera_correlation}).
\vspace{-5cm}

\begin{table}[t]
\centering
\resizebox{\columnwidth}{!}{
\begin{tabular}{lcc}
\toprule
\textbf{Benchmark} & \textbf{Spearman \(r\)} & \textbf{Pearson \(p\)} \\
\hline
EvalPlus \cite{evalplus}       & 0.8909                  & 0.8909                 \\
BigCode (Hard) \cite{bigcodebench}     & 0.9879                  & 0.9879                 \\
\hline
\end{tabular}
}
\caption{Correlation between model ranks on ASPERA with other standard code generation benchmarks. Scores for Gemini 1.5-Flash and 1.0-Pro are not reported on the BigCode (Hard) leaderboard and EvalPlus leaderboard does not include CodeGemma (27B). Hence, we exclude these models from the analyses.}
\label{tab:aspera_correlation}
\end{table}
\end{document}